\documentclass{article}

\usepackage{microtype}
\usepackage{graphicx}
\usepackage{subfigure}
\usepackage{xcolor}
\usepackage{booktabs} 

\usepackage{hyperref}

\usepackage[accepted]{icml2024}

\usepackage{amsmath}
\usepackage{amssymb}
\usepackage{mathtools}
\usepackage{amsthm}

\usepackage[capitalize,noabbrev]{cleveref}

\usepackage{url}
\usepackage{enumitem}
\usepackage{graphicx}
\usepackage{caption}
\usepackage{booktabs}
\usepackage{adjustbox}
\usepackage{listings}
\usepackage{xcolor}

\usepackage{longtable}
\theoremstyle{plain}

\theoremstyle{definition}

\theoremstyle{remark}

\definecolor{c1}{RGB}{24, 209, 207}
\definecolor{c2}{RGB}{21, 182, 198}
\definecolor{c3}{RGB}{18, 156, 189}
\definecolor{c4}{RGB}{16, 130, 180}
\definecolor{c5}{RGB}{13, 104, 171}
\definecolor{c6}{RGB}{10, 78, 162}
\definecolor{c7}{RGB}{8, 52, 154}

\usepackage[textsize=tiny]{todonotes}

\icmltitlerunning{RoboGen: Towards Unleashing Infinite Data for Automated Robot Learning via Generative Simulation}

\begin{document}

\twocolumn[


\icmltitle{
\textcolor{c1}{R}\textcolor{c2}{o}\textcolor{c3}{b}\textcolor{c4}{o}\textcolor{c5}{G}\textcolor{c6}{e}\textcolor{c7}{n}: Towards Unleashing Infinite Data for \\ Automated Robot Learning via Generative Simulation
}



\icmlsetsymbol{equal}{*}

\begin{icmlauthorlist}
\icmlauthor{Yufei Wang}{equal,cmu}
\icmlauthor{Zhou Xian}{equal,cmu}
\icmlauthor{Feng Chen}{equal,thu}
\icmlauthor{Tsun-Hsuan Wang}{mit}
\icmlauthor{Yian Wang}{umass}
\icmlauthor{Katerina Fragkiadaki}{cmu}
\icmlauthor{Zackory Erickson}{cmu}
\icmlauthor{David Held}{cmu}
\icmlauthor{Chuang Gan}{umass,ibm}
\end{icmlauthorlist}

\icmlaffiliation{cmu}{CMU}
\icmlaffiliation{thu}{Tsinghua IIIS}
\icmlaffiliation{mit}{MIT CSAIL}
\icmlaffiliation{umass}{UMass Amherst}
\icmlaffiliation{ibm}{MIT-IBM AI Lab}

\icmlcorrespondingauthor{Yufei Wang}{yufeiw2@andrew.cmu.edu}

\icmlkeywords{Machine Learning, ICML}

\vskip 0.3in
]



\printAffiliationsAndNotice{\icmlEqualContribution} 

\newcommand{\model}{RoboGen}

\newcommand{\etal}{\text{et al.}}
\newcommand{\yufei}[1]{\textcolor{blue}{yufei: #1}}
\newcommand{\xian}[1]{\textcolor{red}{xian: #1}}

\begin{abstract}
We present \textit{\model},   a \textit{generative robotic agent} that automatically learns diverse robotic skills at scale via \textit{generative simulation}. \model{} leverages the latest advancements in foundation and generative models. Instead of directly adapting these models to produce
policies or low-level actions, we advocate for a generative scheme, which uses these models to automatically generate diversified tasks, scenes, and training supervisions, thereby scaling up robotic skill learning with minimal human supervision.
Our approach equips a robotic agent with a self-guided \textit{propose-generate-learn} cycle: the agent first proposes interesting tasks and skills to develop, and then generates simulation environments by populating pertinent assets with proper spatial configurations. Afterwards, the agent decomposes the proposed task into sub-tasks, selects the optimal learning approach (reinforcement learning, motion planning, or trajectory optimization), generates required training supervision, and then learns policies to acquire the proposed skill. 
Our fully generative pipeline can be queried repeatedly, producing an endless stream of skill demonstrations associated with diverse tasks and environments. 
\vspace{-10pt}
\end{abstract}


\section{Introduction}

Simulated environments have become a crucial driving force for teaching robots various complex skills, spanning complex manipulation and locomotion settings \citep{weng2022fabricflownet, xu2023roboninja, chen2022system, haarnoja2023learning, zhuang2023robot}. Compared to exploration and data collection in the real-world, simulated environments provide access to privileged low-level states and unlimited explorations, and support massively parallel computation for significantly faster data collection without considerable investment in robotic hardware.
However, robot learning in simulations also presents its own limitations: while exploration and practicing in simulated environments are cost-effective, constructing these environments requires tremendous human effort, demanding tedious steps including designing tasks, producing relevant and semantically meaningful assets, generating plausible scene layouts and configurations, and crafting training supervisions such as reward or loss functions \citep{james2020rlbench, srivastava2022behavior, gu2023maniskill2, li2023behavior}. The onerous task of creating these components and constructing individualized simulation settings for each one of the countless tasks encountered in our daily life significantly hinders the scalability of robotic skill learning even in simulated worlds.

\begin{figure*}[t!]
    \begin{adjustbox}{center}
    \includegraphics[width=.85\textwidth,keepaspectratio]{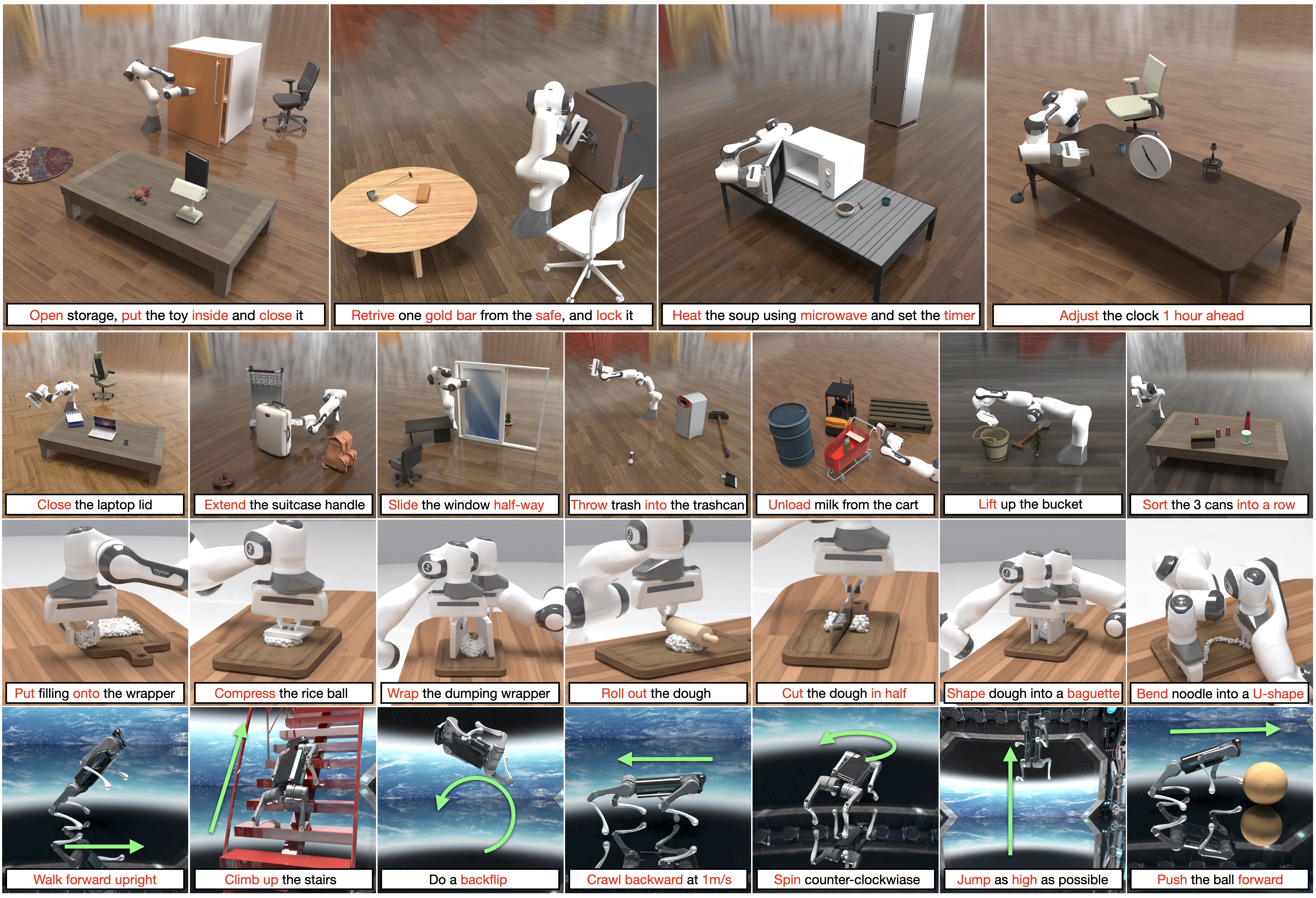}
    \end{adjustbox}
    \vspace{-15pt}
    \caption{
    \small 25 example tasks generated and corresponding skills learned by \model{}. Readers are encouraged to visit our project website for the diverse set of tasks and skills \model{} can produce.
    }
    \vspace{-15pt}
    \label{fig:teaser}
\end{figure*}
In light of this, we propose \textit{Generative Simulation} \citep{xian2023towards}, a new paradigm aiming for scaling up simulated robot learning with the latest advancement in generative models. Generative simulation advocates for \textit{autonomously} generating information for all the stages needed for diverse robotic skill learning in simulation: from high-level task and skill proposals to task-dependent scene descriptions, asset selections and generations, policy learning choices, and training supervisions.  These information is then used for massive skill training, enabling robots to acquire proposed skills. In this paper, as an initial realization of this proposed paradigm, we present \textit{\model{}}, a robotic agent that continuously generates new skills via a self-guided \textit{propose-generate-learn} cycle: it firstly self-proposes skills to learn, and then generates required assets and constructs the scene in simulation conditioned on the proposed task. Afterwards, it labels the tasks with natural language descriptions, decomposes the task into sub-tasks, selects the optimal learning approach (reinforcement learning, motion planning, or trajectory optimization), designs proper training supervisions (e.g. reward functions), and lastly proceeds to policy learning to solve the proposed task. 
One distinct advantage of our proposed paradigm lies in the careful choice of what modes of knowledge to extract from contemporary foundation models. These models have demonstrated impressive capabilities across various modalities \citep{touvron2023llama, driess2023palm, gpt4, rombach2022high, kang2023scaling},
However, due to the absence of training data pertaining to \textit{dynamics}, \textit{actuations}, and \textit{physical interactions}, these models are yet to develop essential understandings for robots to execute physical actions and interact with the surrounding environments (e.g., producing precise joint torques needed for walking or rolling a dough at hand). In contrast to recent efforts that employ foundation models such as Large Language Models (LLMs) for directly yielding policies or low-level actions \citep{liang2022code, huang2023voxposer, wang2023prompt}, our method only extracts information that falls neatly within the capabilities and modalities of these models - object semantics, object affordances, common-sense knowledge regarding what tasks are valuable to learn, etc. These knowledge are used to construct environmental playgrounds, and then augmented with additional help from physics-grounded simulations, for robots to develop understandings of physical interactions and acquire diverse skills. 

Our experiments show that \model{} can deliver a continuous stream of diversified skill demonstrations, spanning tasks including rigid and articulated object manipulation, deformable object manipulation, as well as legged locomotion (see Figure \ref{fig:teaser}). The diversity of tasks and skills generated by \model{} surpasses previous human-crafted robotic datasets, with minimal human involvement beyond several prompt designs and in-context examples. 
Our work attempts to transfer the extensive and versatile knowledge embedded in large-scale models to the field of robotics, making a step towards automated large-scale robotic skill training and demonstration collection for building generalizable robotic systems.
Our code will be made publicly available upon publication. For extensive qualitative results and interactive examples, please refer to our project site at 
\href{https://robogen-ai.github.io/}{\textcolor{red}{\texttt{https://robogen-ai.github.io/}}}.

\section{Related Work}

\textbf{Robotic skill learning in simulations} Various physics-based simulation platforms have been developed in the past to accelerate robotics research \citep{liu2021role}. These include rigid-body simulators \citep{coumans2019, todorov2012mujoco, xiang2020sapien, bousmalis2023robocat}, deformable object simulators \citep{macklin2014unified, lin2020softgym, xu2023roboninja, heiden2021disect}, and environments supporting multi-material and their couplings with robots \citep{xian2023fluidlab, gan2021threedworld, gu2023maniskill2}. Such simulation platforms have been heavily employed in the robotics community for learning diverse skills, including deformable object manipulation \citep{lin2022diffskill, weng2022fabricflownet, wang2023one}, object cutting \citep{heiden2021disect, xu2023roboninja}, fluid manipulation \citep{seita2023toolflownet, xian2023fluidlab}, as well as highly dynamic and complex skills such as in-hand re-orientation \citep{chen2022system, akkaya2019solving}, object tossing \citep{zeng2020tossingbot}, acrobatic flight \citep{kaufmann2020deep, loquercio2021learning, song2023reaching}, and legged locomotion  \citep{cheng2023legs, zhuang2023robot, radosavovic2023learning}. 

\textbf{Scaling up simulation environments}
Apart from building physics engines and simulators, a large body of prior work targeted at building large-scale simulation benchmarks, providing platforms for scalable skill learning and standardized benchmarking \citep{li2023behavior, lin2020softgym, xian2023fluidlab, yu2020meta, james2020rlbench, gu2023maniskill2, srivastava2022behavior}. Notably, most of these prior simulation datasets are manually built with human labeling. Another line of works attempts to scale up tasks and environments using procedural generation, and generate demonstrations with Task and Motion Planning (TAMP) \citep{jiang2023vima, dalal2023imitating, mcdonald2021guided, murali2023cabinet}. These methods primarily build on top of manually-defined rules and planning domains, limiting the diversity of the generated environments and skills to relatively simple pick-and-place tasks \citep{dalal2023imitating,mcdonald2021guided}. Contrary to these works, we leverage the common sense knowledge embedded in foundation models to generate meaningful tasks, relevant scenes, and skill training supervisions, leading to more diverse and plausible skills. 

\textbf{Foundation and generative models for robotics} Following the advancement in foundation and generative models in domains of imagery, language and other modalities, \citep{poole2022dreamfusion, melas2023realfusion, touvron2023llama, driess2023palm, gpt4, liu2023audioldm, girdhar2023imagebind}, a line of works explores using these models for robotics research via approaches such as code generation~\citep{wu2023tidybot, liang2022code}, data augmentation~\citep{yu2023scaling}, visual imagination for skill execution~\citep{du2023learning}, sub-task planning~\citep{ahn2022can, huang2022inner, lin2023text2motion}, concept generalization of learned skills~\citep{brohan2023rt}, outputting low-level control actions~\citep{wang2023prompt}, and goal specification~\citep{kapelyukh2023dall, jiang2023vima}. Related to ours are recent methods using LLMs for reward generation \citep{yu2023language, ma2023eureka}, and sub-task and trajectory generation \citep{ha2023scaling}. Concurrent work \citep{wang2023gen} explored LLM-based task generation, but are limited to table-top rigid object manipulation tasks with limited assets. The task demonstrations are generated by using LLMs to directly write the code script for manipulating the objects; in contrast, we use LLMs to generate the rewards, invoke appropriate algorithms (motion planning, RL, etc) to learn the skill and generate the demonstrations, which is more general. 
Katara et al.~\citep{katara2023gen2sim} also used a LLM to generate table-top rigid and articulated object manipulation tasks and rewards. Ours differ as we additionally perform scene generation and automatic algorithm selection. We also demonstrate our pipeline with more diverse tasks, including more complex and long-horizon articulated object manipulation tasks, as well as locomotion and soft-body manipulation tasks. 

\begin{figure*}
    \centering
    \includegraphics[width=0.9\textwidth]{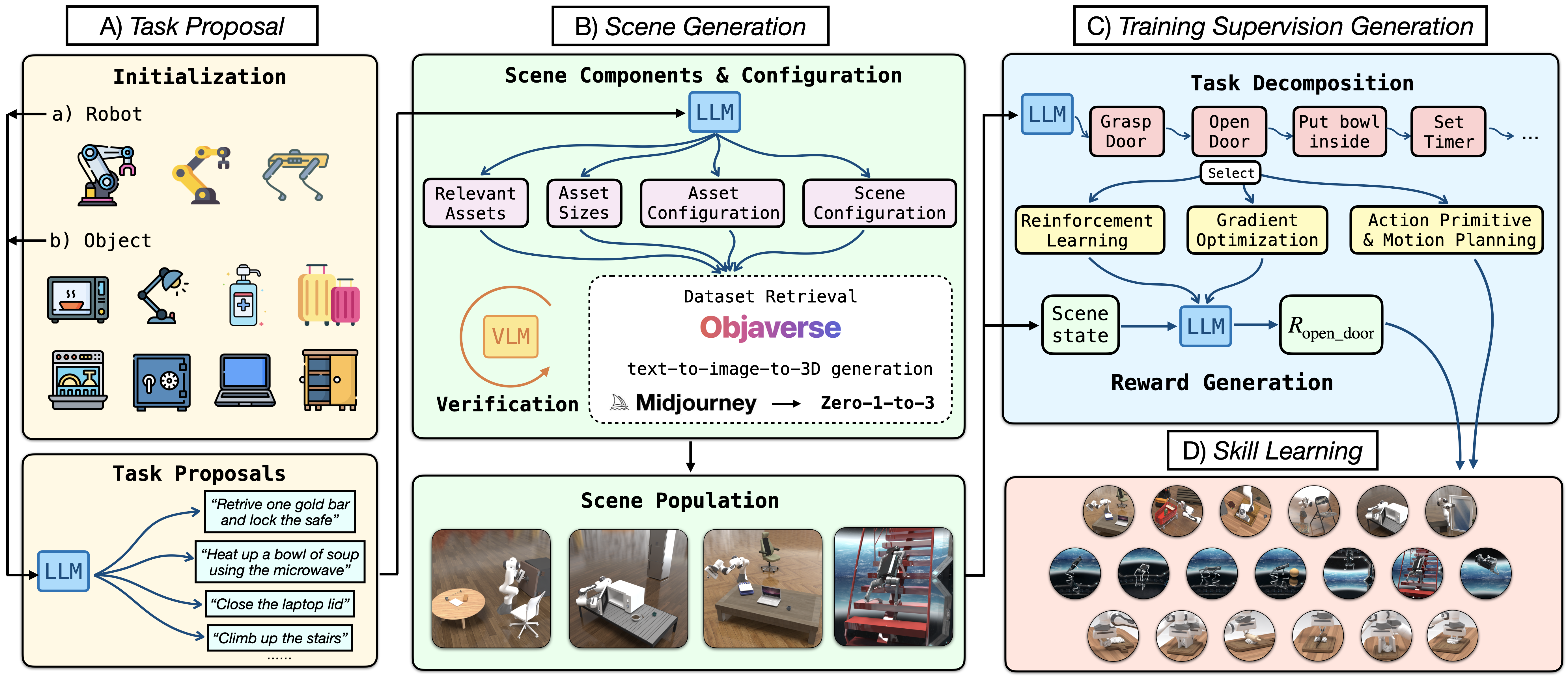}
    \caption{\small \model{} consists of the following stages: A) task proposal, B) scene generation, C) training supervision generation, and D) skill learning with generated information.
    }
    \label{fig:system}
\end{figure*}

\section{\model{}}

\model{} is an automated pipeline that utilizes the embedded common sense and generative capabilities of the latest foundation models~\citep{ChatGPT, alpaca} for automatic task, scene, and training supervision generation, leading to diverse robotic skill learning at scale. We consider tasks including rigid (articulated) object manipulation, soft body manipulation, and legged locomotion. We illustrate the whole pipeline in Figure~\ref{fig:system}, composed of several integral stages: \textit{Task Proposal}, \textit{Scene Generation}, \textit{Training Supervision Generation}, and \textit{Skill Learning}. We detail each of them in the following. 

\subsection{Task Proposal}
\model{} starts with proposing meaningful and diverse tasks for robots to learn. We initialize the system with a specific robot type and an object randomly sampled from a pre-defined pool. The provided robot and sampled object information are then used as input to an LLM to generate task proposal. This initialization step serves as a \textit{seeding} stage, providing a basis upon which the LLM can condition and subsequently reason and extrapolate to generate a variety of tasks, taking into account both robot capability and object affordances. Apart from object-based initialization, another choice is to employ example-based initialization, where we initialize the query with a provided robot and several example tasks sampled from a list of 11 pre-defined tasks (see Appendix \ref{app:ref}). For tasks involving legged robots and soft-body manipulation, we prompt the LLM with only example-based seeding.

We use GPT-4 \citep{gpt4} as our LLM backend to query in the current pipeline, which can be upgraded once better models are available. In the following,  we explain details of \model{} in the context of a robotic arm (e.g., Franka) and tasks generated pertain to object manipulation, using object-based initialization. In this case, the objects used for initialization are sampled from a predefined list, including common articulated and non-articulated objects in household scenarios such as oven, microwave, dispenser, laptop, dishwasher, etc., extracted from PartNetMobility \citep{xiang2020sapien} and RLBench \citep{james2020rlbench}.
The common sense and reasoning capability embedded in LLMs like GPT-4 allow them to produce meaningful tasks considering the object affordances, functionalities, and how they can be interacted with. We instantiate a prompt for task proposal containing the following information: 1) the category of the sampled object, 2) its articulation tree derived from its URDF file, and 3) semantic annotations of the links in the object's articulation, e.g., which link corresponds to the door in a sampled microwave. 
These information are provided by the PartNetMobility dataset. Additionally, we include one example input-output pair in the prompt. We feed the prompt to GPT-4 to obtain a number of semantically meaningful tasks that can be performed with the sampled object, where each task consists of 1) task name, 2) a natural language description of the task, 3) \textit{additional} objects needed for performing the proposed task and 4) joints and links of the sampled articulated object relevant to the task.

As a concrete example, if a sampled articulated object is a \texttt{microwave}, where \texttt{joint\_0} is a revolute joint connecting its door, and \texttt{joint\_1} is another revolute joint controlling the timer knob, GPT-4 could return a task named ``heat up a bowl of soup'', with a task description of ``The robot arm places a bowl of soup inside the microwave, closes the door and sets the microwave timer for an appropriate heating duration'', additional objects that are necessary for the generated task such as ``A bowl of soup'', and task-relevant joints and links including \texttt{joint\_0} (for opening the microwave door), \texttt{joint\_1} (for setting the timer), \texttt{link\_0} (the door), and \texttt{link\_1} (the timer knob). Note that for cases where we sample non-articulated objects or use example-based initialization, the sampled objects and examples are provided only as a hint for task proposal, and the generated tasks will not be tied to them. 
By repeatedly querying with different sampled objects and examples, we can generate a diverse range of manipulation and locomotion tasks, concerning the relevant object affordances when needed.



\subsection{Scene Generation}
\label{sec:scenegen}
Once a task proposal is obtained, \model{} then generates a corresponding scene for solving the task by populating the environment with a number of relevant and necessary objects (\textit{assets}). As shown in Figure~\ref{fig:system} (B), generating a corresponding scene requires obtaining information for 4 different components: a) \textbf{relevant assets} to be used, b) \textbf{asset sizes}, c) \textbf{initial asset configurations} and d) \textbf{initial scene configuration}. We explain details in the following.

\textbf{Relevant assets} In the previous stage of task proposal, we obtained a list of relevant assets that are necessary for performing the proposed task. To further increase the complexity and diversity of the generated scenes while resembling object distributions of real-world scenarios, we query GPT-4 to return a number of additional queries (object names and their descriptions) that are semantically relevant to the task. For example (Figure~\ref{fig:teaser}), for the task ``Open storage, put the toy inside and close it'', the generated scene involves additionally a living room mat, a table-top lamp, a book, and an office chair. These queries (names) of the assets needed for the scene are used to search in existing object mesh databases. 
Specifically, we use Objaverse~\citep{deitke2023Objaverse}, a large-scale dataset containing over 800k object assets (3d meshes, textures, and etc.) as the main database to retrieve the top $k=10$ objects that matches the asset queries.
Due to noises in assets' language annotations and the extreme diversity of objects in Objaverse (e.g. many of the assets are not common household objects), object retrieved this way are potentially not suitable for the proposed task. 
We further use Gemini-Pro \cite{team2023gemini}, a state-of-the-art vision-language models (VLM) to verify the retrieved assets and filter out the undesired ones. 
(See Appendix \ref{app:asset} for more details for the retrieval and verification process.)
In practice, we found objects retrieved this way work well for rigid object manipulation tasks. 
For soft-body manipulation tasks, where a more consistent and controllable target shape for the soft-body under manipulation is desired, and fine-grained details of geometry and texture are secondary, we ask GPT-4 to come up with desired target shapes, and use a text-to-image followed by image-to-mesh generation pipeline to generate the needed mesh. 
We use Midjourney~\citep{midjourney} as our text-to-image generative model, and Zero-1-to-3~\citep{liu2023zero} as our image-to-mesh generative model. See more details of the generation pipeline in Appendix~\ref{app:asset-generation}.


\textbf{Asset size} Assets generated or retreived from Objaverse~\citep{deitke2022objaverse} and PartNetMobility~\citep{xiang2020sapien} are usually not of physically plausible sizes. 
To account for this, we query GPT-4 to generate the sizes of the assets such that: 1) the sizes should match real-world object sizes; 2) the relative sizes between objects allow a plausible solution for solving the task, e.g., for the task of ``putting a book into the drawer'', the size of the drawer should be larger than the book. 

\textbf{Initial asset configuration}
For certain tasks, the articulated object should be initialized with valid states for the robot to learn the skill. For example, for the task of ``close the window'', the window should be initialized in an open state; similarly, for the task of ``opening the door'', the door should be initially closed. Again, we query GPT-4 to set the initial configurations of these articulated objects, specified in joint angles.  

\textbf{Scene configuration}
Spatial configuration specifying the location and relevant poses of each asset in the scene is crucial for both producing plausible environments and allowing valid skill learning. E.g., for the task of ``retrieving a document from the safe'', the document needs to be initialized \texttt{inside} the safe; for the task of ``removing the knife from the chopping board'', the knife needs to be initially placed \texttt{on} the chopping board. \model{} queries GPT-4 to generate the locations for each asset as well as such special spatial relationships with the task description as the input. To avoid collision between objects, \model{} instructs GPT-4 to place objects in a collision-free manner. (See Appendix~\ref{app:collision} for more details.) With the generated scene components and their corresponding configurations, we populate the scene accordingly. See Figure~\ref{fig:teaser} for a collection of example scenes and tasks generated by \model{}. More examples of the generated scenes are available on our project website. 

\subsection{Training Supervision Generation}
To acquire the skill for solving the proposed task, supervisions for skill learning are needed.
To facilitate the learning process, \model{} first queries GPT-4 to plan and decompose the generated task into shorter-horizon sub-tasks.
After the decomposition, \model{} then queries GPT-4 to choose a proper algorithm for solving each sub-task. 
There are three different types of learning algorithms integrated into \model{}: reinforcement learning~\citep{schulman2017proximal, haarnoja2018soft}, gradient-based trajectory optimization~\citep{xian2023fluidlab, xu2023roboninja}, and action primitive with motion planning~\citep{karaman2011sampling}. Each of these is suited for different tasks, e.g., gradient-based trajectory optimization is more suitable for learning fine-grained manipulation tasks involving soft bodies such as shaping a dough into a target shape~\citep{xu2023roboninja, lin2022diffskill}; action primitives coupled with motion planning are more reliable in solving the task such as approaching a target object via a collision-free path; reinforcement learning better suits tasks that are contact rich and involving continuous interaction with other scene components, e.g., legged locomotion, or when the required actions cannot be simply parameterized by discrete end-effector poses, e.g., turning the knob of an oven. We provide examples and let GPT-4 choose which learning algorithm to use conditioned on the generated sub-task. 

We consider several action primitives including grasping, approaching and releasing a target object. Since parallel jaw gripper can be limited when grasping objects with diverse sizes, we consider a robotic manipulator equipped with a suction cup to simplify object grasping. The grasping and approaching primitives are implemented as follows: we first randomly sample a point on the target object or link, compute a gripper pose that aligns with the normal of the sampled point, and then use motion planning to find a collision-free path to reach the target gripper pose. After the pose is reached, we proceed along the normal direction until a contact is made with the target object. For the grasping and approaching primitives, \model{} asks GPT-4 to specify the target object to grasp or approach, conditioned on the sub-task. See Appendix \ref{app:skill} for more implementation details about the action primitives. 


For sub-tasks trained with RL, we prompt  GPT-4 to write corresponding reward functions with three in-context examples. For rigid manipulation and locomotion tasks, the reward functions are based on the low-level states which GPT-4 can query via provided simulator APIs. 
For soft body manipulation tasks, \model{} uses reward functions specified as the earth-mover distance between the particles of current and target shape. We prompt GPT-4 to generate a text description of the target shape, and then use a text-to-3d model~\cite{liu2023zero} to generate the mesh of the target shape using the text description, as described in Section \ref{sec:scenegen}. See Appendix~\ref{app:asset-generation} for more details on this text-to-3d pipeline.

\subsection{Skill Learning}
Once we obtained all the required information for the proposed task, including scene components and configurations, task decompositions, and training supervisions for the decomposed sub-tasks, we are able to construct the scene in simulation for the robot to learn the required skills for completing the task. 
For long-horizon tasks that involve multiple sub-tasks, we adopt a simple scheme of learning each sub-task sequentially: for each sub-task, we run the learning algorithm for $N=8$ times and use the end state with the highest reward as the initial state for the next sub-task. As aforementioned, we use a combination of techniques for skill learning, including reinforcement
learning, gradient-based trajectory optimization, and action primitive with motion planning, selected on the fly conditioned on the task generated. For more details, please refer to Appendix \ref{app:skill}. 

We included the prompts used for all the stages discussed above in Appendix \ref{sec:prmopt} for reference.

\textbf{Discussion on design choices} 
Our framework design prioritizes its foundational structure over specific backend models used in the initial implementation, and our system is agnostic to the backend LLM/VLM/generative model used, ensuring that \model{} can be continuously improved by upgrading the backend modules with newer models once they become available. In addition, while human-designed 3D asset databases currently still present better quality, automated text-to-3D generative pipelines utilize massive 2D image resources available online and holds a better potential in further scaling up. As a result, we intentionally added support for both retrieval-based and generation-based methods for acquiring assets, and anticipate our method evolving towards a fully generative model in the future. 

\section{Experiments}
\model{} is an automated pipeline that can be queried endlessly, and generate a continuous stream of skill demonstrations for diverse tasks. Our experiments aim to answer the following questions: 1) \textbf{Task Diversity}: How diverse are the tasks proposed by \model{} for robotic skill learning? 2) \textbf{Scene Validity}: Does \model{} generate valid simulation environments? 3) \textbf{Training Supervision Validity}: Does \model{} generate valid task decomposition and training supervisions for the task that will induce intended robot skills? 4) \textbf{Skill Learning}: Does integrating different learning algorithms in \model{} improve skill learning performance? 5) \textbf{System}: Can the whole system produce diverse and meaningful robotic skill demonstrations?
\subsection{Experimental Setup}
Our proposed system is generic and agnostic to specific simulation platforms. However, since we consider a wide range of task categories ranging from rigid dynamics to soft body simulation, and also consider skill learning methods such as gradient-based trajectory optimization which necessitates a differentiable simulation platform, we used Genesis for deploying \model{}, a simulation platform for robot learning with diverse materials and fully differentiable\footnote{Genesis is still under development and will be release publicly soon. We build our system on top of an internal version.}. 
For skill learning, we use SAC~\citep{haarnoja2018soft} as the RL algorithm. 
The policy and Q networks are both Multi-layer Perceptrons (MLP) of size $[256, 256, 256]$, trained with a learning rate of $3e-4$. 
For each sub-task, we train with 1M environment steps. We use BIT$^*$~\citep{gammell2015batch} as the motion planning algorithm, and Adam~\cite{kingma2014adam} for gradient-based trajectory optimization for soft body manipulation tasks. More implementation details can be found in Appendix \ref{app:skill}. 

\begin{figure*}[t]
    \centering
    \includegraphics[width=0.85\textwidth]{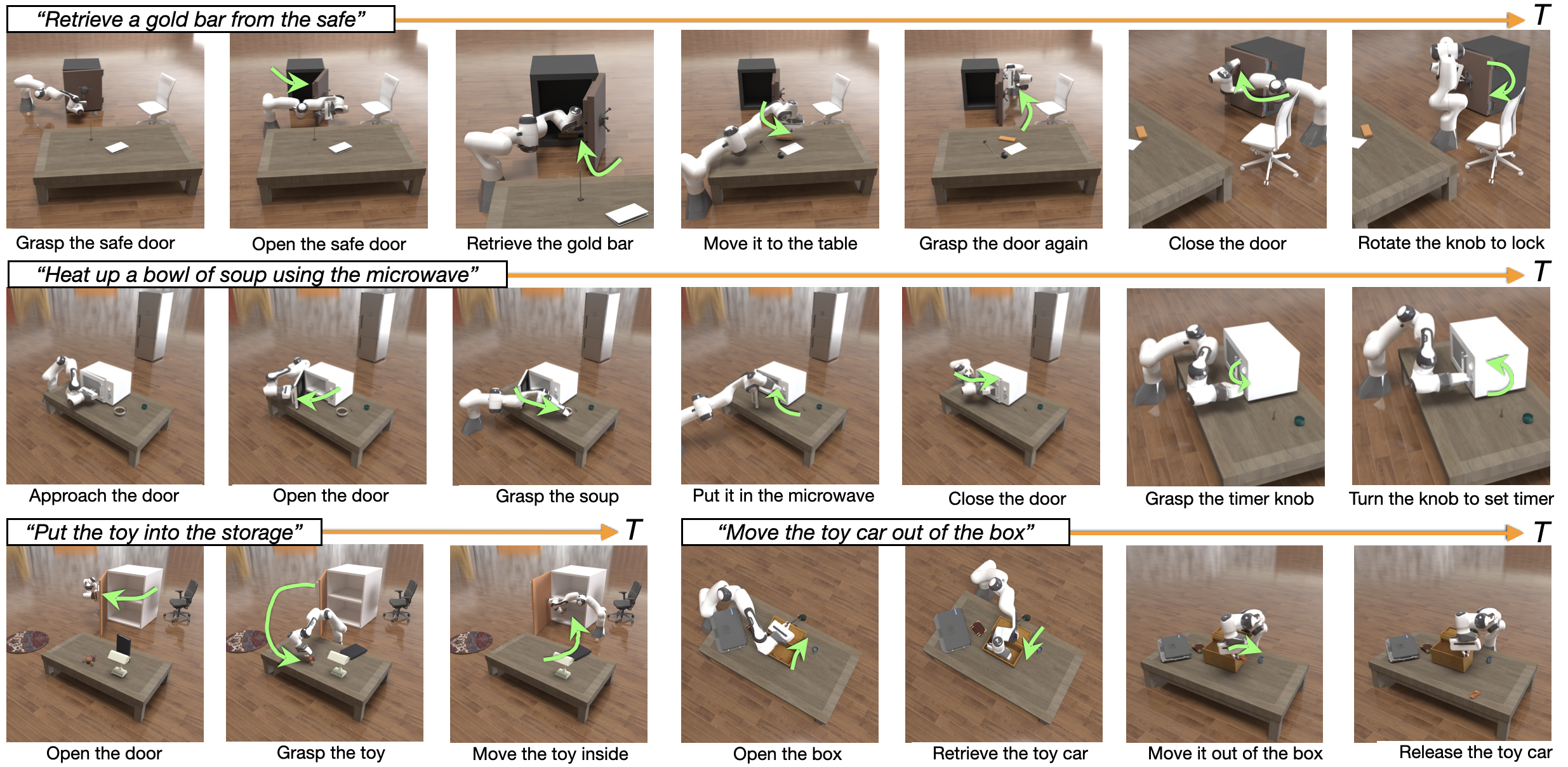}
    \caption{\small Snapshots of the learned skills on 4 example long-horizon tasks. 
    }
    \label{fig:demo}
\end{figure*}

\begin{table*}[h]
    \centering
    \scriptsize
    \begin{tabular}{ccccccc}
      \toprule
      & \model{}  & Behavior-100 & RLbench  & MetaWorld &  Maniskill2 & GenSim \\ \midrule
   Number of Tasks  &  106   & 100  & 106 & 50 & 20 & 70 \\
    Task Description - Self-BLEU $\downarrow$ & $\mathbf{0.284}$ & 0.299 & 0.317 & 0.322 & 0.674 & 	0.378 \\ 
    Task Description - Embedding Similarity (SentenceBert) $\downarrow$ & $\mathbf{0.165}$ & 0.210 & 0.200 & 0.263 & 0.194 & 0.288 \\
    Scene Image - Embedding Similarity (ViT) $\downarrow$ & $\mathbf{0.193}$ & 0.389 & 0.375 & 0.517 & 0.332 & 0.717 \\
    Scene Image - Embedding Similarity (CLIP) $\downarrow$ & $\mathbf{0.762}$ & 0.833 & 0.864 & 0.867 & 0.828 & 0.932\\
    \bottomrule
    \end{tabular}
    \caption{\small\textbf{Comparison on task diversity} with representative human-designed robotics datasets Behavior-100, RLBench, MetaWorld, Maniskill2, and concurrent work GenSim~\cite{wang2023gen}.}
    
    \label{tab:diversity}
\end{table*}
\subsection{Evaluation Metrics and Baselines}
The following evaluation metrics and baselines are used:

\textbf{Task Diversity} The diversity of the generated tasks can be measured in many aspects, such as the semantic meanings of the tasks, scene configurations of the generated simulation environments, the appearances and geometries of the retrieved object assets, and the robot actions required to perform the task. 
For semantic meanings of the tasks, we perform quantitative evaluations by computing the Self-BLEU~\citep{papineni2002bleu,zhu2018texygen} and the embedding similarity~\citep{zhu2018texygen} on the generated task descriptions, where lower scores indicate better diversity. 
In addition to the semantics, we also compare the diversity of the generated tasks in the image space, measured by the embedding similarity of the rendered images of the scenes at the initial state with both ImageNet pre-trained ViT \citep{dosovitskiy2020image} and CLIP models \citep{radford2021learning}.
We compare to established benchmarks, including RLBench~\citep{james2020rlbench}, Maniskill2~\citep{gu2023maniskill2}, Meta-World~\citep{yu2020meta}, and Behavior-100~\citep{srivastava2022behavior}.
We also compare to concurrent work~\cite{wang2023gen}, which leverages LLM to write codes for generating table-top rigid object manipulation tasks.
For robot actions, we evaluate  \model{} qualitatively using the generated environments and visualizations of learned robot skills.  

\textbf{Scene Validity}
To verify that the retrieved objects match the requirements of the task, we compute the BLIP-2 scores~\citep{li2023blip}
between rendered images of the retrieved objects in the simulation scene, and the text descriptions of the objects. 
We compare with two ablations of our system.
A) \textit{w/o object verification: } We retrieve objects based on matching of language descriptions without using a VLM to verify the retrieved object.
B) \textit{w/o size verification: } We use the default size associated with the retrieved asset without quering LLM for plausible sizes. 
We also evaluate scene-level validity via human evaluation, examining whether the generated scenes align with the task descriptions, and if the scene configurations and retrieved objects are correct.

\textbf{Training Supervision Validity} 
We perform human verification by asking a human expert to manually inspect whether the generated decompositions and reward functions are reasonable for solving the task.
We also perform qualitative evaluations by presenting videos of the learned skills using the generated decomposition and training supervisions. 

\textbf{Skill Learning Performance}
We provide quantitative analysis on the skill learning success rate. The success rate is defined as the ratio of runs that successfully learn the skill over all attempting runs for a task.
In addition, we compare to an ablation where we remove the options of using motion planning-based primitive, and rely purely on reinforcement learning to learn the skills on a set of generated articulated-object manipulation tasks. 


\textbf{System} We show qualitative evaluations of the whole system, by providing videos of over 100 learned skills on our website. Figure \ref{fig:teaser} includes snapshots of representative tasks. We also provide a detailed list of generated tasks along with task statistics (e.g., average number of sub-steps) and a detailed failure analysis in Appendix~\ref{app:list-of-generated-tasks} and ~\ref{app:failure}, respectively.

\begin{figure*}[t]
    \centering
    \includegraphics[width=.8\textwidth]{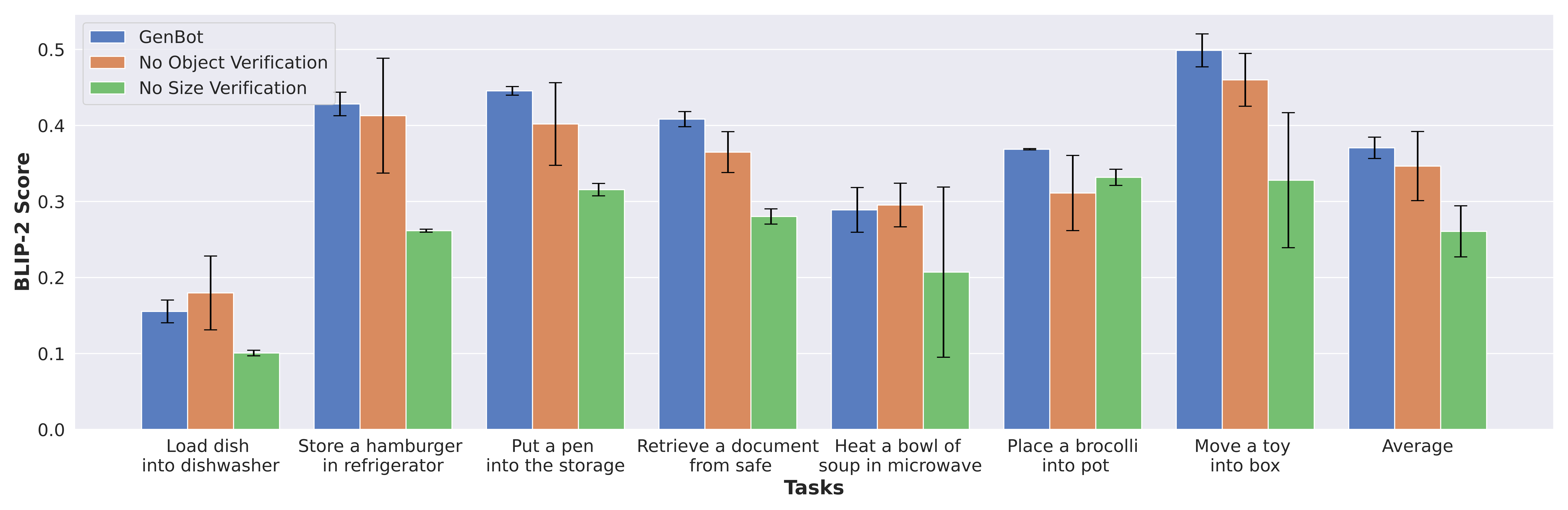}
    \caption{\small We compare the BLIP-2 score of ablations of \model{} on 7 tasks to evaluate the importance of both object and size verification.}
    \label{fig:matching}
\end{figure*}

\begin{figure*}[t]
    \centering
    \includegraphics[width=.8\textwidth]{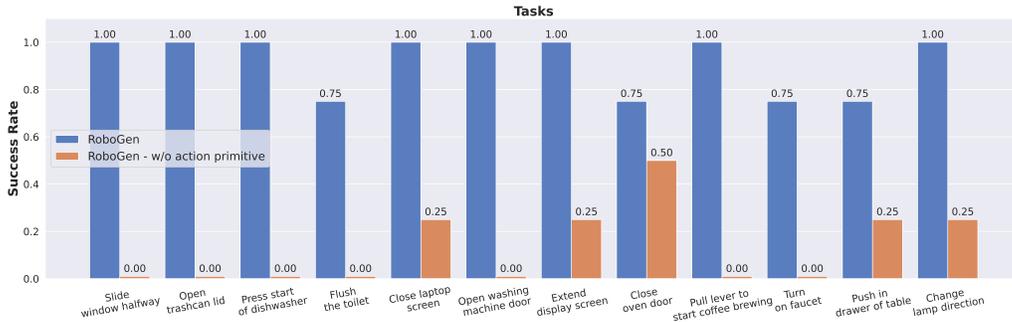}
    \caption{\small Among 12 articulated object manipulation tasks, the success rate decreases drastically if only RL is used for skill learning. }
    \label{fig:skill-learning}
\end{figure*}

\subsection{Results}

\textbf{Task Diversity} We compare \model{} with several established robotics benchmarks in terms of task diversity and report results in Table~\ref{tab:diversity}. Note that \model{} can generate an endless stream of tasks when
queried repeatedly, but here we evaluate a version with 106 tasks generated, comparable to prior works. \model{} achieves the lowest Self-BLEU, as well as the lowest similarity score in both language and image space, 
demonstrating that our pipeline can generate tasks whose semantic and visual diversity matches or surpasses prior manually crafted skill learning benchmarks and datasets. 
We also note the diversity of scenes and tasks generated by RoboGen is noticeably higher than GenSim~\cite{wang2023gen}. 
We believe part of the reason is that GenSim only generates table-top pick-and-place manipulation tasks with a small number of assets from the Ravens benchmarking dataset~\cite{shridhar2022cliport}. In contrast, RoboGen can generate a broader range of tasks such as articulated object manipulation tasks that reason about their affordances and functionalities, legged locomotion, and soft body manipulation tasks, meanwhile leverage more diverse assets retrieved from open-world databases such as Objaverse, resulting in much higher diversity in both task semantics and scene images. 
We also provide the full list of generated tasks, including the task name and task descriptions in Appendix \ref{app:list-of-generated-tasks}, and refer readers to our project website for visualizations of the generated tasks.


\textbf{Scene Validity}
Figure~\ref{fig:matching} shows the BLIP-2 score of all compared methods on an example set of 7 generated tasks. 
As shown, removing the size verification leads to drastic decrease in BLIP-2 score. 
This is expected as the default asset sizes can be drastically different from plausible real-world sizes. 
The ablation ``w/o object verification'' also has a lower BLIP-2 score and a larger variances, indicating our verification step improves validity of the constructed scene. The results demonstrate the importance of using both object and size verification in \model{}.
In addition, we conducted manual evaluations of the generated tasks for scene-level validity. 
Out of 155 generated tasks (full list in Appendix~\ref{app:list-of-generated-tasks}), we found 13 failures due to incorrect scene generation. The failures can be categorized into 1) required functionality not supported by the assets, e.g., loading paper into a printer asset which do not have a movable tray. 
2)  incorrect semantic understanding of articulated object's joint state, i.e., failure to correctly map the joint angle value of an articulated object to its semantic state, e.g., an LLM cannot judge whether the joint angle value 0 corresponds to the door being opened or closed. 
3) failure to find matched assets for tasks that require extremely precise spatial relationships, e.g., it is  hard to retrieve or generate stapler and staples whose size and geometry exactly match each other for the task of loading the staples into the stapler. We provide a detailed analysis in Appendix~\ref{app:failure} on the failure cases and potential solutions to address them in future work. 

\textbf{Training Supervision Validity} 
Figure~\ref{fig:demo} demonstrates the skills learned with the generated training supervisions from \model{}, i.e., the task decompositions and reward functions, on 4 example \textit{long-horizon} tasks. 
As shown, the robot successfully learns skills to complete the corresponding tasks, suggesting that the automatically generated training supervisions are effective in deriving meaningful and useful skills.
We also manually inspected the generated decompositions and reward functions, and found 6 failure cases in the 155 generated object manipulation tasks. The errors can be categorized into 1) referring to undefined variables; 2) reward does not encode the intended behavior. Examples include incorrect semantic understanding of articulated object state, e.g., the task is to fold the chair, yet the generated reward actually encourages unfolding the chair due to misunderstanding of the mapping between joint angle values and object state. We also find it hard to generate correct rewards for continuous motions such as ``moving robotic hand back-and-forth'', or ``knock the door''.  Again, see Appendix~\ref{app:failure} for detailed failure analysis and discussion on potential solutions. 


\textbf{Skill Learning} 
We first evaluate the success rate of our skill learning pipeline on a subset of 50 generated object manipulation tasks, 7 soft-body manipulation tasks, and 12 locomotion tasks. Over all 69 benchmarked tasks, \model{} achieves an average success rate of 0.774, indicating 3 out of 4 runs could lead to successful skill learning. Detailed statistics of the tasks are available in Appendix~\ref{app:skill-learning-success-rate}. 

Further, we compare to an ablated version of \model{} where only RL is used for skill learning. We randomly select 12 tasks that involve interactions with articulated objects for this comparison. The results are shown in Figure~\ref{fig:skill-learning}. As shown, allowing \model{} to select the optimal learning algorithms beneficial for achieving higher performance for completing the tasks. When only RL is used, the skill learning completely fails for most tasks.

\textbf{System} Figure~\ref{fig:teaser} and \ref{fig:demo} show some representative tasks and learned skills generated by \model. As shown in Figure~\ref{fig:teaser}, \model{} can generate diverse tasks for skill learning spanning rigid/articulated object manipulation, legged locomotion and soft body manipulation. Figure~\ref{fig:demo} further shows that \model{} is able to deliver long-horizon manipulation skills with reasonable decompositions. 
For extensive qualitative results of proposed tasks and  learned skills, please refer to our project site. 
Again, please refer to Appendix~\ref{app:tasks} for a list of generated tasks, their statistics, and a detailed failure analysis.


\section{Conclusion \& Limitations}
We introduced \textit{\model}, a generative agent that automatically proposes and learns diverse robotic skills at scale via generative simulation. \model{} utilizes the latest advancements in foundation models to automatically generate diverse tasks, scenes, and training supervisions in simulation, making a foundational step towards scalable robotic skill learning in simulation, while requiring minimal human supervision once deployed. Our system is a fully generative pipeline 
that can be queried endlessly, producing a large number of skill demonstrations associated with diverse tasks and environments. Our current system still has several limitations: 1) Large-scale verification of learned skills is still a challenge in the current pipeline, which could potentially be addressed by incorporating feedback from multi-modal foundation models in the future. 2) Our paradigm is intrinsically constrained by sim-to-real gaps for real-world deployment, which is a stand-alone research field. However, given the recent rapid advancements in physically accurate simulation \citep{li2020incremental} and techniques like domain randomization \citep{tobin2017domain, xu2023roboninja} and realistic sensory signal rendering \citep{zhang2023close}, we anticipate a continual narrowing of this gap in the near future.


\section*{Acknowledgement}

This work is supported by National Science Foundation under Grant No. IIS-2046491, National Science Foundation award No. 1849287, DARPA Machine Common Sense, an Amazon faculty award, an NSF CAREER award, an AFOSR YIP award, and Cisco and Amazon research award. Any opinions, findings, and conclusions or recommendations expressed in this material are those of the author(s) and do not necessarily reflect the views of the National Science Foundation, DARPA, Amazon, AFOSR, or Cisco.

\section*{Impact Statement}
This paper presents work whose goal is to advance the field of Machine Learning. There are many potential societal consequences of our work, none of which we feel must be specifically highlighted here.


\bibliography{references}
\bibliographystyle{icml2024}

\newpage
\appendix
\onecolumn

\section{Implementation Details}
\label{app:implementation}
\subsection{Asset Retrieval and Verification}
\label{app:asset}
For each object in Objaverse, we obtain a list of language descriptions of it by combining the default annotations and a more cleaned version of annotations from~\citep{luo2023scalable}.  
Given the language description of the asset we want to retrieve, we use Sentence-Bert~\citep{reimers2019sentence} to get the embedding of the description, and retrieve $k$ objects from Objaverse whose language embeddings are the most similar to the language embedding of the target asset. Due to noises in the object annotations, there can be significant discrepancies between the actual asset and the intended target, even when the similarity score in the language embedding space is high. To resolve this, we further use Gemini-Pro \cite{team2023gemini} a state-of-the-art vision-language model (VLM) to verify the retrieved assets and filter out the undesired ones. Specifically, we input an image of the retrieved object to the VLM mode to generate a caption of the object. The caption, together with the description of the desired asset and the description of the task, are fed back into GPT-4 to verify if the retrieved asset is appropriate to be used in the proposed task. 

\subsection{Collision Resolving in Scene Generation} 
\label{app:collision}
When the LLM generate the initial pose of the objects, we prompt it to leverage its basic spatial understanding and tries to place the objects in different locations. We use this as the initialization, and check potential collisions in the initial scene configuration. For any detected collision between two objects, we identify the collision node of the objects in contact, and push their center of mass away along the opposite directions of the collision normals to resolve collision.

\subsection{Skill Learning}
\label{app:skill}
For reinforcement learning, we use SAC~\citep{haarnoja2018soft} as the RL algorithm. For object manipulation tasks, the observation space is the low-level state of the objects and robot in the task. The policy and Q networks used in SAC are both Multi-layer Perceptrons (MLP) of size $[256, 256, 256]$. We use a learning rate of $3e-4$ for the actor, the critic, and the entropy regularizer. The horizon of all manipulation tasks are 100, with a frameskip of 2. The action of the RL policy is 6d: where the first 3 elements determines the translation, either as delta translation or target location (suggested by GPT-4), and the second 3 elements determines the delta rotation, expressed as delta-axis angle in the gripper's local frame.
For each sub-task, we train with 1M environment steps. For locomotion tasks, the cross entropy method (CEM (De Boer et al., 2005)) is used for skill learning, which we find to be more stable and efficient than RL. The ground-truth simulator is used as the dynamcis model in CEM, and the actions to be optimized are the joint angle values of the robot. The horizon for all locomotion tasks are 150, with a frameskip of 4

For action primitives, we use BIT$^*$~\citep{gammell2015batch} implemented in the Open Motion Planning Library (OMPL)~\cite{sucan2012open} as the motion planning algorithm. 
For the grasping and the approaching primitive, we first sample a surface point on the target object or link, then compute a gripper pose that aligns the gripper y axis with the normal of the sampled point.
The pre-contact gripper pose is set to be 0.03m above the surface point along the normal direction. 
Motion planning is then used  to find a collision-free path to reach the target gripper pose. 
After the target gripper pose is reached, we keep moving the gripper along the  normal until contact is made. 

For soft body manipulation tasks, we use Adam~\cite{kingma2014adam} for gradient-based trajectory optimization. We run trajectory optimization for 300 gradient steps. We use a learning rate of 0.05 for the optimizer. The horizons of all manipulation tasks are either 150 or 200. We use Earth Mover's distance between object's current and target shape as the cost function for trajectory optimization.

For querying GPT-4, we used a temperature between $0.8 - 1.0$ for task proposal to ensure diversity in the generated tasks. For all other stages of \model{}, we use temperature values between $0-0.3$ to ensure more robust responses from GPT-4.

\section{Generated tasks, Statistics, and Analysis}
\label{app:tasks}
\subsection{List of Tasks and Statistics}
\label{app:list-of-generated-tasks}
Note that \model{} can be used to generate different type of tasks including rigid and articulated object manipulation, soft body object manipulation, and legged locomotion, but the major diversity of the tasks lies in manipulating articulated and rigid objects in the current framework, due to the varied nature of these objects in everyday life.

\begin{figure}
    \centering
    \includegraphics[width=0.32\textwidth]{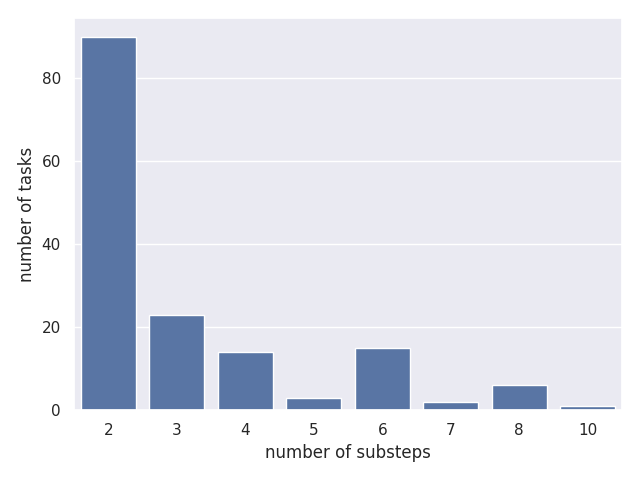}
    \includegraphics[width=0.32\textwidth]{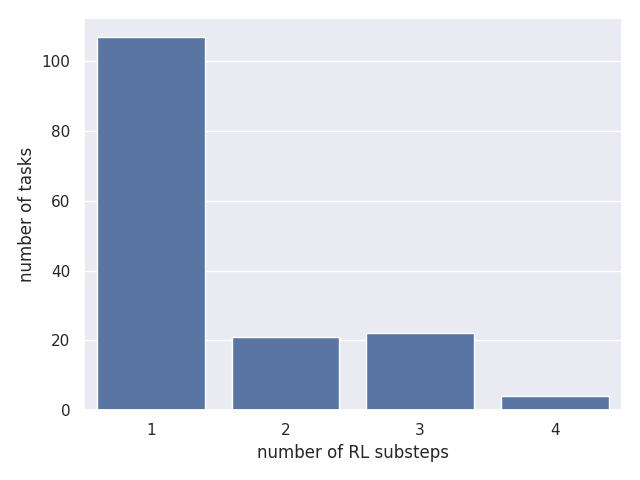}
    \includegraphics[width=0.32\textwidth]{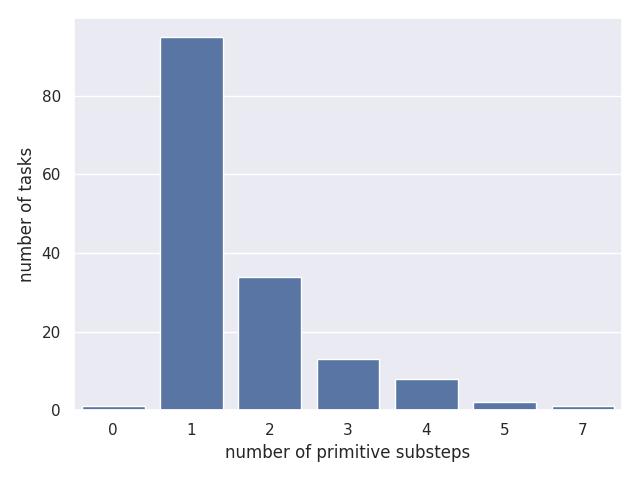}
    \caption{Left: The distribution of number of substeps for the generated rigid and articulated object manipulation tasks in Table~\ref{tab:full_task_list}. The average number of substeps is 3.13. Middle: The distribution of number of substeps that need to be solved using RL for the generated tasks. The average number of RL substeps is 1.5. Right: The distribution of number of substeps that need to be solved using motion planning based primitives for the generated tasks. The average number of such kind of substeps is 1.63. Regarding duration for solving the task: if the task's subgoals can all be solved via planning, typically each task can be solved within 10 minutes. If certain subgoals require RL to solve, it usually takes around 2-3 hours for each RL-necessary step, and the total duration thus depends on both the number and nature of the subtasks. Taking these into account, a task typically takes 4-5 hours on average.  This is done using 8 threads of a CPU running at 2.5Ghz, meaning that each single node in a cluster with a 32-core (64 threads) CPU could run 8 jobs in parallel at the same time.  }
    \label{fig:substep_statistics}
\end{figure}

Table~\ref{tab:full_task_list} provides the list of rigid and articulated object manipulation tasks that are generated using \model{} at the time of submission. We note that \model{} can be constantly queried to generate more tasks. 
Figure~\ref{fig:substep_statistics} shows the distribution of number of substeps for these generated tasks. As shown, most tasks are short-horizon and can be solved within 4 substeps. Longer-horizon tasks require 8 and up to 10 substeps to solve. The average number of substeps for all tasks is 3.13. Figure~\ref{fig:substep_statistics} also presents the distribution of substeps to be solved using RL or motion planning based primitives. Please refer to the caption of the figure for more details. 

Table~\ref{tab:softbody} shows a list of representative soft body manipulation tasks that \model{} generates, and Table~\ref{tab:locomotion} shows the a list of example generated locomotion tasks.

\scriptsize
\begin{longtable}{p{3cm}|p{4.5cm}|p{1.5cm}|p{1.5cm}|p{1.5cm}}
\caption{List of generated tasks. } \label{tab:full_task_list} \\
\toprule
Task name & Task description & \# of substeps & \# of RL substeps & \# of primitive substeps \\
\midrule
\endfirsthead

\multicolumn{5}{c}
{{\bfseries Table \thetable\ continued from previous page}} \\
\toprule
Task name & Task description & \# of substeps & \# of RL substeps & \# of primitive substeps \\
\endhead

\hline
\multicolumn{5}{c}{{Continued on next page}}  
\endfoot

\hline
\endlastfoot


        Rotate Laptop Screen& The robot arm rotates the laptop screen to a certain angle for better view& 2& 1& 1\\ \hline
Move Laptop& The robot arm lifts and moves the laptop to a new location& 3& 2& 1\\ \hline
Close Laptop Lid& The robotic arm will close the laptop lid& 2& 1& 1\\ \hline
Open Laptop Lid& The robotic arm will open the laptop lid& 2& 1& 1\\ \hline
Pack Item In Suitcase& The robot arm places an item .for example, a folded shirt. inside the suitcase& 4& 2& 2\\ \hline
Extend Suitcase Handle& The robotic arm will extend the suitcases handle in order to pull or push the suitcase& 2& 1& 1\\ \hline
Pull Suitcase on Wheels& The robot arm extends the suitcase handle, grips it in a way to let the suitcase stand on its wheels and pulls it& 3& 2& 1\\ \hline
Lift Suitcase& The robotic arm will lift the suitcase by its handle& 2& 1& 1\\ \hline
Partially Close Window& The robotic arm partially closes one of the slider translation windows& 2& 1& 1\\ \hline
Open Window Halfway& The robotic arm will open one of the slider translation windows halfway to let fresh air in& 2& 1& 1\\ \hline
Fully Open Window& The robotic arm will open both of the slider translation windows to their full extent for maximum ventilation& 4& 2& 2\\ \hline
Close Window& The robotic arm closes both slider translation windows& 4& 2& 2\\ \hline
Open and Close Toilet Lid& The robot arm will interact with the hinge lid of the toilet to first open it and then close it& 4& 2& 2\\ \hline
Open and Close Toilet Pump Lid& The robot arm will interact with the slider pump lid to first open it and then close it& 3& 2& 1\\ \hline
Flush the Toilet& The robotic arm will interact with the hinge lever of the toilet to flush it& 3& 1& 2\\ \hline
Set Clock Time& The robotic arm adjusts the hinge hands of the clock to set the desired time& 6& 2& 4\\ \hline
Move Clock Ahead for Daylight Saving& The robotic arm moves the clock hands ahead by 1 hour to adjust for daylight saving& 2& 1& 1\\ \hline
Move Clock Back at End of Daylight Saving& The robot arm moves the clock hands back by 1 hour to adjust to the end of daylight saving& 2& 1& 1\\ \hline
close the oven door& The robot arm needs to close the oven door after use This task involves moving towards the oven door and applying force to close it& 2& 1& 1\\ \hline
Extend Display Screen& The robotic arm will extend the slider translation screen to enlarge the display& 2& 1& 1\\ \hline
Retract Display Screen& The robotic arm will retract the slider translation screen to make the display smaller& 2& 1& 1\\ \hline
Adjust Display Angle& The robotic arm adjusts the display base link to change the viewing angle& 2& 1& 1\\ \hline
Rotate Display Base& The robotic arm will rotate the display base to point the display to a different direction& 2& 1& 1\\ \hline
Rinse a Plate& The robot arm holds a plate under the spout, turns on the faucet to rinse the plate, then turns off the faucet& 8& 3& 5\\ \hline
Turn On Faucet& The robotic arm operates the hinge switch of the faucet in order for water to flow from the spout& 2& 1& 1\\ \hline
Wash Hands& The robot arm acts as if its washing hands to demonstrate good hygiene& 8& 4& 4\\ \hline
Fill a Glass of Water& The robot arm first turns on the faucet, waits for a glass to fill, then turns off the faucet& 5& 3& 2\\ \hline
Fold Chair& The robotic arm will fold the chair to save room or for easy carrying& 3& 1& 2\\ \hline
Position Chair for Seating& The robotic arm positions the unfolded chair in a desired location for a person to sit& 3& 1& 2\\ \hline
Unfold Chair& The robotic arm will unfold the folding chair to make it suitable for sitting& 3& 1& 2\\ \hline
Lift Chair& The robotic arm lifts the chair from the ground to place it into another location& 4& 2& 2\\ \hline
Staple Papers& The robot arm gathers a few loose sheets of paper and uses the stapler to staple them together& 6& 2& 4\\ \hline
Close Stapler Lid& The robot arm closes the lid of the stapler after it has been opened& 2& 1& 1\\ \hline
Open Stapler Lid& The robotic arm will open the lid of the stapler& 2& 1& 1\\ \hline
Load Staples into Stapler& The robot arm inserts new staples into the stapler& 6& 3& 3\\ \hline
Turn On the Printer& The robot arm pushes the slider button to turn on the printer& 2& 1& 1\\ \hline
Load Paper into Printer& The robot arm loads paper into the printer via the input tray, typically located on the printer body& 2& 1& 1\\ \hline
Print a Document& The robot interacts with the printer to print a document The robot arm first places a document on the printer, then moves the button to initiate the print& 4& 2& 2\\ \hline
Stop a Printer& The robot arm stops a printer by moving the slider button to the stop position& 2& 1& 1\\ \hline
Fill Kettle with Water& The robot arm opens the kettle lid, holds a water jug to fill the kettle with water, and then closes the lid& 6& 3& 3\\ \hline
Pour Water from Kettle& The robot arm holds the kettle handle, tilts the kettle to pour water into a cup& 4& 2& 2\\ \hline
Open Kettle Lid& The robotic arm will open the kettle lid& 2& 1& 1\\ \hline
Lift Kettle by Handle& The robotic arm will lift the kettle by its handle& 2& 1& 1\\ \hline
close the drawer of the table& The robot arm will close the drawer of the table& 2& 1& 1\\ \hline
Close Door& The robotic arm will close the door& 2& 1& 1\\ \hline
Knock On Door& The robotic arm will knock on the door in a typical way a human would& 3& 3& 0\\ \hline
Partially Open Door& Open the door partially for ventilation or for casual conversation without fully opening it& 2& 1& 1\\ \hline
Open Door& The robotic arm will open the door& 2& 1& 1\\ \hline
Open Partial Box Lid& The robotic arm will partially open the box lid based on certain degree, to demonstrate kinematic control& 2& 1& 1\\ \hline
Store an Object Inside Box& The robot arm places a small object inside the box and closes the lid& 6& 3& 3\\ \hline
Open Box Lid& The robotic arm will open the box lid& 2& 1& 1\\ \hline
Retrieve an Object From Box& The robot arm opens the box lid, takes a small object from the box, and then closes the lid& 6& 3& 3\\ \hline
Push Drawer In& After retrieving an item from the drawer, the robot arm slides the drawer back into the box& 2& 1& 1\\ \hline
Close Box Lid& The robotic arm closes the lid of the box& 2& 1& 1\\ \hline
Pull Drawer Out& The robotic arm uses the prismatic joint to slide the drawer out from the box& 2& 1& 1\\ \hline
Making Coffee& The robot arm opens the lid of the container, places coffee grounds inside, then closes the lid and starts the brewing process by adjusting the knob& 8& 4& 4\\ \hline
Turning On Coffee Machine& The robotic arm will adjust the hinge knob on the coffee machine to the on setting& 2& 1& 1\\ \hline
Change Cleaning Cycle& Robot changes the cleaning cycle of the dishwasher by interacting with one of the slider buttons& 2& 1& 1\\ \hline
Open Dishwasher Door& The robotic arm will open the dishwasher door& 2& 1& 1\\ \hline
Load Dishwasher& Robot arm places a plate inside the dishwasher& 6& 3& 3\\ \hline
Press Start Button& The robot will press the start button on the dishwasher to begin the washing cycle& 3& 1& 2\\ \hline
Close Dispenser Lid& After filling or extracting contents, the robotic arm will close the lid of the dispenser& 2& 1& 1\\ \hline
Extract Contents& The robot arm will open the dispenser lid and proceed to extract the contents inside the dispenser& 6& 3& 3\\ \hline
Open Dispenser Lid& The robotic arm will open the lid of the dispenser& 2& 1& 1\\ \hline
Fill Dispenser& The robotic arm opens the dispenser lid and then pours the desired content into the dispenser& 5& 3& 2\\ \hline
Rotate Fan Rotor& The robotic arm will apply a force to the rotor of the fan, causing it to rotate& 3& 1& 2\\ \hline
Change Fan Direction& The robotic arm will change the direction of the fan by physically moving the entire fan& 2& 1& 1\\ \hline
Position Fan To Cool Off a Room& The robot arm moves the fan to a location in order to cool off a specific area in a room& 2& 1& 1\\ \hline
Turn Off Water Faucet& The robotic arm will rotate the switch of the faucet to cut off the water supply& 2& 1& 1\\ \hline
Angle Laptop Screen& The robot positions the laptop screen to a desired angle for better visibility& 2& 1& 1\\ \hline
Opening Refrigerator Door& The robotic arm will open one of the refrigerator doors& 2& 1& 1\\ \hline
Opening Both Refrigerator Doors& The robotic arm opens both the refrigerator doors one after the other& 4& 2& 2\\ \hline
Load item into the refrigerator& The robotic arm will open one of the refrigerator doors, place an item inside, and close the door& 6& 3& 3\\ \hline
Retrieving an item from the refrigerator& The robotic arm will open one of the refrigerator doors, retrieve an item, and then close the door& 6& 3& 3\\ \hline
Dispose Toilet Paper into Toilet& A robotic arm picks up a piece of toilet paper and disposes of it in the toilet by dropping it in and then closing the lid& 10& 3& 7\\ \hline
Close Trashcan Lid& The robotic arm will close the trashcans lid& 2& 1& 1\\ \hline
Open Trashcan Lid& The robotic arm will open the trashcans lid& 2& 1& 1\\ \hline
Move the Trashcan& The robot arm pushes the trashcan from one place to another& 2& 1& 1\\ \hline
Change Lamp Direction& The robotic arm will alter the lamp's light direction by manipulating the lamps head& 2& 1& 1\\ \hline
Rotate Lamp Base& The robot arm will rotate the lamp base to adjust the lamps general orientation& 2& 1& 1\\ \hline
Adjust Lamp Position& The robotic arm will adjust the position of the lamp using its hinge rotation bars, enabling the robot to direct the lamps light to a specific area& 6& 3& 3\\ \hline
Change Lamp Direction& The robotic arm will alter the lamp's light direction by manipulating the lamps head& 2& 1& 1\\ \hline
Close Drawer& The robotic arm will push the drawer closed& 2& 1& 1\\ \hline
Retrieve Object from Drawer& The robot arm opens the drawer, retrieves an object from inside, and then closes the drawer& 6& 3& 3\\ \hline
Open Drawer& The robotic arm will pull the drawer open& 2& 1& 1\\ \hline
Store Object in Table Drawer& The robot arm puts an item, like a book, into a drawer in the table& 6& 3& 3\\ \hline
Throw Trash Away& The robotic arm places an item of trash inside the trash can& 7& 3& 4\\ \hline
Insert New Trash Bag& The robotic arm inserts a new trash bag into the trash can& 5& 3& 2\\ \hline
Check Contents of the Pot& The robot arm slides the lid of the pot to check the contents inside the pot& 3& 1& 2\\ \hline
Stir Contents in Pot& The robot arm removes the lid of the pot and stirs the pots contents with a stirring spoon& 4& 2& 2\\ \hline
Remove Pot Lid& The robotic arm will slide the lid of the pot aside& 3& 1& 2\\ \hline
Select Washing Cycle& The robotic arm will push one of the washing machines slider buttons to select a washing cycle& 2& 1& 1\\ \hline
Load Clothes Into Washing Machine& The robot arm opens the washing machine door and places clothes inside& 4& 2& 2\\ \hline
Adjust Washing Settings& The robot arm rotates a knob to adjust washing settings such as temperature or spin speed& 2& 1& 1\\ \hline
Open Washing Machine Door& The robotic arm will open the washing machine door& 2& 1& 1\\ \hline
Move Door Slightly Open& The robotic arm opens the door slightly to allow for some air circulation without fully opening it& 3& 1& 2\\ \hline
Deliver an Object& The robot arm holds an object, opens the door, passes through, then closes the door behind it This represents the robot arm delivering an object from one room to another& 8& 3& 5\\ \hline
Find Door Position& The robot arm would touch different parts of the door to find its initial position It is useful to know the initial position for actions like opening or closing& 4& 2& 2\\ \hline
Regulate Coffee Strength& The robot arm rotates a knob to adjust the strength of the coffee& 2& 1& 1\\ \hline
Insert Portafilter& The robot arm inserts the portafilter into the coffee machine& 3& 2& 1\\ \hline
Adjust Machine Settings& The robot arm adjusts a knob to alter machine settings& 2& 1& 1\\ \hline
Pull Lever to Start Coffee Brewing& The robot arm pulls a lever to start the brewing process of the coffee machine& 2& 1& 1\\ \hline
Steam Milk& The robot operates a lever to steam milk for the coffee& 3& 2& 1\\ \hline
Unload Dishes from Dishwasher& The robot arm retrieves clean dishes from the dishwasher& 6& 3& 3\\ \hline
Start Dishwasher Cycle& The robot arm turns the dishwasher knob to start the washing cycle& 2& 1& 1\\ \hline
Open Dishwasher Door& The robotic arm will open the dishwasher door for placing or removing dishes& 2& 1& 1\\ \hline
Straighten Display Screen& The robotic arm will straighten the display screen if it has been tilted or rotated& 3& 1& 2\\ \hline
Tilt Display Screen& The robotic arm will tilt the display screen to adjust viewing angle& 3& 1& 2\\ \hline
Position Display Screen& The robotic arm will move the display screen to a desired location& 2& 1& 1\\ \hline
Orient Globe Towards Specific Country& The robot arm rotates the globe such that a specific country on the globes surface faces the viewer& 2& 1& 1\\ \hline
Rotate Globe Horizontally& The robotic arm will rotate the globe horizontally to display various continents and countries on its surface& 2& 1& 1\\ \hline
Spin Globe Gently for Leisure& The robot arm spins the globe gently, as a relaxing activity or a playful interaction& 2& 1& 1\\ \hline
Adjust Lamp Height& The robot arm will adjust the height of the lamp by manipulating the rotation bars& 6& 3& 3\\ \hline
Turn On Lamp& The robotic arm turns on the lamp by pressing the toggle button& 3& 1& 2\\ \hline
Set Soup Bowl in Microwave& The robot arm will set a bowl of soup on the microwaves rotation tray and set the timer& 7& 3& 4\\ \hline
Rotate Power Knob& The robotic arm rotates the power know to set the heating power level& 2& 1& 1\\ \hline
Press Microwave Button& The robot arm slides the microwave button& 3& 1& 2\\ \hline
Set Timer& The robotic arm rotates the timer knob to set the duration for heating& 2& 1& 1\\ \hline
Open Microwave Door& The robotic arm will open the microwave door& 2& 1& 1\\ \hline
Open Oven Door& The robot arm is programmed to open the door of the oven& 2& 1& 1\\ \hline
Adjust Oven Timer& The robot arm is to manipulate one of the ovens hinge knobs to set an appropriate timer& 2& 1& 1\\ \hline
Set Oven Temperature& The robot arm is to adjust another knob to set the appropriate temperature for cooking& 2& 1& 1\\ \hline
Set Oven Function& The robot arm needs to adjust another knob to set the desired oven function – for example, circulating air, grilling or bottom heat& 2& 1& 1\\ \hline
Open Fridges Freezer Door& The robot arm opens the freezer compartment door of the refrigerator& 2& 1& 1\\ \hline
Move Cart Forward& The robotic arm will push the cart forward& 2& 1& 1\\ \hline
Turn Cart& The robotic arm will turn the cart to change its direction& 2& 1& 1\\ \hline
Load Object onto Cart& The robot arm places an object onto the cart& 3& 1& 2\\ \hline
Unload Object from Cart& The robot arm takes an object off from the cart& 3& 1& 2\\ \hline
Adjust Chair Height& The robotic arm will adjust the height of the chair by interacting with the knob& 2& 1& 1\\ \hline
Move Chair& The robot arm will move the chair using the wheels& 2& 1& 1\\ \hline
Rotate Chair& The robot arm rotates the chair to a desired direction& 2& 1& 1\\ \hline
Tilt Chair Seat& The robot arm tilts the chair seat to a desired angle& 2& 1& 1\\ \hline
Open Eyeglasses& The robotic arm will unfold the legs of the eyeglasses& 4& 2& 2\\ \hline
Place Eyeglasses on Table& The robot arm picks up the eyeglasses and places them on a table& 3& 1& 2\\ \hline
Store an Item in Safe& The robot arm opens the safe, places an item inside, and then closes and locks the safe& 8& 4& 4\\ \hline
Turn Safe Knob& The robotic arm will turn one of the safes knobs to unlock it& 2& 1& 1\\ \hline
Retrieve an Item from Safe& The robot arm unlocks the safe, opens the door, retrieves an item from inside, and then closes and locks the safe& 8& 4& 4\\ \hline
Open Safe Door& The robotic arm will open the safe door& 2& 1& 1\\ \hline
Open Trashcan Lid& The robotic arm will open the lid of the trashcan& 2& 1& 1\\ \hline
Open Dispenser Lid& The robotic arm will open the lid of the dispenser& 2& 1& 1\\ \hline
Turn On Water Faucet& The robotic arm will rotate the switch of the faucet to turn on the water& 2& 1& 1\\ \hline
Open Laptop& The robotic arm opens the unfolded state of the laptops screen& 2& 1& 1\\ \hline
Open Toilet Lid& The robotic arm will carefully open the lid of the toilet& 2& 1& 1\\ \hline
Close Dispenser Lid& The robotic arm will close the dispenser lid after use& 2& 1& 1\\ \hline
Close Table Drawer& The robotic arm will close the open drawer on the table& 2& 1& 1\\ \hline
Open Trash Can& The robotic arm will open the trash can lid& 2& 1& 1\\ \hline
Close Toilet Lid& The robotic arm will put down the lid of the toilet& 3& 1& 2\\ \hline
Open Door& The robotic arm will open the door by rotating the hinge& 2& 1& 1\\ \hline
Turn Off Faucet& The robotic arm turns off the faucet by rotating one of the hinge switches& 2& 1& 1\\ \hline
Close Window& The robotic arm will close the window to preserve indoor temperature& 2& 1& 1\\ \hline
Open Box& The robot arm opens the box by manipulating the hinge lid& 2& 1& 1\\ \hline
Rotate Clock Hands& Rotate the minute and hour hands of the clock with the robotic arm, simulating the passing of time& 4& 2& 2\\ \hline
Unfold the Chair& The robotic arm will unfold the chair to prepare it for use& 4& 2& 2\\ \hline
Open Kettle Lid& The robot arm lifts the kettle lid& 2& 1& 1\\ 
\end{longtable}

\begin{table}[]
    \centering
    \scriptsize
    \begin{tabular}{c|c}
    \toprule
    Task Name & Task Description \\ \hline
      Bend the noodle into a U shape & The robot needs to to bend an initial straight noodle into the shape of letter "U" \\ \hline
      Flatten the rice dough &  The robot uses a square dough flattener to flatten a rice dough \\ \hline
      Cut dough in half & The robot uses a knife to cut a dough in half \\ \hline
      Shape dough & The robot uses two square dough flatteners to shape the dough into a baguette\\ \hline
      Lift up dumping & The robot uses two square dough flatteners to lift up a dumpling \\ \hline
      Roll out dough & The robot uses a rolling pin to flatten a dough \\ \hline
      Put filling onto wrapper & The robot needs to grasp a dumpling filling and put it on top of the dumpling wrapper \\ 
    \bottomrule
    \end{tabular}
    \caption{List of soft body manipulation tasks \model{} generated. }
    \label{tab:softbody}
\end{table}


\begin{table}[]
    \centering
    \scriptsize
    \begin{tabular}{c|c}
    \toprule
    Task Name & Task Description \\ \hline
      Jump forward & The legged robot needs to do a jump forward \\ \hline
      Spin counter-clockwise &  The legged robot needs to spin itself counter-clockwise around the vertical axis \\ \hline
      Run forward & The legged robot needs to do fun forward at a high speed \\ \hline
      Spin left without using right hind leg &  The legged robot needs to spin itself to the left while not letting the right hind leg touch the ground \\ \hline
      Jump higher than 5 meters & The legged robot needs to jump and reach a height higher than 5 meters \\ \hline
      Flip forward & The legged  robot makes a flip forward \\ \hline
      Climb up stairs & The legged robot climbs up a staircase in the environment\\ \hline
      Kick the soccer ball to the left & The legged robot needs to kick the soccer ball and make it move to the left\\ \hline
      Walk backwards & The legged robot needs to move backwards \\ \hline 
      Push Ball & The legged robot needs to push the ball forward \\ \hline
      Turn Right & The legged robot needs to turn itself to face right \\ \hline
      Crawl forward & The legged robot needs to move forward while keeping the body in a low position \\ 
    \bottomrule
    \end{tabular}
    \caption{List of locomotion tasks \model{} generated. }
    \label{tab:locomotion}
\end{table}

\normalsize

\subsection{Skill Learning Success Rate}
\label{app:skill-learning-success-rate}
Due to the randomness in the skill learning process (sampling is used in the motion planning-based action primitive, and RL inherently has randomness during exploration and training), we also provide quantitative analysis on the skill learning success rate, i.e., given a generated task with correct training supervisions, if we run the skill learning pipeline for multiple times, how many of the runs would succeed in learning the skill. The success in learning a skill is determined by a human evaluator watching the video of the learned policy. 

We present the detailed skill learning success rate of 50 articulated object manipulation tasks in Table~\ref{tab:success_rate}. The average skill learning success rate among these tasks is 0.745. 
We also benchmark the success rate for the soft-body manipulation and locomotion tasks, shown in Table~\ref{tab:success_rate_softbody} and Table~\ref{tab:success_rate_locomotion}. 

\scriptsize
\begin{longtable}{p{3cm}|p{4.5cm}|p{3cm}}
\caption{Skill learning success rate on 50 articulated object manipulation tasks.} \label{tab:success_rate} \\
\toprule
Task name & Task description & Skill Learning Success Rate \\
\midrule
\endfirsthead

\multicolumn{3}{c}
{{\bfseries Table \thetable\ continued from previous page}} \\
\toprule
Task name & Task description & Skill Learning Success Rate \\
\endhead

\hline
\multicolumn{3}{c}{{Continued on next page}}  
\endfoot

\hline
\endlastfoot

Rotate Laptop Screen & The robot arm rotates the laptop screen to a certain angle for better view & 1.0 \\ \hline
Extend Suitcase Handle & The robotic arm will extend the suitcases handle in order to pull or push the suitcase & 1.0 \\ \hline
Open Window Halfway & The robotic arm will open one of the slider translation windows halfway to let fresh air in & 1.0 \\ \hline
Flush the Toilet & The robotic arm will interact with the hinge lever of the toilet to flush it & 0.67 \\ \hline
Move Clock Ahead for Daylight Saving & The robotic arm moves the clock hands ahead by 1 hour to adjust for daylight saving & 0.38 \\ \hline
close the oven door & The robot arm needs to close the oven door after use This task involves moving towards the oven door and applying force to close it & 0.83 \\ \hline
Open Trashcan Lid & The robotic arm will open the lid of the trashcan & 1.0 \\ \hline
Extend Display Screen & The robotic arm will extend the slider translation screen to enlarge the display & 1.0 \\ \hline
Turn On Faucet & The robotic arm operates the hinge switch of the faucet in order for water to flow from the spout & 0.83 \\ \hline
Unfold Chair & The robotic arm will unfold the folding chair to make it suitable for sitting & 0.5 \\ \hline
Open Stapler Lid & The robotic arm will open the lid of the stapler & 0.5 \\ \hline
Turn On the Printer & The robot arm pushes the slider button to turn on the printer & 1.0 \\ \hline
Lift Kettle by Handle & The robotic arm will lift the kettle by its handle & 0.83 \\ \hline
close the drawer of the table & The robot arm will close the drawer of the table & 0.75 \\ \hline
Close Door & The robotic arm will close the door & 0.5 \\ \hline
Open Partial Box Lid & The robotic arm will partially open the box lid based on certain degree, to demonstrate kinematic control & 0.83 \\ \hline
Pull Drawer Out & The robotic arm uses the prismatic joint to slide the drawer out from the box & 0.67 \\ \hline
Turning On Coffee Machine & The robotic arm will adjust the hinge knob on the coffee machine to the on setting & 0.5 \\ \hline
Press Start Button & The robot will press the start button on the dishwasher to begin the washing cycle & 1.0 \\ \hline
Close Dispenser Lid & After filling or extracting contents, the robotic arm will close the lid of the dispenser & 0.25 \\ \hline
Open Dispenser Lid & The robotic arm will open the lid of the dispenser & 0.0 \\ \hline
Rotate Fan Rotor & The robotic arm will apply a force to the rotor of the fan, causing it to rotate & 1.0 \\ \hline
Turn On Water Faucet & The robotic arm will rotate the switch of the faucet to turn on the water & 1.0 \\ \hline
Open Laptop & The robotic arm opens the unfolded state of the laptops screen & 0.8 \\ \hline
Opening Both Refrigerator Doors & The robotic arm opens both the refrigerator doors one after the other & 0.8 \\ \hline
Open Toilet Lid & The robotic arm will carefully open the lid of the toilet & 1.0 \\ \hline
Close Trashcan Lid & The robotic arm will close the trashcans lid & 0.33 \\ \hline
Change Lamp Direction & The robotic arm will alter the lamp's light direction by manipulating the lamps head & 1.0 \\ \hline
Partially Close Window & The robotic arm partially closes one of the slider translation windows & 0.5 \\ \hline
Close Dispenser Lid & The robotic arm will close the dispenser lid after use & 1.0 \\ \hline
Open Drawer & The robotic arm will pull the drawer open & 0.75 \\ \hline
Close Table Drawer & The robotic arm will close the open drawer on the table & 0.75 \\ \hline
Open Trash Can & The robotic arm will open the trash can lid & 1.0 \\ \hline
Remove Pot Lid & The robotic arm will slide the lid of the pot aside & 1.0 \\ \hline
Close Toilet Lid & The robotic arm will put down the lid of the toilet & 1.0 \\ \hline
Open Washing Machine Door & The robotic arm will open the washing machine door & 1.0 \\ \hline
Move Door Slightly Open & The robotic arm opens the door slightly to allow for some air circulation without fully opening it & 0.67 \\ \hline
Open Door & The robotic arm will open the door by rotating the hinge & 0.5 \\ \hline
Turn Off Faucet & The robotic arm turns off the faucet by rotating one of the hinge switches & 0.67 \\ \hline
Close Window & The robotic arm will close the window to preserve indoor temperature & 0.8 \\ \hline
Open Box & The robot arm opens the box by manipulating the hinge lid & 0.8 \\ \hline
Rotate Clock Hands & Rotate the minute and hour hands of the clock with the robotic arm, simulating the passing of time & 0.4 \\ \hline
Pull Lever to Start Coffee Brewing & The robot arm pulls a lever to start the brewing process of the coffee machine & 1.0 \\ \hline
Open Dishwasher Door & The robotic arm will open the dishwasher door for placing or removing dishes & 0.6 \\ \hline
Tilt Display Screen & The robotic arm will tilt the display screen to adjust viewing angle & 0.6 \\ \hline
Unfold the Chair & The robotic arm will unfold the chair to prepare it for use & 1.0 \\ \hline
Rotate Globe Horizontally & The robotic arm will rotate the globe horizontally to display various continents and countries on its surface & 1.0 \\ \hline
Turn On Lamp & The robotic arm turns on the lamp by pressing the toggle button & 0.25 \\ \hline
Open Kettle Lid & The robot arm lifts the kettle lid & 0.75 \\ \hline
Open Microwave Door & The robotic arm will open the microwave door & 0.25 \\ \hline

\end{longtable}

\scriptsize
\begin{longtable}{p{3cm}|p{4.5cm}|p{3cm}}
\caption{Skill learning success rate on 7 soft-body manipulation tasks.} \label{tab:success_rate_softbody} \\
\toprule
Task name & Task description & Skill Learning Success Rate \\
\midrule
\endfirsthead

\multicolumn{3}{c}
{{\bfseries Table \thetable\ continued from previous page}} \\
\toprule
Task name & Task description & Skill Learning Success Rate \\
\endhead

\hline
\multicolumn{3}{c}{{Continued on next page}}  
\endfoot

\hline
\endlastfoot

 Bend the noodle into a U shape & The robot needs to to bend an initial straight noodle into the shape of letter "U" &  0.8 \\ \hline
      Flatten the rice dough &  The robot uses a square dough flattener to flatten a rice dough & 1.0 \\ \hline
      Cut dough in half & The robot uses a knife to cut a dough in half & 1.0 \\ \hline
      Shape dough & The robot uses two square dough flatteners to shape the dough into a baguette & 0.6 \\ \hline
      Lift up dumping & The robot uses two square dough flatteners to lift up a dumpling & 1.0 \\ \hline
      Roll out dough & The robot uses a rolling pin to flatten a dough & 1.0\\ \hline
      Put filling onto wrapper & The robot needs to grasp a dumpling filling and put it on top of the dumpling wrapper & 0.8 \\ 

\end{longtable}

\scriptsize
\begin{longtable}{p{3cm}|p{4.5cm}|p{3cm}}
\caption{Skill learning success rate on 12 locomotion tasks.} \label{tab:success_rate_locomotion} \\
\toprule
Task name & Task description & Skill Learning Success Rate \\
\midrule
\endfirsthead

\multicolumn{3}{c}
{{\bfseries Table \thetable\ continued from previous page}} \\
\toprule
Task name & Task description & Skill Learning Success Rate \\
\endhead

\hline
\multicolumn{3}{c}{{Continued on next page}}  
\endfoot

\hline
\endlastfoot

Jump forward & The legged robot needs to do a jump forward & 1.0 \\ \hline
Spin counter-clockwise &  The legged robot needs to spin itself counter-clockwise around the vertical axis & 1.0 \\ \hline
Run forward & The legged robot needs to do fun forward at a high speed & 0.6 \\ \hline
Spin left without using right hind leg &  The legged robot needs to spin itself to the left while not letting the right hind leg touch the ground & 0.8 \\ \hline
Jump higher than 5 meters & The legged robot needs to jump and reach a height higher than 5 meters & 1.0 \\ \hline
Flip forward & The legged  robot makes a flip forward & 0.6 \\ \hline
Climb up stairs & The legged robot climbs up a staircase in the environment & 0.4\\ \hline
Kick the soccer ball to the left & The legged robot needs to kick the soccer ball and make it move to the left & 0.8 \\ \hline
Walk backwards & The legged robot needs to move backwards & 1.0 \\ \hline 
Push Ball & The legged robot needs to push the ball forward  & 1.0 \\ \hline
Turn Right & The legged robot needs to turn itself to face right & 1.0 \\ \hline
Crawl forward & The legged robot needs to move forward while keeping the body in a low position & 0.8 \\ 

\end{longtable}

\normalsize
\subsection{Failure Analysis}
\label{app:failure}

Through manual inspection on the 155 generated tasks in Table~\ref{tab:full_task_list}, we found 19 failure cases in total, due to either error in the generated scene or the generated training supervisions. Table~\ref{tab:failure_case} provides a detailed analysis on the failure cases.

Among the 19 failure cases, 13 failures can be attributed to incorrect scene generation. The failures can be categorized into 1) required functionality not supported by the assets, e.g., loading paper into a printer asset which do not have a movable tray. 2)  incorrect semantic understanding of articulated object's joint state, i.e., failure to correctly map the joint angle value of an articulated object to its semantic state, e.g., an LLM cannot judge whether the joint angle value 0 corresponds to the door being opened or closed. 
3) failure to find matched assets for tasks that require extremely precise spatial relationships, e.g., it's hard to retrieve or generate stapler and staples whose size and geometry exactly match each other for the task of loading the staples into the stapler.
Some of these failures can be addressed with additional checks, e.g., using a vision language model to verify the mapping between the joint angle values and the semantic state of the asset, while others (generate assets with required functionalities, or pair of matched assets) might require more fundamental research to address. 

6 failures are caused by incorrect reward generation. The errors can be categorized into 1) referring to undefined variables; 2) reward does not encode the intended behavior. Examples include incorrect semantic understanding of articulated object state, e.g., the task is to fold the chair, yet the generated reward actually encourages unfolding the chair due to misunderstanding of the mapping between joint angle values and object state. We also find it hard to generate correct rewards for continuous motions such as ``moving robotic hand back-and-forth'', or ``knock the door''. 
Again, the incorrect semantic understanding of articulated object state can be potentially fixed by using a vision-language model to figure out the mapping between the joint angle and object state. For syntax errors such as undefined variables, one could feed the error back to the LLM and ask it to correct itself. 
The reward function generation can also be improved to better match the intended goal by incorporating environment feedback into the system~\citep{ma2023eureka}, which we leave as future work.

\scriptsize
\begin{longtable}{p{3cm}|p{4.5cm}|p{6cm}}
\caption{Failure case analysis} \label{tab:failure_case} \\
\toprule
Task name & Task description & Failure case \\
\midrule
\endfirsthead

\multicolumn{3}{c}
{{\bfseries Table \thetable\ continued from previous page}} \\
\toprule
Task name & Task description & Failure case \\
\endhead

\hline
\multicolumn{3}{c}{{Continued on next page}}  
\endfoot

\hline
\endlastfoot

Pack Item In Suitcase & The robot arm places an item .for example, a folded shirt. inside the suitcase & Limited asset functionality: the suitcase cannot be opened.
 \\ \hline
Open Window Halfway & The robotic arm will open one of the slider translation windows halfway to let fresh air in & Incorrect semantic understanding of articulated object state: setting both joint angles to 0 make the window opened already
 \\ \hline
Correct Clock Time & The robotic arm corrects the time displayed on the clock based on the standard time & Generated reward refers to undefined variables "standard time"
 \\ \hline
Wash Hands & The robot arm acts as if its washing hands to demonstrate good hygiene & Reward error in one of the substeps: moving hands back and forth
 \\ \hline
Fold Chair & The robotic arm will fold the chair to save room or for easy carrying & Wrong reward due to incorrect understanding of the joint state of articulated object. The reward actually encourages unfolding the chair
 \\ \hline
Unfold Chair & The robotic arm will unfold the folding chair to make it suitable for sitting & Wrong reward due to incorrect understanding of the joint state of articulated object. The reward actually encourages folding the chair
 \\ \hline
Staple Papers & The robot arm gathers a few loose sheets of paper and uses the stapler to staple them together & Too delicate initial spatial relationship -- the task requires the sheet of paper to be initialized into the stapler, which is hard for a random stapler and a sheet of paper sampled from PartNetMobility / Objaverse
 \\ \hline
Load Staples into Stapler & The robot arm inserts new staples into the stapler & Asset mismatch: randomly sampled stapler and staple won't easily match each other
 \\ \hline
Load Paper into Printer & The robot arm loads paper into the printer via the input tray, typically located on the printer body & Limited asset functionality: the printer cannot really be loaded with paper
 \\ \hline
Print a Document & The robot interacts with the printer to print a document The robot arm first places a document on the printer, then moves the button to initiate the print & Limited asset functionality: the printer cannot really be loaded with paper
 \\ \hline
Fill Kettle with Water & The robot arm opens the kettle lid, holds a water jug to fill the kettle with water, and then closes the lid & Limited asset functionality: the kettle lid cannot be really moved away from the kettle body
 \\ \hline
Pour Water from Kettle & The robot arm holds the kettle handle, tilts the kettle to pour water into a cup & Limited asset functionality: the kettle lid cannot be really moved away from the kettle body
 \\ \hline
Knock On Door & The robotic arm will knock on the door in a typical way a human would & Reward error: not really correct reward function for the knocking motion.
 \\ \hline
Making Coffee & The robot arm opens the lid of the container, places coffee grounds inside, then closes the lid and starts the brewing process by adjusting the knob & Limited asset functionality: the coffeemachine lid cannot really be moved away from the body
 \\ \hline
Extract Contents & The robot arm will open the dispenser lid and proceed to extract the contents inside the dispenser & Limited asset functionality: the lid of the dispenser canont be removed from the body to enable the pouring motion.
 \\ \hline
Fill Dispenser & The robotic arm opens the dispenser lid and then pours the desired content into the dispenser & Limited asset functionality: the lid of the dispenser canont be removed from the body to enable the pouring motion.
 \\ \hline
Stir Contents in Pot & The robot arm removes the lid of the pot and stirs the pots contents with a stirring spoon & Limited asset functionality: the lid cannot really be removed from the pod
 \\ \hline
Deliver an Object & The robot arm holds an object, opens the door, passes through, then closes the door behind it This represents the robot arm delivering an object from one room to another & Reward error for delivering an object through the door
 \\ \hline
Open Eyeglasses & The robotic arm will unfold the legs of the eyeglasses & Incorrect semantic understanding of the object joint state. Setting the joint angle to 0 actually make the eyeglass already unfolded.
 \\ \hline

\end{longtable}

\normalsize
\section{Asset Generation Results}
\label{app:asset-generation}
We provide more details on our text-to-3D asset generation pipeline here. This asset generation pipeline is majorly used for generating goal meshes for deformable object manipulation tasks. It works as follows. First, given the text descriptions of the object, we use Midjourney~\citep{midjourney} to generate a 2D image of it. We prompt Midjourney to generate the image with white background, in either front view or top-down view, as images in these formats are more suitable inputs for the following text-to-3D generation models. 
Midjourney usually generates 4 images in a batch, and a random image is chosen as input for the following image-to-3d model. 
Then, the generated image and text descriptions are used as input to zero-1-to-3~\citep{liu2023zero}, an image to mesh generative model. The generated mesh is then refined using Deep Marching Tetrahedra (DMTet)~\citep{shen2021deep}.
Figure~\ref{tab:asset-generation} shows some example results.

\begin{table}[]
    \centering
    \begin{tabular}{cccc}
       \includegraphics[width=0.23\textwidth]{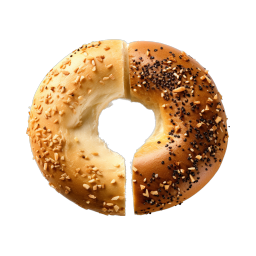}  &
       \includegraphics[width=0.23\textwidth]{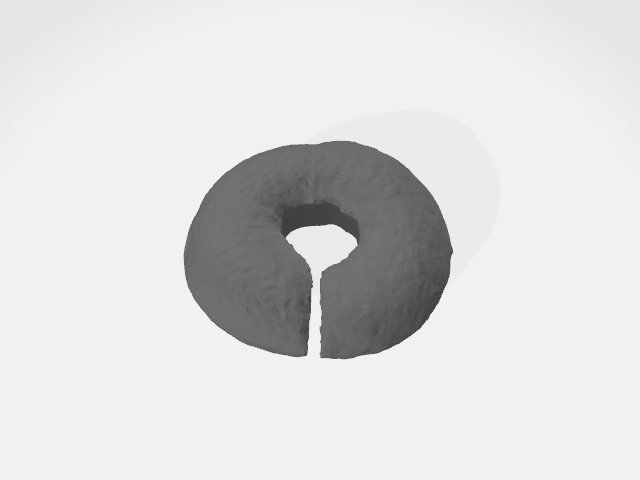} &
     \includegraphics[width=0.23\textwidth]{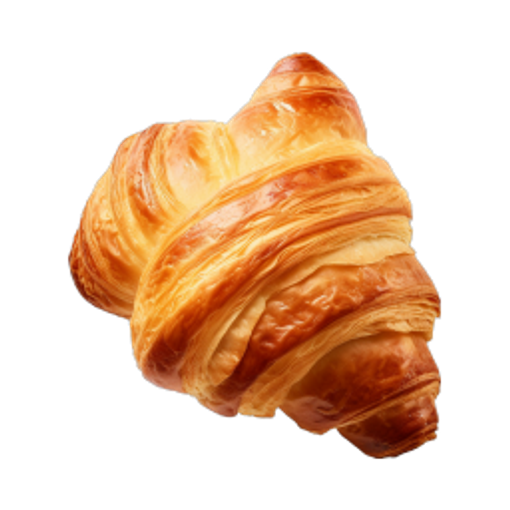}  &  
     \includegraphics[width=0.23\textwidth]{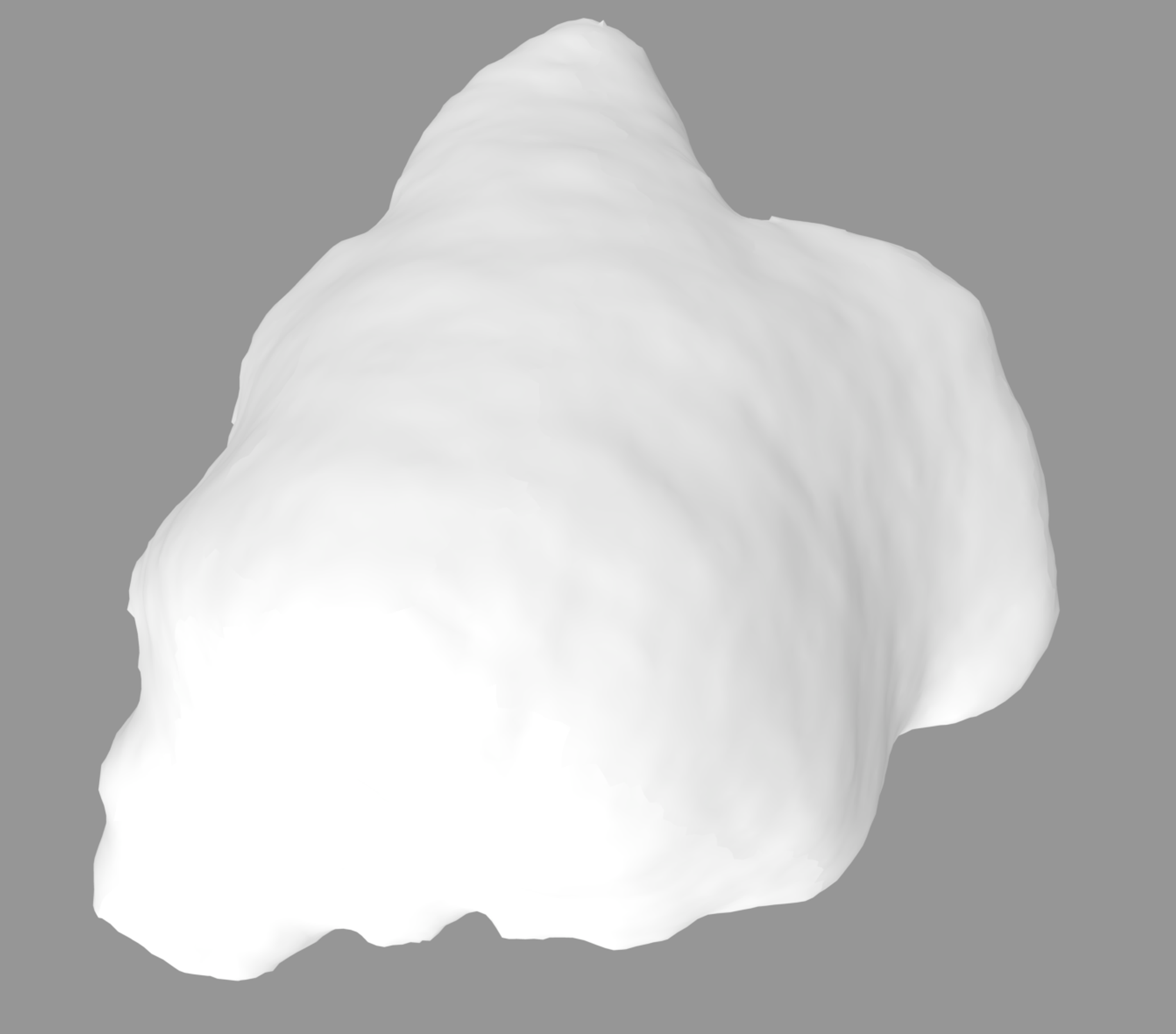}   
       \\
         image of bagel  & mesh of  bagel   &  image of croissant &  mesh of  croissant \\

        \includegraphics[width=0.23\textwidth]{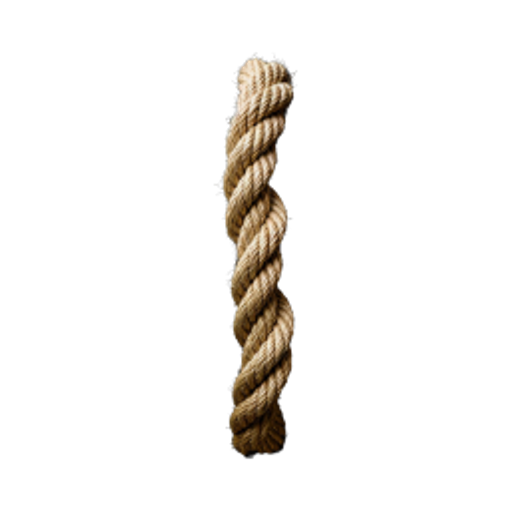}  &
       \includegraphics[width=0.13\textwidth]{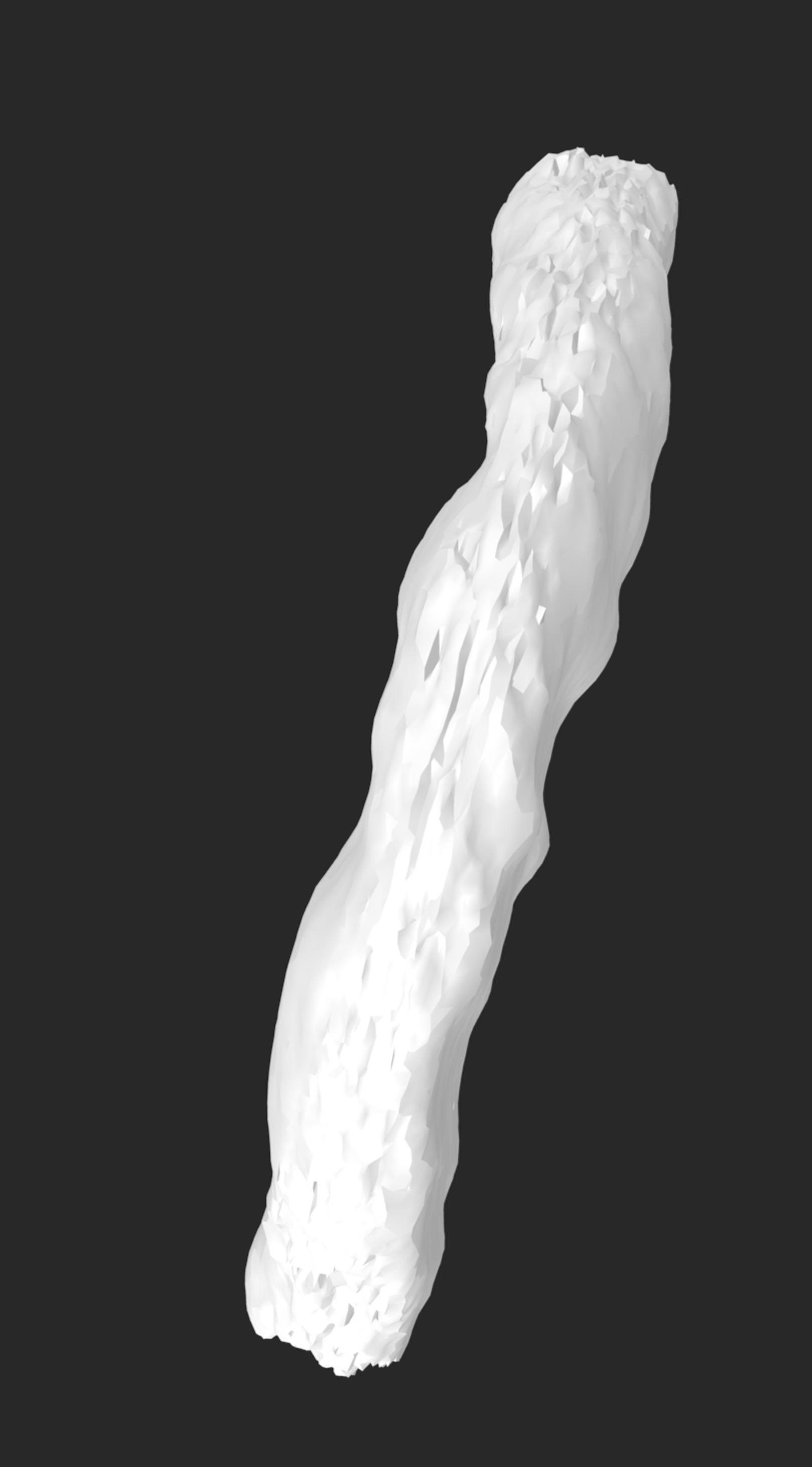} &
     \includegraphics[width=0.23\textwidth]{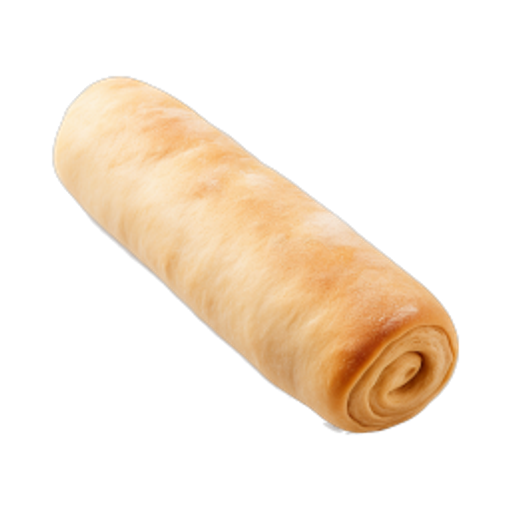}  &  \includegraphics[width=0.23\textwidth]{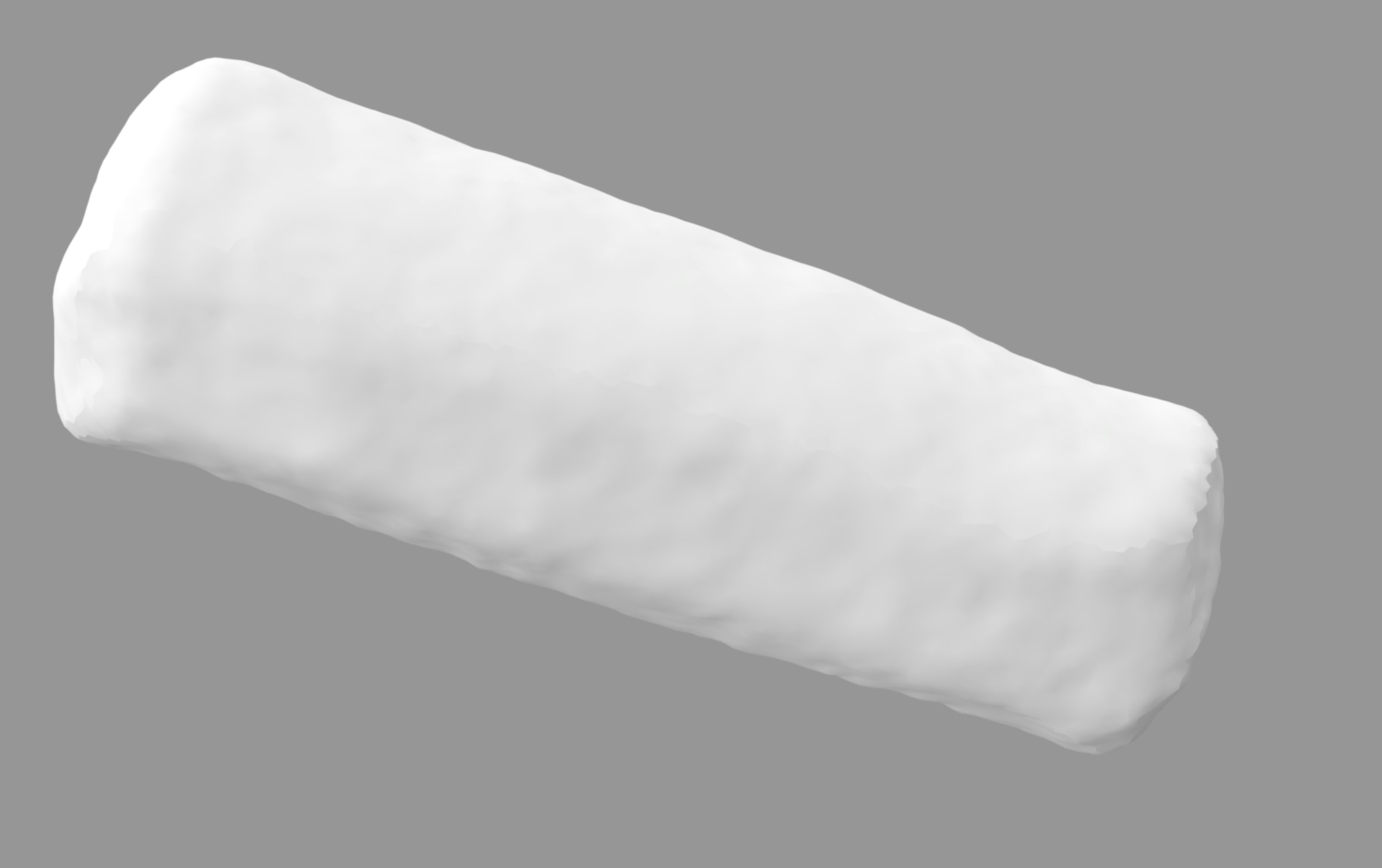}   
       \\
         image of rope  & mesh of  rope   &  image of cylinder dough &  mesh of  cylinder dough \\
    \includegraphics[width=0.23\textwidth]{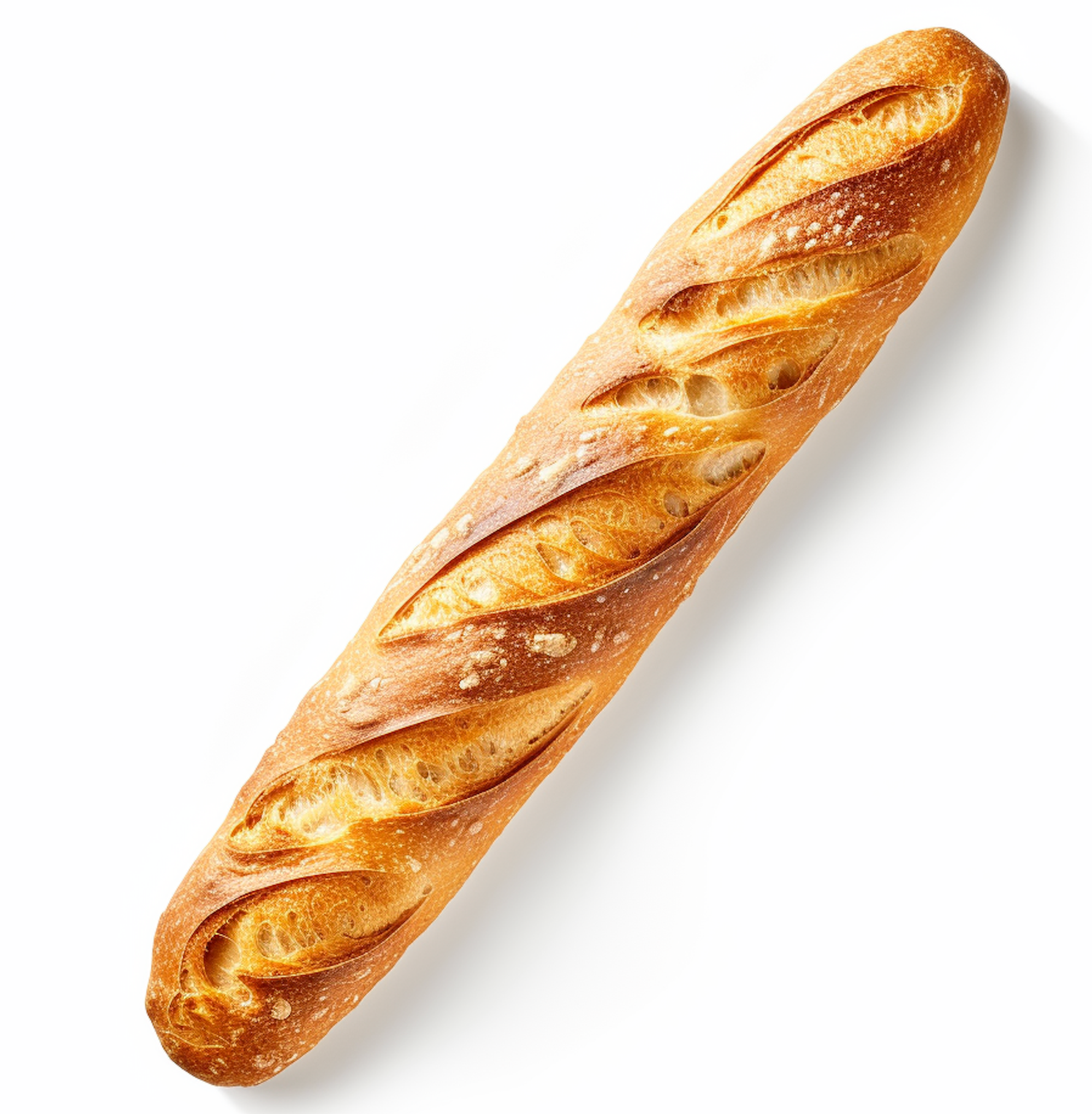}  &
       \includegraphics[width=0.23\textwidth]{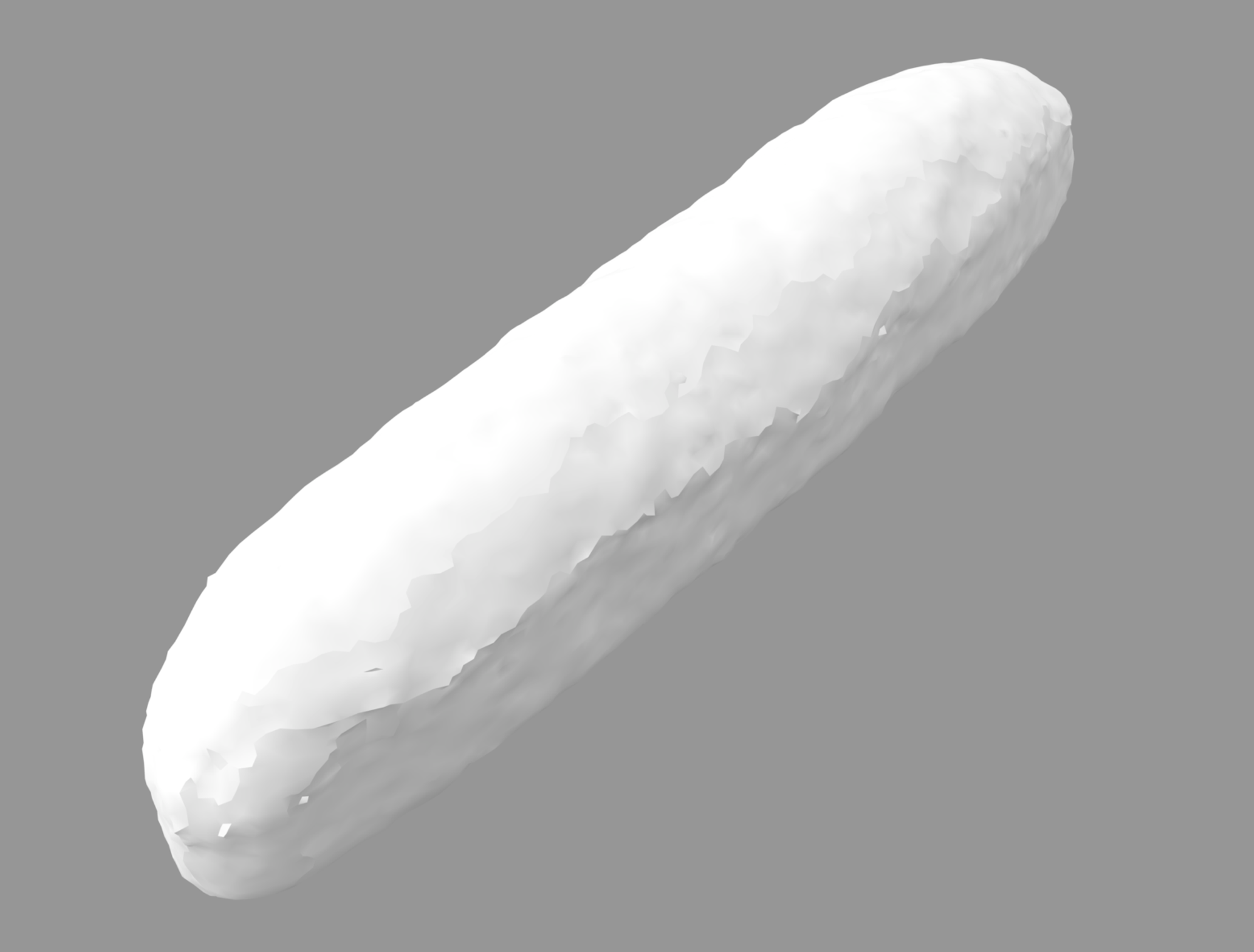} &
     \includegraphics[width=0.23\textwidth]{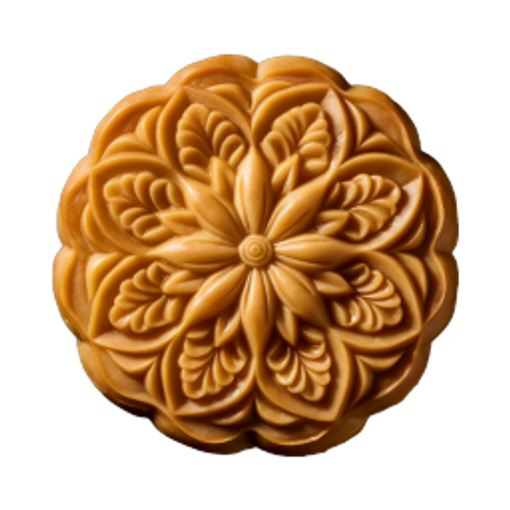}  &  \includegraphics[width=0.23\textwidth]{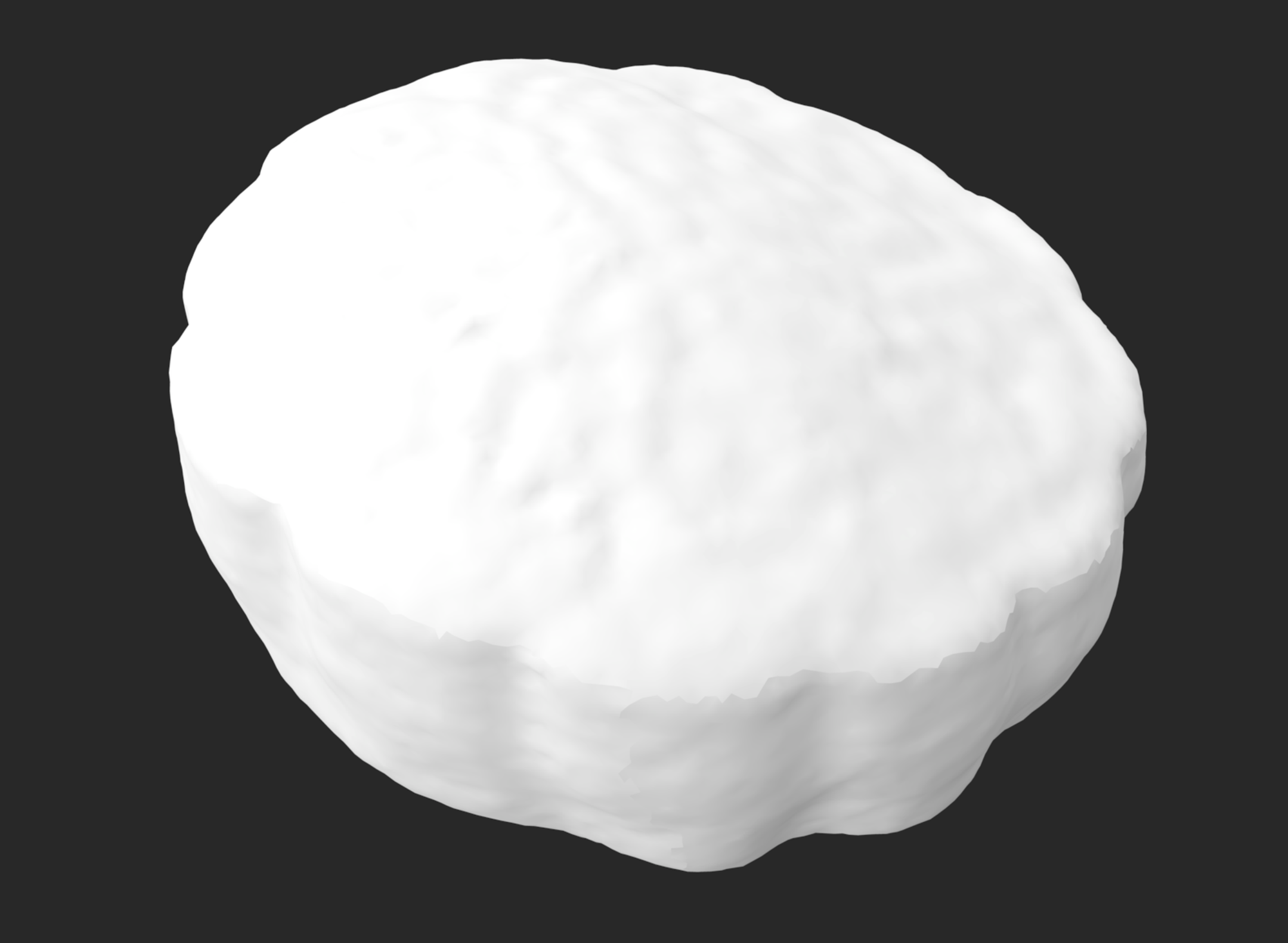}   
       \\
         image of baguette  & mesh of  baguette   &  image of mooncake  &  mesh of  mooncake \\
    \includegraphics[width=0.23\textwidth]{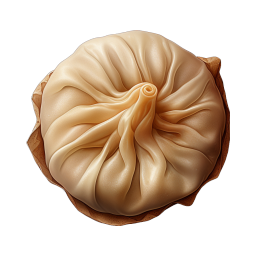}  &
       \includegraphics[width=0.23\textwidth]{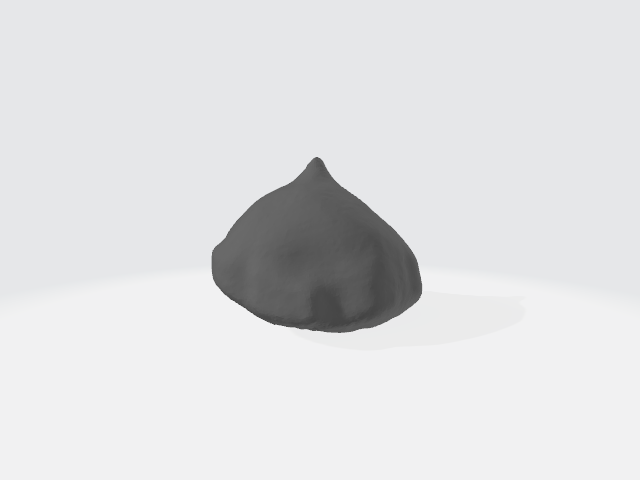} &
     \includegraphics[width=0.23\textwidth]{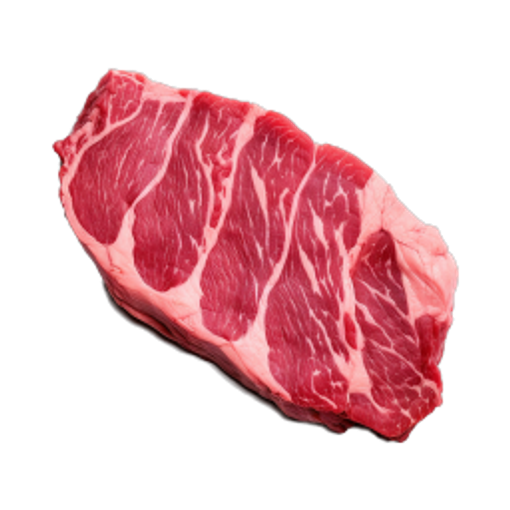}  &  \includegraphics[width=0.23\textwidth]{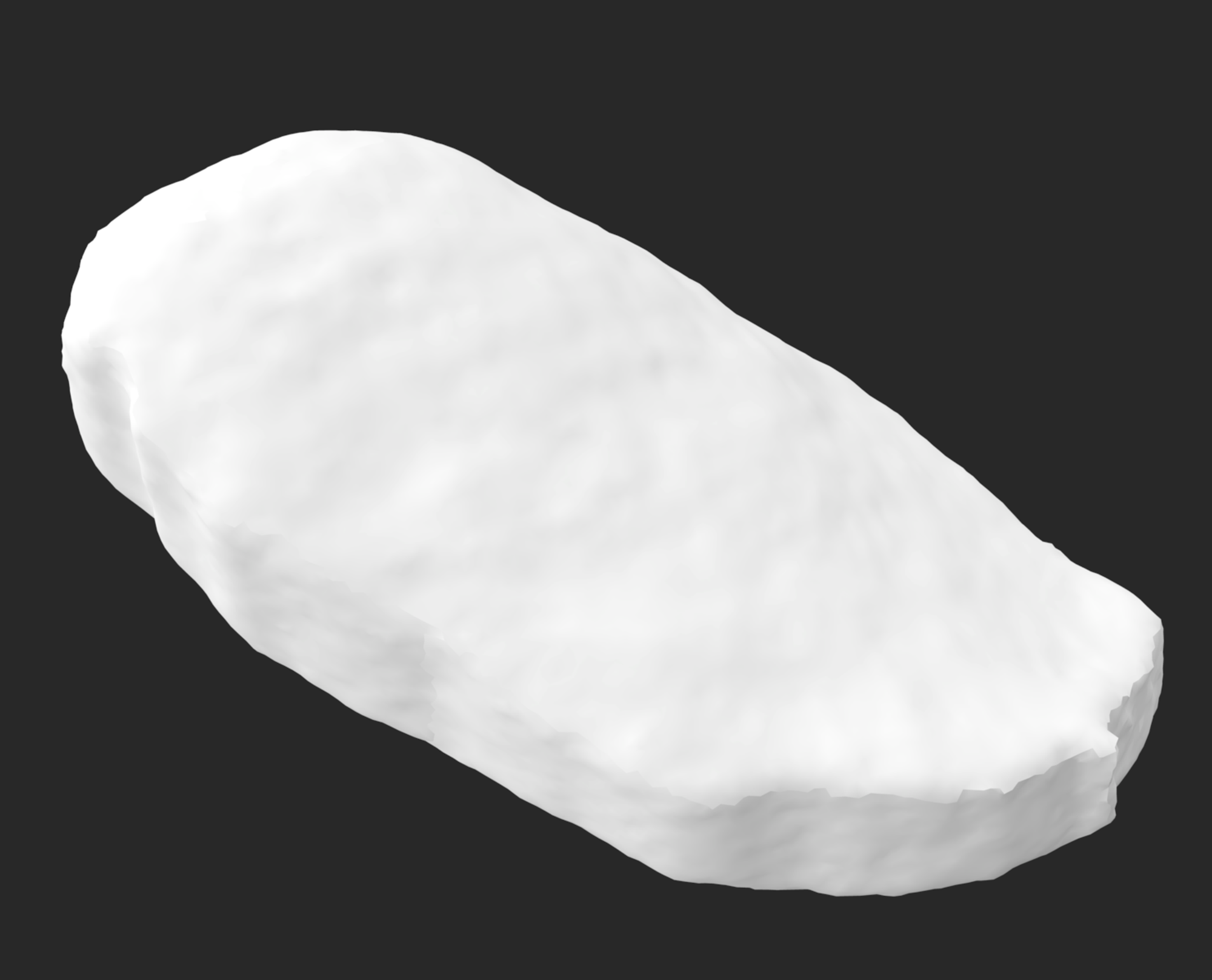}   
       \\
         image of dumpling  & mesh of  dumpling   &  image of meat  &  mesh of  meat \\

    \end{tabular}
    \caption{Example generated images and meshes from our text-to-image-to-3d pipeline.}
    \label{tab:asset-generation}
\end{table}

\section{Prompts}
\label{sec:prmopt}

\subsection{Pre-defined tasks for example-based initialization of 
\label{app:ref}
\model{} (purely non-articulated object manipulation tasks).}
For task proposal of non-articulated object manipulation tasks, we use example-based seeding for \model{}. Below is the list of example tasks. 
\begin{lstlisting}
"""
Task: stack two blocks, with the larger one at the bottom.
Object: A small block, and a large block.
""",

"""
Taks: Put the broccoli on the grill pan
Objects: a broccoli, a grill pan
""",

"""
Task: Put 1 mug on the cup holder
Objects: A mug, a mug tree holder
""",

"""
Task: Pick up the hanger and place it on the clothing rack
Objects: a cloth hanger, a clothing rack
""",

"""
Task: Put 1 book into the bookshelf
Objects: a book, a bookshelf
""",

"""
Taks: Put the knife on the chopping board
Objects: a kitchen knife, a board
""",

"""
Task: Put a old toy in bin
Objects: A old toy, a rubbish bin
""",

"""
Task: Place the dishes and cutlery on the table in preparation for a meal
Objects: a dish plate, a fork, a spoon, a steak knife
""",

"""
Task: Stack one cup on top of the other
Objects: Two same cups
""",

"""
Task: Remove the green pepper from the weighing scales and place it on the floor
Objects: A green pepper, a weighing scale 
""",

"""
Task: Put the apple on the weighing scale to weigh it
Objects: An apple, a weighing scale
""",

\end{lstlisting}

\subsection{Pre-defined tasks for example-based initialization of 
\model{} (locomotion tasks).} 
For task proposal of locomotion tasks, we included 3 examples in the prompt as the seeding for \model{}.  
\begin{lstlisting}
"""
Skill: flip rightwards
Reward:
```python
def _compute_reward(self):
    # we first get some information of the quadruped/humanoid robot.
    # COM_pos and COM_quat are the position and orientation (quaternion) of the center of mass of the quadruped/humanoid.
    COM_pos, COM_quat = get_robot_pose(self)
    # COM_vel, COM_ang are the velocity and angular velocity of the center of mass of the quadruped/humanoid.
    COM_vel, COM_ang = get_robot_velocity(self)

    # face_dir, side_dir, and up_dir are three axes of the rotation of the quadruped/humanoid.
    # face direction points from the center of mass towards the face direction of the quadruped/humanoid.
    # side direction points from the center of mass towards the side body direction of the quadruped/humanoid.
    # up direction points from the center of mass towards up, i.e., the negative direction of the gravity. 
    # gravity direction is [0, 0, -1].
    # when initialized, the face of the robot is along the x axis, the side of the robot is along the y axis, and the up of the robot is along the z axis.
    face_dir, side_dir, up_dir = get_robot_direction(self, COM_quat)

    target_side = np.array([0, 1, 0]) # maintain initial side direction during flip
    target_ang = np.array([50, 0, 0.0]) # spin around x axis to do the rightwards flip, since x is the face direction of the robot.

    alpha_ang = 1.0
    alpha_side = 1.0

    r_ang    = - alpha_ang * np.linalg.norm(COM_ang - target_ang)
    r_side   = - alpha_side * np.linalg.norm(side_dir - target_side)
    r += r_ang + r_side

    # there is a default energy term that penalizes the robot for consuming too much energy. This should be included for all skill.
    r_energy = get_energy_reward(self)
    return r + r_energy
```
""",

"""
Skill: jump backward
Reward:
```python
def _compute_reward(self):
    # we first get some information of the quadruped/humanoid.
    # COM_pos and COM_quat are the position and orientation (quaternion) of the center of mass of the quadruped/humanoid.
    COM_pos, COM_quat = get_robot_pose(self)
    # COM_vel, COM_ang are the velocity and angular velocity of the center of mass of the quadruped/humanoid.
    COM_vel, COM_ang = get_robot_velocity(self)

    # face_dir, side_dir, and up_dir are three axes of the rotation of the quadruped/humanoid.
    # face direction points from the center of mass towards the face direction of the quadruped/humanoid.
    # side direction points from the center of mass towards the side body direction of the quadruped/humanoid.
    # up direction points from the center of mass towards up, i.e., the negative direction of the gravity. 
    # gravity direction is [0, 0, -1].
    # when initialized, the face of the robot is along the x axis, the side of the robot is along the y axis, and the up of the robot is along the z axis.
    face_dir, side_dir, up_dir = get_robot_direction(self, COM_quat)

    if self.time_step <= 30: # first a few steps the robot are jumping
        target_height = 5.0
    else: # then it should not jump
        target_height = 0.0

    target_v = np.array([-5.0, 0, 0.0]) # jump backwards
    target_up = np.array([0, 0, 1]) # maintain up direction
    target_face = np.array([1, 0, 0]) # maintain initial face direction
    target_side = np.array([0, 1, 0]) # maintain initial side direction
    target_ang = np.array([0, 0, 0.0]) # don't let the robot spin

    alpha_vel = 5.0
    alpha_ang = 1.0
    alpha_face = 1.0
    alpha_up = 1.0
    alpha_side = 1.0
    alpha_height = 10.0

    r_vel    = - alpha_vel * np.linalg.norm(COM_vel - target_v)
    r_ang    = - alpha_ang * np.linalg.norm(COM_ang - target_ang)
    r_face   = - alpha_face * np.linalg.norm(face_dir - target_face)
    r_up     = - alpha_up * np.linalg.norm(up_dir - target_up)
    r_side   = - alpha_side * np.linalg.norm(side_dir - target_side)
    r_height = - alpha_height * np.linalg.norm(COM_pos[2] - target_height)
    r = r_vel + r_ang + r_face + r_up + r_side + r_height

    # there is a default energy term that penalizes the robot for consuming too much energy. This should be included for all skill.
    r_energy = get_energy_reward(self)
    return r + r_energy
```
""",

"""
Skill: walk to ball
Object: ball # for this task there is a ball in the environment
Location: [1, 0, 0] # we put it at the position [1, 0, 0]. The robot is initialized at the origin [0, 0, 0].
Reward:
```python
def _compute_reward(self):
    # we first get some information of the quadruped/humanoid.
    # COM_pos and COM_quat are the position and orientation (quaternion) of the center of mass of the quadruped/humanoid.
    COM_pos, COM_quat = get_robot_pose(self)
    # COM_vel, COM_ang are the velocity and angular velocity of the center of mass of the quadruped/humanoid.
    COM_vel, COM_ang = get_robot_velocity(self)

    # face_dir, side_dir, and up_dir are three axes of the rotation of the quadruped/humanoid.
    # face direction points from the center of mass towards the face direction of the quadruped/humanoid.
    # side direction points from the center of mass towards the side body direction of the quadruped/humanoid.
    # up direction points from the center of mass towards up, i.e., the negative direction of the gravity. 
    # gravity direction is [0, 0, -1].
    # when initialized, the face of the robot is along the x axis, the side of the robot is along the y axis, and the up of the robot is along the z axis.
    face_dir, side_dir, up_dir = get_robot_direction(self, COM_quat)

    target_v = np.array([0.0, 0, 0.0])
    target_up = np.array([0, 0, 1])
    target_face = np.array([1, 0, 0])
    target_side = np.array([0, 1, 0])
    target_ang = np.array([0, 0, 0.0])

    alpha_vel = 0.0
    alpha_ang = 0.0
    alpha_face = 1.0
    alpha_up = 1.0
    alpha_side = 1.0
    alpha_height = 1.0

    r_vel    = - alpha_vel * np.linalg.norm(COM_vel - target_v)
    r_ang    = - alpha_ang * np.linalg.norm(COM_ang - target_ang)
    r_face   = - alpha_face * np.linalg.norm(face_dir - target_face)
    r_up     = - alpha_up * np.linalg.norm(up_dir - target_up)
    r_side   = - alpha_side * np.linalg.norm(side_dir - target_side)
    r_height = - alpha_height * np.linalg.norm(COM_pos[2] - self.COM_init_pos[2])
    r += r_vel + r_ang + r_face + r_up + r_height

    # don't want the ball to move
    obj_pos, obj_quat = get_obj_pose("ball")
    obj_vel, obj_ang = get_obj_vel("ball")
    target_obj_vel = np.array([0.0, 0, 0.0])
    alpha_obj_vel = 1.0
    r_obj_vel = - alpha_obj_vel * np.linalg.norm(obj_vel - target_obj_vel)
    r += r_obj_vel
    
    # move towards the ball
    r_dist  = - np.linalg.norm(COM_pos - obj_pos)
    r += r_dist

    # there is a default energy term that penalizes the robot for consuming too much energy. This should be included for all skill.
    r_energy = get_energy_reward(self)
    return r + r_energy
``` 
""",

\end{lstlisting}

\subsection{Prompt for soft body manipulation tasks}
The task proposal prompt for soft body manipulation is as follows:
\begin{lstlisting}
We need you to generate some robot learning tasks involving manipulation of soft materials, especially those focused on making baked foods.

Please think of 5 suitable table-top tasks involving manipulating soft body objects in common household scenarios. 

You should first think of the soft-body object to manipulate, and then you can choose what tools to use in the next conversation.

Note: you should only think of meaningful tabletop manipulation tasks in household settings. Please do not think of tasks that are not common in household scenarios. You should make sure that the tasks can be solved by a robot arm.

Please do not think of tasks that involve chemical change of the objects, such as boiling the water, or frying a steak. 

Please output a list with 5 different task names.

\end{lstlisting}

After we obtain the task names, we use the following prompt for scene generation as well as training supervision generation. We assume soft body manipulation tasks can be described using a initial configuration and a goal configuration. For both the initial and the goal configuration, we ask GPT-4 to generate a text description of the soft body. This text description is used as input to our text-to-image-to-3d pipeline to generate the mesh of the softbody. The training supervision takes a fixed form as the earth mover distance between the mesh in the initial configuration and the mesh in the goal configuration. 

\begin{lstlisting}
We need you to generate some robot learning tasks involving soft materials.

We have successfully built a system that builds simulation environments using the following YAML file format, and we also have a system that translates language descriptions into meshes. The YAML files can import the mesh into the simulation environment.

We believe that a successful task consists of two parts: a start YAML file and a goal YAML file.

The YAML file must follow the following rules.

I will give you the name of the task you need to build.

Here is an example: 

Task1: Make a pretzel

Description: Reshape dough into pretzel.

start.yaml

```yaml
# note: every child leads a different object
# note: You can only use type 'mesh'.
childs:
  - type: 'Mesh' 
    obj_cfg:
      file: cube_dough.obj
      scale: (0.1, 0.1, 0.1) 
      # The first dimension is the x-axis, which expands horizontally, the second dimension is the y-axis, which expands vertically, and the third dimension is the z-axis, which represents the horizontal height.
      pos: (0.5, 0.5, 0.5)
    material_cfg: 
      # note: Define name for the material
      name: 'cube_dough'
      # note: Use rgba to control the color
      color: (0.9, 0.9, 0.9, 1.0)
      # We need to provide a detailed language description as the image_prompt, which is used as input for a text-to-image model to generate an image, so we want a fully detailed one.
      # Prompt must be a string.    
      Image_prompt: 'a cube dough, no background, top-view'
      # Mesh_prompt helps an image-to-mesh model to generate a mesh from an image, so we want a fully detailed one. It must follow the rule of prompt.
      Mesh_prompt: 'a cube dough'
```

goal.yaml

```yaml
childs:
  - type: 'Mesh'
    obj_cfg:
      file: pretzel.obj
      pos: (0.5, 0.5, 0.5)
      scale: (0.1, 0.1, 0.1) 
      # contorl the size of the mesh
      # all the mesh will be scaled into default size as 1 meter in length, 1 meter in height, 1 meter in width
      # In this case, the mesh will be 0.1 meter in width, 0.1 meter in length, 0.1 meter in height.
    material_cfg: 
      name: 'pretzel'
      color: (0.9, 0.9, 0.9, 1.0)
      # We need to provide a detailed language description as the image_prompt, which is used as input for a text-to-image model to generate an image, so we want a fully detailed one.
      # Prompt must be a string.    
      Image_prompt: 'a pretzel, no background, top-view'
       # Mesh_prompt helps an image-to-mesh model to generate a mesh from an image, so we want a fully detailed one. It must follow the rule of prompt.
      Mesh_prompt: 'a pretzel'
     ```

I want you to generate a task with the name: {TASK NAME}

\end{lstlisting}

\subsection{All prompts for articulated object centric manipulation tasks}
In the following, we show all prompts used for generating an articulated object manipulation task using \model{}.

\subsubsection{Prompt for Task Proposal.} 
The first stage of \model{} is task proposal, where it proposes meaningful and diverse tasks for robots to learn. For tasks related to articulated object manipulation, we randomly sample an object from a pre-defined pool, and ask GPT-4 to propose tasks related to the functionality and affordance of the sampled object. We show the prompt we use for this stage here. 

We include one input-output example (when the sampled object is an oven) in the prompt. For the prompt shown here, we ask GPT-4 to generate meaningful tasks related to a trashcan: 
\begin{lstlisting}
I will give you an articulated object, with its articulation tree and semantics. Your goal is to imagine some tasks that a robotic arm can perform with this articulated object in household scenarios. You can think of the robotic arm as a Franka Panda robot. The task will be built in a simulator for the robot to learn it. 

Focus on manipulation or interaction with the object itself. Sometimes the object will have functions, e.g., a microwave can be used to heat food, in these cases, feel free to include other objects that are needed for the task. 
Please do not think of tasks that try to assemble or disassemble the object. Do not think of tasks that aim to clean the object or check its functionality. 

For each task you imagined, please write in the following format: 
Task name: the name of the task.
Description: some basic descriptions of the tasks. 
Additional Objects: Additional objects other than the provided articulated object required for completing the task. 
Links: Links of the articulated objects that are required to perform the task. 
- Link 1: reasons why this link is needed for the task
- Link 2: reasons why this link is needed for the task
- ...
Joints: Joints of the articulated objects that are required to perform the task. 
- Joint 1: reasons why this joint is needed for the task
- Joint 2: reasons why this joint is needed for the task
- ...


Example Input: 

```Oven articulation tree
links: 
base
link_0
link_1
link_2
link_3
link_4
link_5
link_6
link_7

joints: 
joint_name: joint_0 joint_type: revolute parent_link: link_7 child_link: link_0
joint_name: joint_1 joint_type: continuous parent_link: link_7 child_link: link_1
joint_name: joint_2 joint_type: continuous parent_link: link_7 child_link: link_2
joint_name: joint_3 joint_type: continuous parent_link: link_7 child_link: link_3
joint_name: joint_4 joint_type: continuous parent_link: link_7 child_link: link_4
joint_name: joint_5 joint_type: continuous parent_link: link_7 child_link: link_5
joint_name: joint_6 joint_type: continuous parent_link: link_7 child_link: link_6
joint_name: joint_7 joint_type: fixed parent_link: base child_link: link_7
```

```Oven semantics
link_0 hinge door
link_1 hinge knob
link_2 hinge knob
link_3 hinge knob
link_4 hinge knob
link_5 hinge knob
link_6 hinge knob
link_7 heavy oven_body
```

Example output:

Task Name: Open Oven Door
Description: The robotic arm will open the oven door.
Additional Objects: None
Links:
- link_0: from the semantics, this is the door of the oven. The robot needs to approach this door in order to open it. 
Joints: 
- joint_0: from the articulation tree, this is the revolute joint that connects link_0. Therefore, the robot needs to actuate this joint for opening the door.


Task Name: Adjust Oven Temperature
Description: The robotic arm will turn one of the oven's hinge knobs to set a desired temperature.
Additional Objects: None
Links:
- link_1: the robot needs to approach link_1, which is assumed to be the temperature knob, to rotate it to set the temperature.
Joints:
- joint_1: joint_1 connects link_1 from the articulation tree. The robot needs to actuate it to rotate link_1 to the desired temperature.


Task Name: Heat a hamburger Inside Oven 
Description: The robot arm places a hamburger inside the oven, and sets the oven temperature to be appropriate for heating the hamburger.
Additional Objects: hamburger
Links:
- link_0: link_0 is the oven door from the semantics. The robot needs to open the door in order to put the hamburger inside the oven.
link_1: the robot needs to approach link_1, which is the temperature knob, to rotate it to set the desired temperature.
Joints:
- joint_0: from the articulation tree, this is the revolute joint that connects link_0 (the door). Therefore, the robot needs to actuate this joint for opening the door.
- joint_1: from the articulation tree, joint_1 connects link_1, which is the temperature knob. The robot needs to actuate it to rotate link_1 to the desired temperature.

Task Name: Set Oven Timer
Description: The robot arm turns a timer knob to set cooking time for the food.
Additional Objects: None.
Links: 
- link_2: link_2 is assumed to be the knob for controlling the cooking time. The robot needs to approach link_2 to set the cooking time.
Joints:
- joint_2: from the articulation tree, joint_2 connects link_2. The robot needs to actuate joint_2 to rotate link_2 to the desired position, setting the oven timer.


Can you do the same for the following object:

```TrashCan articulation tree
links: 
base
link_0
link_1

joints: 
joint_name: joint_0 joint_type: revolute parent_link: link_1 child_link: link_0
joint_name: joint_1 joint_type: fixed parent_link: base child_link: link_1

```
```TrashCan semantics
link_0 hinge door
link_1 free trashcan_body

```

\end{lstlisting}

Here we provide an example response from GPT-4 with the sampled object being a trashcan. GPT-4 come up with two tasks: Open Trash Can and Throw Trash Away. It also returns an additional object as "a pile of trash" for the task of Throw Trash Away. It returns the involved links and joints for these two tasks as well. 
\begin{lstlisting}
Task Name: Open Trash Can
Description: The robotic arm will open the trash can lid.
Additional Objects: None
Links:
- link_0: This link represents the lid of the trash can, based on the sematics. The robot needs to approach this lid to open it.
Joints: 
- joint_0: The joint_0 in the articulation tree connects link_0 (the lid). Thus, the robot needs to actuate this joint to open the lid.

Task Name: Throw Trash Away
Description: The robotic arm places an item of trash inside the trash can.
Additional Objects: A pile of trash
Links:
- link_0: Link_0, as the trash can lid, needs to be approached and opened by the robot in order to throw away the trash.
Joints:
- joint_0: Joint_0, connecting to link_0, needs to be actuated for the robot to open the trash can lid.
\end{lstlisting}

\subsubsection{Scene Generation Prompt. } 
Once a task proposal is obtained, \model{} then generates a corresponding scene for solving the task by populating the environment with a number of relevant and necessary objects (\textit{assets}).
Continuing the above example, we show the scene generation prompt for the task ``Throw Trash Away''. We include 3 input-output examples in the prompt. 
As shown in the prompt, we ask GPT-4 to output the scene in a yaml file format, such that it can be easily processed and subsequently used to build the scene in simulation. 
\begin{lstlisting}
I need you to describe the initial scene configuration for a given task in the following format, using a yaml file. This yaml file will help build the task in a simulator. The task is for a mobile Franka panda robotic arm to learn a manipulation skill in the simulator. The Franka panda arm is mounted on a floor, at location (1, 1, 0). It can move freely on the floor. The z axis is the gravity axis. 

The format is as follows:
```yaml 
- use_table: whether the task requires using a table. This should be decided based on common sense. If a table is used, its location will be fixed at (0, 0, 0). The height of the table will be 0.6m. Usually, if the objects invovled in the task are usually placed on a table (not directly on the ground), then the task requires using a table.
# for each object involved in the task, we need to specify the following fields for it.
- type: mesh
  name: name of the object, so it can be referred to in the simulator
  size: describe the scale of the object mesh using 1 number in meters. The scale should match real everyday objects. E.g., an apple is of scale 0.08m. You can think of the scale to be the longest dimension of the object. 
  lang: this should be a language description of the mesh. The language should be a concise description of the obejct, such that the language description can be used to search an existing database of objects to find the object.
  path: this can be a string showing the path to the mesh of the object. 
  on_table: whether the object needs to be placed on the table (if there is a table needed for the task). This should be based on common sense and the requirement of the task. E.g., a microwave is usually placed on the table.
  center: the location of the object center. If there isn't a table needed for the task or the object does not need to be on the table, this center should be expressed in the world coordinate system. If there is a table in the task and the object needs to be placed on the table, this center should be expressed in terms of the table coordinate, where (0, 0, 0) is the lower corner of the table, and (1, 1, 1) is the higher corner of the table. In either case, you should try to specify a location such that there is no collision between objects.
```

An example input includes the task names, task descriptions, and objects involved in the task. I will also provide with you the articulation tree and semantics of the articulated object. 
This can be useful for knowing what parts are already in the articulated object, and thus you do not need to repeat those parts as separate objects in the yaml file.

Your task includes two parts:
1. Output the yaml configuration of the task.
2. Sometimes, the task description / objects involved will refer to generic/placeholder objects, e.g., to place an "item" into the drawer, and to heat "food" in the microwave. In the generated yaml config, you should change these placeholder objects to be concrete objects in the lang field, e.g., change "item" to be a toy or a pencil, and "food" to be a hamburger, a bowl of soup, etc. 

Example input:
Task Name: Insert Bread Slice 
Description: The robotic arm will insert a bread slice into the toaster.
Objects involved: Toaster, bread slice. Only the objects specified here should be included in the yaml file.

```Toaster articulation tree
links: 
base
link_0
link_1
link_2
link_3
link_4
link_5

joints: 
joint_name: joint_0 joint_type: continuous parent_link: link_5 child_link: link_0
joint_name: joint_1 joint_type: prismatic parent_link: link_5 child_link: link_1
joint_name: joint_2 joint_type: prismatic parent_link: link_5 child_link: link_2
joint_name: joint_3 joint_type: prismatic parent_link: link_5 child_link: link_3
joint_name: joint_4 joint_type: prismatic parent_link: link_5 child_link: link_4
joint_name: joint_5 joint_type: fixed parent_link: base child_link: link_5
```

```Toaster semantics
link_0 hinge knob
link_1 slider slider
link_2 slider button
link_3 slider button
link_4 slider button
link_5 free toaster_body
```


An example output:
```yaml
- use_table: True ### Toaster and bread are usually put on a table. 
- type: mesh
  name: "Toaster"
  on_table: True # Toasters are usually put on a table.
  center: (0.1, 0.1, 0) # Remember that when an object is placed on the table, the center is expressed in the table coordinate, where (0, 0, 0) is the lower corner and (1, 1, 1) is the higher corner of the table. Here we put the toaster near the lower corner of the table.  
  size: 0.35 # the size of a toaster is roughly 0.35m
  lang: "a common toaster"
  path: "toaster.urdf"
- type: mesh
  name: "bread slice"
  on_table: True # Bread is usually placed on the table as well. 
  center: (0.8, 0.7, 0) # Remember that when an object is placed on the table, the center is expressed in the table coordinate, where (0, 0, 0) is the lower corner and (1, 1, 1) is the higher corner of the table. Here we put the bread slice near the higher corner of the table.  
  size: 0.1 # common size of a bread slice 
  lang: "a slice of bread"
  Path: "bread_slice.obj"
```

Another example input:
Task Name: Removing Lid From Pot
Description: The robotic arm will remove the lid from the pot.
Objects involved: KitchenPot. Only the objects specified here should be included in the yaml file.

```KitchenPot articulation tree
links: 
base
link_0
link_1

joints: 
joint_name: joint_0 joint_type: prismatic parent_link: link_1 child_link: link_0
joint_name: joint_1 joint_type: fixed parent_link: base child_link: link_1
```

```KitchenPot semantics
link_0 slider lid
link_1 free pot_body
```
Output:
```yaml
- use_table: True # A kitchen pot is usually placed on the table.
- type: mesh
  name: "KitchenPot"
  on_table: True # kitchen pots are usually placed on a table. 
  center: (0.3, 0.6, 0) # Remember that when an object is placed on the table, the center is expressed in the table coordinate, where (0, 0, 0) is the lower corner and (1, 1, 1) is the higher corner of the table. Here we put the kitchen pot just at a random location on the table.  
  size: 0.28 # the size of a common kitchen pot is roughly 0.28m
  lang: "a common kitchen pot"
  path: "kitchen_pot.urdf"
```
Note in this example, the kitchen pot already has a lid from the semantics file. Therefore, you do not need to include a separate lid in the yaml file.


One more example input:
Task Name: Heat a hamburger in the oven.
Description: The robotic arm will put a hamburger in the oven and use the oven to heat it.
Objects involved: A hamburger, an oven. Only the objects here should be included in the yaml file.

```Oven articulation tree
links: 
base
link_0
link_1
link_2
link_3
link_4
link_5
link_6
link_7

joints: 
joint_name: joint_0 joint_type: revolute parent_link: link_7 child_link: link_0
joint_name: joint_1 joint_type: continuous parent_link: link_7 child_link: link_1
joint_name: joint_2 joint_type: continuous parent_link: link_7 child_link: link_2
joint_name: joint_3 joint_type: continuous parent_link: link_7 child_link: link_3
joint_name: joint_4 joint_type: continuous parent_link: link_7 child_link: link_4
joint_name: joint_5 joint_type: continuous parent_link: link_7 child_link: link_5
joint_name: joint_6 joint_type: continuous parent_link: link_7 child_link: link_6
joint_name: joint_7 joint_type: fixed parent_link: base child_link: link_7
```

```Oven semantics
link_0 hinge door
link_1 hinge knob
link_2 hinge knob
link_3 hinge knob
link_4 hinge knob
link_5 hinge knob
link_6 hinge knob
link_7 heavy oven_body
```

Output:
```yaml
- use_table: True # Although an oven is usually just placed on the ground, a hamburger is usually placed on the table.
- type: mesh
  name: "Oven"
  on_table: False # An oven is usually just placed on the floor.
  center: (1.0, 0, 0) # Remember that when not on a table, the center is expressed in the world coordinate. Since the robot is at (1, 1, 0) and the table is at (0, 0, 0), we place the oven at (1.8, 2, 0) to avoid collision with the table and the robot.
  size: 0.9 # the size of an oven is roughly 0.9m
  lang: "a common oven"
  path: "oven.urdf"
- type: mesh
  name: "hamburger"
  on_table: True # a hamburger is usually placed on the table. 
  center: (0.2, 0.4, 0) # Remember that when an object is placed on the table, the center is expressed in the table coordinate, where (0, 0, 0) is the lower corner and (1, 1, 1) is the higher corner of the table. Here we put the hamburger just at a random location on the table.  
  size: 0.1 # common hamburgers' diameters are 10 cm.
  lang: "a common hamburger"
  path: "hamburger.obj"
```

Rules: 
- You do not need to include the robot in the yaml file.
- The yaml file should only include the objects listed in "Objects involved".
- Sometimes, the task description / objects involved will refer to generic/placeholder objects, e.g., to place an "item" into the drawer, and to heat "food" in the microwave. In the generated yaml config, you should change these placeholder objects to be concrete objects in the lang field, e.g., change "item" to be a toy or a pencil, and "food" to be a hamburger, a bowl of soup, etc. 


Can you do this for the following task:
Task Name: Throw Trash Away
Description: The robotic arm places an item of trash inside the trash can
Objects involved: TrashCan, A pile of trash

```TrashCan articulation tree
links: 
base
link_0
link_1

joints: 
joint_name: joint_0 joint_type: revolute parent_link: link_1 child_link: link_0
joint_name: joint_1 joint_type: fixed parent_link: base child_link: link_1

```
```TrashCan semantics
link_0 hinge door
link_1 free trashcan_body

```

\end{lstlisting}

Here we show an example response from GPT-4. The generated scene yaml file specifies the location of each object, its physical size, along with a language description of the object, which will be used when retrieving the object from Objaverse. 
\begin{lstlisting}
Output:
```yaml
- use_table: False # Trash cans are usually placed on the floor.
- type: mesh
  name: "TrashCan"
  on_table: False # Trash cans are usually placed on the floor.
  center: (1.5, 1.5, 0) # Remember that when not on a table, the center is expressed in the world coordinate. Since the robot is at (1, 1, 0) and the table is at (0, 0, 0), we place the trash can at (1.5, 1.5, 0) to avoid collision with the table and the robot.
  size: 0.6 # the size of a common trash can is roughly 0.6m
  lang: "a common trash can"
  path: "trashcan.urdf"
- type: mesh
  name: "Trash"
  on_table: True # Trash is usually picked up from a table or the floor. Here we assume it's on the table. 
  center: (0.5, 0.5, 0) # Remember that when an object is placed on the table, the center is expressed in the table coordinate, where (0, 0, 0) is the lower corner and (1, 1, 1) is the higher corner of the table. Here we put the trash in the middle of the table.  
  size: 0.1 # common size of a pile of trash 
  lang: "a pile of paper waste" # Note here, we changed the generic/placeholder "pile of trash" object to be a more concrete object: a pile of paper waste.
  path: "paper_waste.obj"
```

\end{lstlisting}

\subsubsection{Prompt for retrieving semantically relevant assets. }
To further increase the complexity and diversity of the generated scenes while resembling object distributions of real-world scenarios, we query GPT-4 to return a number of additional queries (object names and their descriptions) that are semantically relevant to the task. Here we show the prompt for this part. We include one input-output example in the prompt.
We again ask GPT-4 to output the list of semantically relevant assets in the yaml file format. 
\begin{lstlisting}
Given a task, which is for a mobile Franka panda robotic arm to learn a manipulation skill in the simulator, your goal is to add more objects into the task scene such that the scene looks more realistic. The Franka panda arm is mounted on a floor, at location (1, 1, 0). It can move freely on the floor. The z axis is the gravity axis. 

The input to you includes the following:
Task name, task description, the essential objects involved in the task, and a config describing the current task scene, which contains only the essential objects needed for the task. The config is a yaml file in the following format:

```yaml 
- use_table: whether the task requires using a table. This should be decided based on common sense. If a table is used, its location will be fixed at (0, 0, 0). The height of the table will be 0.6m. 
# for each object involved in the task, we need to specify the following fields for it.
- type: mesh
  name: name of the object, so it can be referred to in the simulator
  size: describe the scale of the object mesh using 1 number in meters. The scale should match real everyday objects. E.g., an apple is of scale 0.08m. You can think of the scale to be the longest dimension of the object. 
  lang: this should be a language description of the mesh. The language should be a bit detailed, such that the language description can be used to search an existing database of objects to find the object.
  path: this can be a string showing the path to the mesh of the object. 
  on_table: whether the object needs to be placed on the table (if there is a table needed for the task). This should be based on common sense and the requirement of the task.     
  center: the location of the object center. If there isn't a table needed for the task or the object does not need to be on the table, this center should be expressed in the world coordinate system. If there is a table in the task and the object needs to be placed on the table, this center should be expressed in terms of the table coordinate, where (0, 0, 0) is the lower corner of the table, and (1, 1, 1) is the higher corner of the table. In either case, you should try to specify a location such that there is no collision between objects.
```

Your task is to think about what other distractor objects can be added into the scene to make the scene more complex and realistic for the robot to learn the task. These distractor objects are not necessary for the task itself, but their existence makes the scene look more interesting and complex. You should output the distractor objects using the same format as the input yaml file. You should try to put these distractor objects at locations such that they don't collide with objects already in the scene.

Here is one example:

Input:

Task name: Heat up a bowl of soup in the microwave
Task description: The robot will grab the soup and move it into the microwave, and then set the temperature to heat it.
Objects involved: Microwave, a bowl of soup
Config:
```yaml
-   use_table: true
-   center: (0.3, 0.7, 0)
    lang: A standard microwave with a turntable and digital timer
    name: Microwave
    on_table: true
    path: microwave.urdf
    size: 0.6
    type: urdf
-   center: (0.2, 0.2, 0)
    lang: A ceramic bowl full of soup
    name: Bowl of Soup
    on_table: true
    path: bowl_soup.obj
    size: 0.15
    type: mesh
```

Output: 
```yaml
- name: plate # a plate is a common object placed when there is microwave and bowl of soup, in a kitchen setup
  lang: a common kitchen plate
  on_table: True
  center: (0.8, 0.8, 0)
  type: mesh
  path: "plate.obj"
  size: 0.15 # a plate is usually of scale 0.15m
- name: sponge # a sponge is a common object placed when there is microwave and bowl of soup, in a kitchen setup
  lang: a common sponge
  on_table: True
  center: (0.5, 0.2, 0)
  type: mesh
  path: "sponge.obj"
  size: 0.1 # a sponge is usually of scale 0.1m
- name: Oven # a oven is a common object placed when there is microwave and bowl of soup, in a kitchen setup
  lang: a kitchen oven
  on_table: False # an oven is usually a standalone object on the floor
  center: (1.8, 0.5, 0) # remember robot is at (1, 1, 0) and table is at (0, 0, 0). So the oven is placed at (1.8, 0.5, 0) in the world coordinate system to avoid collision with other objects.
  type: mesh
  path: "oven.obj"
  size: 0.8 # an oven is usually of scale 0.8m
```

Can you do it for the following task:
Task name: Throw Trash Away
Task description: The robotic arm places an item of trash inside the trash can
Initial config:
```yaml
- use_table: false
- center: (1.5, 1.5, 0)
  lang: a common trash can
  name: TrashCan
  on_table: false
  path: trashcan.urdf
  size: 0.6
  type: urdf
- center: (0.5, 0.5, 0)
  lang: a pile of paper waste
  name: Trash
  on_table: true
  path: paper_waste.obj
  size: 0.05
  type: mesh

```
\end{lstlisting}

Here is an example response from GPT-4. For the task of throwing away trash, the addition objects are: a broom, a dustpan, and a recycling bin, which are common objects for a scene where one needs to throw trash away. 
\begin{lstlisting}
Sure, here is the output for the given task:

```yaml
- name: broom # a broom is a common object found near a trash can
  lang: a common broom
  on_table: False
  center: (2, 1.5, 0)
  type: mesh
  path: "broom.obj"
  size: 1.2 # a broom is usually of scale 1.2m
- name: dustpan # a dustpan is a common object found near a trash can
  lang: a common dustpan
  on_table: False
  center: (2, 2, 0)
  type: mesh
  path: "dustpan.obj"
  size: 0.3 # a dustpan is usually of scale 0.3m
- name: recycling_bin # a recycling bin is a common object found near a trash can
  lang: a common recycling bin
  on_table: False
  center: (1, 2, 0)
  type: mesh
  path: "recycling_bin.obj"
  size: 0.6 # a recycling bin is usually of scale 0.6m
- name: soda_can # a soda can is a common object found near a trash can
  lang: a common soda can
  on_table: True
  center: (0.7, 0.7, 0)
  type: mesh
  path: "soda_can.obj"
  size: 0.1 # a soda can is usually of scale 0.1m
```
\end{lstlisting}

\subsubsection{Asset size adjustment prompt. } 
Because the assets retrieved from Objaverse is usually not of correct physical sizes, we ask GPT-4 to correct any unreasonable object sizes. 
For each object in the scene, we query GPT-4 to correct its size such that the size matches the real-world size. 
We continue the above example of throw trash away. 
\begin{lstlisting}
A robotic arm is trying to manipulate some objects to learn corresponding skills in a simulator. However, the size of the objects might be wrong. Your task is to adjust the size of the objects, such that they match each other when interact with each other; and the size should also match what is commonly seen in everyday life, in household scenarios. 

Now I will give you the name of the task, the object and their sizes, please correct any unreasonable sizes. 

Objects are represented using a mesh file, you can think of size as the longest dimension of the object. 

I will write in the following format:
```
Task: task description
obj1, mesh, size 
obj2, mesh, size
```

Please reply in the following format:
explanations of why some size is not reasonable.
```yaml
obj1, mesh, corrected_size
obj2, mesh, corrected_radius
```

Here is an example:
Input: 
```
Task: The robotic arm lowers the toilet seat from an up position to a down position
Toilet, mesh, 0.2
```

Output:
A toilet is usually 0.6 - 0.8m in its back height, so the size is not reasonable -- it is a bit too small. Below is the corrected size.
```yaml
Toilet, mesh, 0.7
```

Another example:
Input:
```
Task: Fill a cup with water under the faucet
Faucet, mesh, 0.25
Cup, mesh, 0.3
```

Output:
The size of the faucet makes senes. However, the size of the cup is too large for 2 reasons: it does not match the size of tha faucet for getting water under the faucet; and it is not a common size of cup in everyday life. Below is the corrected size.
```yaml
Faucet, mesh, 0.25 
Cup, mesh, 0.12 
```

One more example to show that even if no change is needed, you should still reply with the same size.
Input:
```
Task: Open Table Drawer The robotic arm will open a table drawer
table, mesh, 0.8
```

Output:
The size of the table is reasonable, so no change is needed.
```yaml
table, mesh, 0.8
```
This is also a good example to show that sometimes, the task description might include two objects, e.g., a table and a drawer, yet there is only one object size provided (here the table). This is not an error, but that the other object is part of the provided object, i.e., here the drawer is part of the table. It's fine, you should then just reply with the corrected size of the object provided, here, the table, in such cases.

\end{lstlisting}

\begin{lstlisting}
```
Task: Throw Trash Away The robotic arm places an item of trash inside the trash can
trashcan, mesh, 0.6
trash, mesh, 0.1
```
\end{lstlisting}

An example response from GPT-4, where it adjusted the size of the trash from 0.1m to 0.05m. 
\begin{lstlisting}
The size of the trashcan is reasonable as it is a common size for a household trashcan. However, the size of the trash is a bit too large. Trash items are usually smaller to fit inside the trashcan. Below is the corrected size.
```yaml
trashcan, mesh, 0.6
trash, mesh, 0.05
```
\end{lstlisting}

\subsubsection{Prompt for articulated object joint initialization. } 
For scene generation, we also need to specify the asset configuration, which is the initial joint angle of the articulated object. E.g., for the robot to learn to open the drawer, the drawer needs to be initially closed. 
We continue the example of throwing the trash away. We include multiple input-output examples in the prompt. 

\begin{lstlisting}
Your goal is to set the  joint angles of some articulated objects to the right value in the initial state, given a task. The task is for a robot arm to learn the corresponding skills to manipulate the articulated object. 

The input to you will include the task name, a short description of the task, the articulation tree of the articulated object, a semantic file of the articulated object, the links and joints of the articulated objects that will be involved in the task, and the substeps for doing the task. 

You should output for each joint involved in the task, what joint value it should be set to. You should output a number in the range [0, 1], where 0 corresponds to the lower limit of that joint angle, and 1 corresponds to the upper limit of the joint angle. You can also output a string of "random", which indicates to sample the joint angle within the range.

By default, the joints in an object are set to their lower joint limits. You can assume that the lower joint limit corresponds to the natural state of the articulated object. E.g., for a door's hinge joint, 0 means it is closed, and 1 means it is open. For a lever, 0 means it is unpushed, and 1 means it is pushed to the limit. 

Here is an example:

Input:
Task Name: Close the door
Description: The robot arm will close the door after it was opened. 


```door articulation tree
links: 
base
link_0
link_1
link_2

joints: 
joint_name: joint_0 joint_type: revolute parent_link: link_1 child_link: link_0
joint_name: joint_1 joint_type: fixed parent_link: base child_link: link_1
joint_name: joint_2 joint_type: revolute parent_link: link_0 child_link: link_2
```

```door semantics
link_0 hinge rotation_door
link_1 static door_frame
link_2 hinge rotation_door
```

Links: 
- link_0: link_0 is the door. This is the part of the door assembly that the robot needs to interact with.
Joints:
- joint_0: Joint_0 is the revolute joint connecting link_0 (the door) as per the articulation tree. The robot needs to actuate this joint cautiously to ensure the door is closed.

substeps:
approach the door	
close the door


Output:
The goal is for the robot arm to learn to close the door after it is opened. Therefore, the door needs to be initially opened, thus, we are setting its value to 1, which corresponds to the upper joint limit. 
```joint values
joint_0: 1
```

Another example:
Task Name: Turn Off Faucet
Description: The robotic arm will turn the faucet off by manipulating the switch

```Faucet articulation tree
links: 
base
link_0
link_1

joints: 
joint_name: joint_0 joint_type: fixed parent_link: base child_link: link_0
joint_name: joint_1 joint_type: revolute parent_link: link_0 child_link: link_1
```

```Faucet semantics
link_0 static faucet_base
link_1 hinge switch
```

Links: 
- link_0: link_0 is the door. This is the part of the door assembly that the robot needs to interact with.
Joints:
- joint_0: Joint_0 is the revolute joint connecting link_0 (the door) as per the articulation tree. The robot needs to actuate this joint cautiously to ensure the door is closed.

substeps:
grasp the faucet switch
turn off the faucet

Output:
For the robot to learn to turn off the faucet, it cannot be already off initially. Therefore, joint_1 should be set to its upper joint limit, or any value that is more than half of the joint range, e.g., 0.8.
```joint value
joint_1: 0.8
```

One more example:
Task Name: Store an item inside the Drawer
Description: The robot arm picks up an item and places it inside the drawer of the storage furniture

```StorageFurniture articulation tree
links: 
base
link_0
link_1
link_2

joints: 
joint_name: joint_0 joint_type: revolute parent_link: link_1 child_link: link_0
joint_name: joint_1 joint_type: fixed parent_link: base child_link: link_1
joint_name: joint_2 joint_type: prismatic parent_link: link_1 child_link: link_2
```

```StorageFurniture semantics
link_0 hinge rotation_door
link_1 heavy furniture_body
link_2 slider drawer
```

Links:
- link_2: link_2 is the drawer link from the semantics. The robot needs to open this drawer to place the item inside. 
Joints: 
- joint_2: joint_2, from the articulation tree, connects to link_2 (the drawer). Thus, the robot would need to actuate this joint to open the drawer to store the item.

substeps:
 grasp the drawer
 open the drawer
 grasp the item
 put the item into the drawer
 grasp the drawer again
 close the drawer
 release the grasp


Output:
This task involves putting one item into the drawer of the storage furniture. As noted in the substeps, the robot needs to first open the drawer, put the item in, and then close it. Since the articulated object is initialized with the lower joint limit, i.e., the drawer is initially closed, it aligns with the task where the robot needs to first learn to open the drawer. Therefore, no particular joint angle needs to be set, and we just output None. 

```joint value
None
```

Can you do it for the following task:

Task Name: Throw Trash Away
Description: The robotic arm places an item of trash inside the trash can


```TrashCan articulation tree
links: 
base
link_0
link_1

joints: 
joint_name: joint_0 joint_type: revolute parent_link: link_1 child_link: link_0
joint_name: joint_1 joint_type: fixed parent_link: base child_link: link_1

```


```TrashCan semantics
link_0 hinge door
link_1 free trashcan_body

```

Links:
link_0: Link_0, as the trash can lid, needs to be approached and opened by the robot in order to throw away the trash.

Joints:
joint_0: Joint_0, connecting to link_0, needs to be actuated for the robot to open the trash can lid.

substeps:
 grasp the trash can lid
 open the trash can lid
 grasp the trash
 put the trash into the trash can
 release the trash
 grasp the trash can lid again
 close the trash can lid
\end{lstlisting}

An example output from GPT-4. It correctly reasons that to throw the trash away in the trashcan, the robot needs to learn to open the trashcan, thus the trashcan needs to be initialized as closed (joint angle set to be 0). 
\begin{lstlisting}
    Output:
The task involves the robot learning to throw trash away. The robot needs to learn to open the trash can lid, throw the trash in, and then close the lid. Therefore, we need to set the initial joint angle such that the trash can lid is not already open. We can set joint_0 to its lower limit, which corresponds to the trash can lid being closed. 

```joint value
joint_0: 0
```

\end{lstlisting}

\subsubsection{Prompt for special spatial relationships. } 
For certain tasks, the objects in the scene need to satisfy certain spatial relationships. For example, for the task of retrieve a gold bar from the safe, the gold bar needs to be initially inside the safe. Here we show the prompt we use for that, and we continue the ``Throw away trash'' example. 
\begin{lstlisting}
Your goal is to output any special spatial relationships certain objects should have in the initial state, given a task. The task is for a robot arm to learn the corresponding skills in household scenarios.  

The input to you will include 
the task name, 
a short description of the task, 
objects involved in the task, 
substeps for performing the task,
If there is an articulated object involved in the task, the articulation tree of the articulated object, the semantic file of the articulated object, and the links and joints of the articulated objects that will be involved in the task. 

We have the following spatial relationships:
on, obj_A, obj_B: object A is on top of object B, e.g., a fork on the table.
in, obj_A, obj_B: object A is inside object B, e.g., a gold ring in the safe.
in, obj_A, obj_B, link_name: object A is inside the link with link_name of object B. For example, a table might have two drawers, represented with link_0, and link_1, and in(pen, table, link_0) would be that a pen is inside one of the drawers that corresponds to link_0. 

Given the input to you, you should output any needed spatial relationships of the involved objects. 

Here are some examples:

Input:
Task Name:Fetch Item from Refrigerator 
Description: The robotic arm will open a refrigerator door and reach inside to grab an item and then close the door.
Objects involved: refrigerator, item

```refrigerator articulation tree
links: 
base
link_0
link_1
link_2

joints: 
joint_name: joint_0 joint_type: fixed parent_link: base child_link: link_0
joint_name: joint_1 joint_type: revolute parent_link: link_0 child_link: link_1
joint_name: joint_2 joint_type: revolute parent_link: link_0 child_link: link_2
```

```refrigerator semantics
link_0 heavy refrigerator_body
link_1 hinge door
link_2 hinge door
```

Links:
link_1: The robot needs to approach and open this link, which represents one of the refrigerator doors, to reach for the item inside.
Joints:
joint_1: This joint connects link_1, representing one of the doors. The robot needs to actuate this joint to open the door, reach for the item, and close the door. 


substeps:
 grasp the refrigerator door
 open the refrigerator door
 grasp the item
 move the item out of the refrigerator
 grasp the refrigerator door again
 close the refrigerator door


Output:
The goal is for the robot arm to learn to retrieve an item from the refrigerator. Therefore, the item needs to be initially inside the refrigerator. From the refrigerator semantics we know that link_0 is the body of the refrigerator, therefore we should have a spatial relationship as the following:
```spatial relationship
In, item, refrigerator, link_0
```

Another example:
Task Name: Turn Off Faucet
Description: The robotic arm will turn the faucet off by manipulating the switch
Objects involved: faucet

```Faucet articulation tree
links: 
base
link_0
link_1

joints: 
joint_name: joint_0 joint_type: fixed parent_link: base child_link: link_0
joint_name: joint_1 joint_type: revolute parent_link: link_0 child_link: link_1
```

```Faucet semantics
link_0 static faucet_base
link_1 hinge switch
```

Links: 
link_0: link_0 is the door. This is the part of the door assembly that the robot needs to interact with.
Joints:
joint_0: Joint_0 is the revolute joint connecting link_0 (the door) as per the articulation tree. The robot needs to actuate this joint cautiously to ensure the door is closed.

substeps:
grasp the faucet switch
turn off the faucet

Output:
There is only 1 object involved in the task, thus no special spatial relationships are required.
```spatial relationship
None
```

One more example:
Task Name: Store an item inside the Drawer
Description: The robot arm picks up an item and places it inside the drawer of the storage furniture.
Objects involved: storage furniture, item

```StorageFurniture articulation tree
links: 
base
link_0
link_1
link_2

joints: 
joint_name: joint_0 joint_type: revolute parent_link: link_1 child_link: link_0
joint_name: joint_1 joint_type: fixed parent_link: base child_link: link_1
joint_name: joint_2 joint_type: prismatic parent_link: link_1 child_link: link_2
```

```StorageFurniture semantics
link_0 hinge rotation_door
link_1 heavy furniture_body
link_2 slider drawer
```

Links:
link_2: link_2 is the drawer link from the semantics. The robot needs to open this drawer to place the item inside. 
Joints: 
joint_2: joint_2, from the articulation tree, connects to link_2 (the drawer). Thus, the robot would need to actuate this joint to open the drawer to store the item.

substeps:
 grasp the drawer
 open the drawer
 grasp the item
 put the item into the drawer
 grasp the drawer again
 close the drawer
 release the grasp


Output:
This task involves putting one item into the drawer of the storage furniture. The item should initially be outside of the drawer, such that the robot can learn to put it into the drawer. Therefore, no special relationships of in or on are needed. Therefore, no special spatial relationships are needed.
```spatial relationship
None
```

Can you do it for the following task: 

Task Name: Throw Trash Away
Description: The robotic arm places an item of trash inside the trash can
Objects involved: TrashCan, Trash


```TrashCan articulation tree
links: 
base
link_0
link_1

joints: 
joint_name: joint_0 joint_type: revolute parent_link: link_1 child_link: link_0
joint_name: joint_1 joint_type: fixed parent_link: base child_link: link_1

```


```TrashCan semantics
link_0 hinge door
link_1 free trashcan_body

```

Links:
link_0: Link_0, as the trash can lid, needs to be approached and opened by the robot in order to throw away the trash.

Joints:
joint_0: Joint_0, connecting to link_0, needs to be actuated for the robot to open the trash can lid.

substeps:
 grasp the trash can lid
 open the trash can lid
 grasp the trash
 put the trash into the trash can
 release the trash
 grasp the trash can lid again
 close the trash can lid
\end{lstlisting}

An example output from GPT-4: it decides that there is no speical relationship needed for the task of throw trash away. 
\begin{lstlisting}
Output:
The task involves the robot arm placing an item of trash into the trash can. Initially, the trash should not be inside the trash can, so the robot can learn to put it in. Therefore, no special spatial relationships are needed.
```spatial relationship
None
```
\end{lstlisting}

We have finished all steps needed for scene generation now. 

\subsubsection{Prompt for generating training supervision. } 
To acquire the skill for solving the proposed task, supervisions for skill learning are needed.
To facilitate the learning process, \model{} first queries GPT-4 to plan and decompose the generated task into shorter-horizon sub-tasks.
After the decomposition, \model{} then queries GPT-4 to choose a proper algorithm for solving each sub-task. 
For sub-tasks to be learned using RL, we prompt  GPT-4 to write corresponding reward functions with three in-context examples. For object manipulation and locomotion tasks, the reward functions are based on the low-level states which GPT-4 can query via a provided list of simulator APIs. 
Here we show the prompt we use for this, which includes 3 input-output examples, including the decomposition, the algorithm selection, and the reward if RL is selected as the algorithm. 
With the generated scene and training supervision, we can then perform skill learning to let the robot learn the skill to perform this task. 

\begin{lstlisting}
A robotic arm is trying to solve some household object manipulation tasks to learn corresponding skills in a simulator.

We will provide with you the task description, the initial scene configurations of the task, which contains the objects in the task and certain information about them. 
Your goal is to decompose the task into executable sub-steps for the robot, and for each substep, you should either call a primitive action that the robot can execute, or design a reward function for the robot to learn, to complete the substep. 
For each substep, you should also write a function that checks whether the substep has been successfully completed. 

Common substeps include moving towards a location, grasping an object, and interacting with the joint of an articulated object.

An example task:

Task Name: Fetch item from refrigerator
Description: The robotic arm will open a refrigerator door reach inside to grab an item, place it on the table, and then close the door
Initial config:
```yaml
-   use_table: true 
-   center: (1.2, 0, 0)
    lang: a common two-door refrigerator
    name: Refrigerator
    on_table: false 
    path: refrigerator.urdf
    size: 1.8
    type: urdf
-   center: (1.2, 0, 0.5) 
    lang: a can of soda
    name: Item
    on_table: false 
    path: soda_can.obj
    size: 0.2
    type: mesh
```

I will also give you the articulation tree and semantics file of the articulated object in the task. Such information will be useful for writing the reward function/the primitive actions, for example, when the reward requires accessing the joint value of a joint in the articulated object, or the position of a link in the articulated object, or when the primitive needs to access a name of the object.
```Refrigerator articulation tree
links: 
base
link_0
link_1
link_2

joints: 
joint_name: joint_0 joint_type: fixed parent_link: base child_link: link_0
joint_name: joint_1 joint_type: revolute parent_link: link_0 child_link: link_1
joint_name: joint_2 joint_type: revolute parent_link: link_0 child_link: link_2
```


```Refrigerator semantics
link_0 heavy refrigerator_body
link_1 hinge door
link_2 hinge door
```

I will also give you the links and joints of the articulated object that will be used for completing the task:

Links:
link_1: This link is one of the refrigerator doors, which the robot neesd to reach for the item inside.
Joints:
joint_1: This joint connects link_1, representing one of the doors. The robot needs to actuate this joint to open the door, reach for the item, and close the door.

For each substep, you should decide whether the substep can be achieved by using the provided list of primitives. If not, you should then write a reward function for the robot to learn to perform this substep. 
If you choose to write a reward function for the substep, you should also specify the action space of the robot when learning this reward function. 
There are 2 options for the action space: "delta-translation", where the action is the delta translation of the robot end-effector, suited for local movements; and "normalized-direct-translation", where the action specifies the target location the robot should move to, suited for moving to a target location.
For each substep, you should also write a condition that checks whether the substep has been successfully completed.

Here is a list of primitives the robot can do. The robot is equipped with a suction gripper, which makes it easy for the robot to grasp an object or a link on an object. 
grasp_object(self, object_name): the robot arm will grasp the object specified by the argument object name.
grasp_object_link(self, object_name, link_name): some object like an articulated object is composed of multiple links. The robot will grasp a link with link_name on the object with object_name. 
release_grasp(self): the robot will release the grasped object.
approach_object(self, object_name): this function is similar to grasp_object, except that the robot only approaches the object, without grasping it.
approach_object_link(self, object_name, link_name): this function is similar to grasp_object_link, except that the robot only approaches the object's link, without grasping it.
Note that all primitives will return a tuple (rgbs, final_state) which represents the rgb images of the execution process and the final state of the execution process. 
You should always call the primitive in the following format:
rgbs, final_state = some_primitive_function(self, arg1, ..., argn)

Here is a list of helper functions that you can use for designing the reward function or the success condition:
get_position(self, object_name): get the position of center of mass of object with object_name.
get_orientation(self, object_name): get the orientation of an object with object_name.
detect(self, object_name, object_part): detect the position of a part in object. E.g., the opening of a toaster, or the handle of a door. 
get_joint_state(self, object_name, joint_name): get the joint angle value of a joint in an object.
get_joint_limit(self, object_name, joint_name): get the lower and upper joint angle limit of a joint in an object, returned as a 2-element tuple.
get_link_state(self, object_name, link_name): get the position of the center of mass of the link of an object.
get_eef_pos(self): returns the position, orientation of the robot end-effector as a list.
get_bounding_box(self, object_name): get the axis-aligned bounding box of an object. It returns the min and max xyz coordinate of the bounding box.
get_bounding_box_link(self, object_name, link_name): get the axis-aligned bounding box of the link of an object. It returns the min and max xyz coordinate of the bounding box.
in_bbox(self, pos, bbox_min, bbox_max): check if pos is within the bounding box with the lowest corner at bbox_min and the highest corner at bbox_max. 
get_grasped_object_name(self): return the name of the grasped object. If no object is grasped by the robot, return None. The name is automatically converted to the lower case.
get_grasped_object_and_link_name(self): return a tuple, the first is the name of the grasped object, and the second is the name of the grasped link. If no object is grasped by the robot, return (None, None). The name is automatically converted to the lower case.
gripper_close_to_object(self, object_name): return true if the robot gripper is close enough to the object specified by object_name, otherwise false.
gripper_close_to_object_link(self, object_name, link_name): return true if the robot gripper is close enough to the object link, otherwise false.

You can assume that for objects, the lower joint limit corresponds to their natural state, e.g., a box is closed with the lid joint being 0, and a lever is unpushed when the joint angle is 0.

For the above task "Fetch item from refrigerator", it can be decomposed into the following substeps, primitives, and reward functions:

substep 1: grasp the refrigerator door
```primitive
    rgbs, final_state = grasp_object_link(self, "Refrigerator", "link_1")  
    grasped_object, grasped_link = get_grasped_object_and_link_name(self)
    success = (grasped_object == "Refrigerator".lower() and grasped_link == "link_1".lower())
```

substep 2: open the refrigerator door
```reward
def _compute_reward(self):
    # this reward encourages the end-effector to stay near door to grasp it.
    eef_pos = get_eef_pos(self)[0]
    door_pos = get_link_state(self, "Refrigerator", "link_1")
    reward_near = -np.linalg.norm(eef_pos - door_pos)

    # Get the joint state of the door. We know from the semantics and the articulation tree that joint_1 connects link_1 and is the joint that controls the rotation of the door.
    joint_angle = get_joint_state(self, "Refrigerator", "joint_1") 
    # The reward is the negative distance between the current joint angle and the joint angle when the door is fully open (upper limit).
    joint_limit_low, joint_limit_high = get_joint_limit(self, "Refrigerator", "joint_1")
    target_joint_angle = joint_limit_high
    diff = np.abs(joint_angle - target_joint_angle)
    reward_joint =  -diff

    reward = reward_near + 5 * reward_joint

    success = diff < 0.1 * (joint_limit_high - joint_limit_low)

    return reward, success
```

```action space
delta-translation
```
In the last substep the robot already grasps the door, thus only local movements are needed to open it. 

substep 3: grasp the item
```primitive
    rgbs, final_state = grasp_object(self, "Item")
    success = get_grasped_object_name(self) == "Item".lower()
```

substep 4: move the item out of the refrigerator
```reward
def _compute_reward(self):
    # Get the current item position
    item_position = get_position(self, "Item")

    # The first reward encourages the end-effector to stay near the item
    eef_pos = get_eef_pos(self)[0]
    reward_near = -np.linalg.norm(eef_pos - item_position)

    # The reward is to encourage the robot to grasp the item and move the item to be on the table. 
    # The goal is not to just move the soda can to be at a random location out of the refrigerator. Instead, we need to place it somewhere on the table. 
    # This is important for moving an object out of a container style of task.
    table_bbox_low, table_bbox_high = get_bounding_box(self, "init_table") # the table is referred to as "init_table" in the simulator. 
    table_bbox_range = table_bbox_high - table_bbox_low

    # target location is to put the item at a random location on the table
    target_location = np.zeros(3)
    target_location[0] = table_bbox_low[0] + 0.2 * table_bbox_range[0] # 0.2 is a random chosen number, any number in [0, 1] should work
    target_location[1] = table_bbox_low[1] + 0.3 * table_bbox_range[1] # 0.3 is a random chosen number, any number in [0, 1] should work
    target_location[2] = table_bbox_high[2] # the height should be the table height
    diff = np.linalg.norm(item_position - target_location)
    reward_distance = -diff

    reward = reward_near + 5 * reward_distance

    success = diff < 0.06
    
    return reward, success
```

```action space
normalized-direct-translation
```
Since this substep requires moving the item to a target location, we use the normalized-direct-translation.

substep 5: grasp the refrigerator door again
```primitive
    rgbs, final_state = grasp_object_link(self, "Refrigerator", "link_1")
    grasped_object, grasped_link = get_grasped_object_and_link_name(self)
    success = (grasped_object == "Refrigerator".lower() and grasped_link == "link_1".lower())  
```

substep 6: close the refrigerator door
```reward
def _compute_reward(self):
    # this reward encourages the end-effector to stay near door
    eef_pos = get_eef_pos(self)[0]
    door_pos = get_link_state(self, "Refrigerator", "link_1")
    reward_near = -np.linalg.norm(eef_pos - door_pos)

    # Get the joint state of the door. The semantics and the articulation tree show that joint_1 connects link_1 and is the joint that controls the rotation of the door.
    joint_angle = get_joint_state(self, "Refrigerator", "joint_1") 
    # The reward encourages the robot to make joint angle of the door to be the lower limit to clost it.
    joint_limit_low, joint_limit_high = get_joint_limit(self, "Refrigerator", "joint_1")
    target_joint_angle = joint_limit_low
    
    diff = np.abs(target_joint_angle - joint_angle)
    reward_joint =  -diff

    reward = reward_near + 5 * reward_joint

    success = diff < 0.1 * (joint_limit_high - joint_limit_low)       

    return reward, success
```

```action space
delta-translation
```

I will give some more examples of decomposing the task. Reply yes if you understand the goal.

=====================================

Yes, I understand the goal. Please proceed with the next example.

=====================================

Another example:

Task Name: Set oven temperature
Description: The robotic arm will turn the knob of an oven to set a desired temperature.
Initial config:
```yaml
-   use_table: false
-   center: (1, 0, 0) # when an object is not on the table, the center specifies its location in the world coordinate. 
    lang: a freestanding oven 
    name: oven
    on_table: false
    path: oven.urdf
    size: 0.85
    type: urdf
```

```Oven articulation tree:
links: 
base
link_0
link_1
link_2
link_3
link_4

joints: 
joint_name: joint_0 joint_type: continuous parent_link: link_4 child_link: link_0
joint_name: joint_1 joint_type: continuous parent_link: link_4 child_link: link_1
joint_name: joint_2 joint_type: continuous parent_link: link_4 child_link: link_2
joint_name: joint_3 joint_type: continuous parent_link: link_4 child_link: link_3
joint_name: joint_4 joint_type: fixed parent_link: base child_link: link_4
```

```Oven semantics
link_0 hinge knob
link_1 hinge knob
link_2 hinge knob
link_3 hinge knob
link_4 heavy oven_body
```

Links:
link_0: We know from the semantics that link_0 is a hinge knob. It is assumed to be the knob that controls the temperature of the oven. The robot needs to actuate this knob to set the temperature of the oven.

Joints:
joint_0: from the articulation tree, joint_0 connects link_0 and is a continuous joint. Therefore, the robot needs to actuate joint_0 to turn link_0, which is the knob.

This task can be decomposed as follows:

substep 1: grasp the temperature knob
```primitive
	rgbs, final_state = grasp_object_link(self, "oven", "link_0") 
	grasped_object, grasped_link = get_grasped_object_and_link_name(self)
	success = (grasped_object == "oven".lower() and grasped_link == "link_0".lower())
```

substep 2: turn the temperature knob to set a desired temperature
```reward
def _compute_reward(self):
    # This reward encourages the end-effector to stay near the knob to grasp it.
    eef_pos = get_eef_pos(self)[0]
    knob_pos = get_link_state(self, "oven", "link_0")
    reward_near = -np.linalg.norm(eef_pos - knob_pos)

    joint_angle = get_joint_state(self, "oven", "joint_0") 
    
    joint_limit_low, joint_limit_high = get_joint_limit(self, "oven", "joint_0")
    desired_temperature = joint_limit_low + (joint_limit_high - joint_limit_low)  / 3 # We assume the target desired temperature is one third of the joint angle. It can also be 1/3, or other values between joint_limit_low and joint_limit_high. 

    # The reward is the negative distance between the current joint angle and the joint angle of the desired temperature.
    diff = np.abs(joint_angle - desired_temperature)
    reward_joint =  -diff

    reward = reward_near + 5 * reward_joint

    success = diff < 0.1 * (joint_limit_high - joint_limit_low)

    return reward, success
```

```action space
delta-translation
```

I will provide more examples in the following messages. Please reply yes if you understand the goal.

=====================================

Yes, I understand the goal. Please proceed with the next example.

=====================================

Here is another example:

Task Name:  Put a toy car inside a box
Description: The robotic arm will open a box, grasp the toy car and put it inside the box.
Initial config:
```yaml
-  use_table: True 
-   center: (0.2, 0.3, 0)
    on_table: True
    lang: a box
    name: box
    size: 0.25
    type: urdf
-   center: (0.1, 0.6, 0)
    on_table: True
    lang: a toy car
    name: toy_car
    size: 0.1
    type: mesh
```

```box articulation tree
links: 
base
link_0
link_1
link_2

joints: 
joint_name: joint_0 joint_type: revolute parent_link: link_2 child_link: link_0
joint_name: joint_1 joint_type: revolute parent_link: link_2 child_link: link_1
joint_name: joint_2 joint_type: fixed parent_link: base child_link: link_2
```

```box semantics
link_0 hinge rotation_lid
link_1 hinge rotation_lid
link_2 free box_body
```

Links:
link_0: To fully open the box, the robot needs to open both box lids. We know from the semantics that link_0 is one of the lids.
link_1: To fully open the box, the robot needs to open both box lids. We know from the semantics that link_1 is another lid.
Joints:
joint_0: from the articulation tree, joint_0 connects link_0 and is a hinge joint. Thus, the robot needs to actuate joint_0 to open link_0, which is the lid of the box.
joint_1: from the articulation tree, joint_1 connects link_1 and is a hinge joint. Thus, the robot needs to actuate joint_1 to open link_1, which is the lid of the box.

This task can be decomposed as follows:

substep 1: grasp the first lid of the box
```primitive
	# The semantics shows that link_0 and link_1 are the lid links. 
	rgbs, final_state = grasp_object_link(self, "box", "link_0")  
    grasped_object, grasped_link = get_grasped_object_and_link_name(self)
    success = (grasped_object == "box".lower() and grasped_link == "link_0".lower())
```

substep 2: open the first lid of the box
```reward
def _compute_reward(self):
    # This reward encourages the end-effector to stay near the lid to grasp it.
    eef_pos = get_eef_pos(self)[0]
    lid_pos = get_link_state(self, "box", "link_0")
    reward_near = -np.linalg.norm(eef_pos - lid_pos)

    # Get the joint state of the first lid. The semantics and the articulation tree show that joint_0 connects link_0 and is the joint that controls the rotation of the first lid link_0.
    joint_angle = get_joint_state(self, "box", "joint_0") 
    # The reward is the negative distance between the current joint angle and the joint angle when the lid is fully open (upper limit).
    joint_limit_low, joint_limit_high = get_joint_limit(self, "box", "joint_0")
    target_joint_angle = joint_limit_high

    diff = np.abs(joint_angle - target_joint_angle)
    reward_joint =  -diff

    reward = reward_near + 5 * reward_joint

    success = diff < 0.1 * (joint_limit_high - joint_limit_low)

    return reward, success
```

```action space
delta-translation
```

substep 3: grasp the second lid of the box
```primitive
	# We know from the semantics that link_0 and link_1 are the lid links. 
	rgbs, final_state = grasp_object_link(self, "box", "link_1")  
    grasped_object, grasped_link = get_grasped_object_and_link_name(self)
    success = (grasped_object == "box".lower() and grasped_link == "link_1".lower())
```

substep 4: open the second lid of the box
```reward
def _compute_reward(self):
    # This reward encourages the end-effector to stay near the lid to grasp it.
    eef_pos = get_eef_pos(self)[0]
    lid_pos = get_link_state(self, "box", "link_1")
    reward_near = -np.linalg.norm(eef_pos - lid_pos)

    # Get the joint state of the second lid. The semantics and the articulation tree show that joint_1 connects link_1 and is the joint that controls the rotation of the second lid link_1.
    joint_angle = get_joint_state(self, "box", "joint_1") 
    # The reward is the negative distance between the current joint angle and the joint angle when the lid is fully open (upper limit).
    joint_limit_low, joint_limit_high = get_joint_limit(self, "box", "joint_1")
    target_joint_angle = joint_limit_high

    diff = np.abs(joint_angle - target_joint_angle)
    reward_joint =  -diff

    reward = reward_near + 5 * reward_joint
    
    success = diff < 0.1 * (joint_limit_high - joint_limit_low)
    return reward, success
```

```action space
delta-translation
```

substep 5: grasp the toy car
```primitive
	rgbs, final_state = grasp_object(self, "toy_car")
    success = get_grasped_object_name(self) == "toy_car".lower()
```

substep 6: put the toy car into the box
```reward
def _compute_reward(self):
    # Get the current car position
    car_position = get_position(self, "toy_car")

    # This reward encourages the end-effector to stay near the car to grasp it.
    eef_pos = get_eef_pos(self)[0]
    reward_near = -np.linalg.norm(eef_pos - car_position)

    
    # Get the box body bounding box
    min_aabb, max_aabb = get_bounding_box_link(self, "box", "link_4") # from the semantics, link_4 is the body of the box.
    diff = np.array(max_aabb) - np.array(min_aabb)
    min_aabb = np.array(min_aabb) + 0.05 * diff  # shrink the bounding box a bit
    max_aabb = np.array(max_aabb) - 0.05 * diff
    center = (np.array(max_aabb) + np.array(min_aabb)) / 2

    # another reward is one if the car is inside the box bounding box
    reward_in = 0
    if in_bbox(self, car_position, min_aabb, max_aabb): reward_in += 1

    # another reward is to encourage the robot to move the car to be near the box
    # we need this to give a dense reward signal for the robot to learn to perform this task. 
    reward_reaching = -np.linalg.norm(center - car_position)

    # The task is considered to be successful if the car is inside the box bounding box
    success = in_bbox(self, car_position, min_aabb, max_aabb)

    # We give more weight to reward_in, which is the major goal of the task.
    reward = 5 * reward_in + reward_reaching + reward_near
    return reward, success
```

```action space
normalized-direct-translation
```
Since this substep requires moving the item to a target location, we use the normalized-direct-translation.


Please decompose the following task into substeps. For each substep, write a primitive/a reward function, write the success checking function, and the action space if the reward is used. 

The primitives you can call for the robot to execute:
grasp_object(self, object_name): the robot arm will grasp the object specified by the argument object name.
grasp_object_link(self, object_name, link_name): some object like an articulated object is composed of multiple links. The robot will grasp a link with link_name on the object with object_name. 
release_grasp(self): the robot will release the grasped object.
approach_object(self, object_name): this function is similar to grasp_object, except that the robot only approaches the object, without grasping it.
approach_object_link(self, object_name, link_name): this function is similar to grasp_object_link, except that the robot only approaches the object's link, without grasping it.
Note that all primitives will return a tuple (rgbs, final_state) which represents the rgb images of the execution process and the final state of the execution process. 
You should always call the primitive in the following format:
rgbs, final_state = some_primitive_function(self, arg1, ..., argn)

The APIs you can use for writing the reward function/success checking function:
get_position(self, object_name): get the position of center of mass of object with object_name.
get_orientation(self, object_name): get the orientation of an object with object_name.
get_joint_state(self, object_name, joint_name): get the joint angle value of a joint in an object.
get_joint_limit(self, object_name, joint_name): get the lower and upper joint angle limit of a joint in an object, returned as a 2-element tuple.
get_link_state(self, object_name, link_name): get the position of the center of mass of the link of an object.
get_eef_pos(self): returns the position, orientation of the robot end-effector as a list.
get_bounding_box(self, object_name): get the axis-aligned bounding box of an object. It returns the min and max xyz coordinate of the bounding box.
get_bounding_box_link(self, object_name, link_name): get the axis-aligned bounding box of the link of an object. It returns the min and max xyz coordinate of the bounding box.
in_bbox(self, pos, bbox_min, bbox_max): check if pos is within the bounding box with the lowest corner at bbox_min and the highest corner at bbox_max. 
get_grasped_object_name(self): return the name of the grasped object. If no object is grasped by the robot, return None. The name is automatically converted to the lower case.
get_grasped_object_and_link_name(self): return a tuple, the first is the name of the grasped object, and the second is the name of the grasped link. If no object is grasped by the robot, return (None, None). The name is automatically converted to the lower case.
gripper_close_to_object(self, object_name): return true if the robot gripper is close enough to the object specified by object_name, otherwise false.
gripper_close_to_object_link(self, object_name, link_name): return true if the robot gripper is close enough to the object link, otherwise false.

The action space you can use for learning with the reward: delta-translation is better suited for small movements, and normalized-direct-translation is better suited for directly specifying the target location of the robot end-effector. 
You can assume that for objects, the lower joint limit corresponds to their natural state, e.g., a box is closed with the lid joint being 0, and a lever is unpushed when the joint angle is 0.

Task name: Throw Trash Away
Description: The robotic arm places an item of trash inside the trash can
Initial config:
```yaml
- use_table: false
- center: (1.5, 1.5, 0)
  lang: a common trash can
  name: TrashCan
  on_table: false
  path: trashcan.urdf
  reward_asset_path: '102187'
  size: 0.6
  type: urdf
- center: (0.5, 0.5, 0)
  lang: a pile of paper waste
  name: Trash
  on_table: true
  path: paper_waste.obj
  size: 0.05
  type: mesh

```


```TrashCan articulation tree
links: 
base
link_0
link_1

joints: 
joint_name: joint_0 joint_type: revolute parent_link: link_1 child_link: link_0
joint_name: joint_1 joint_type: fixed parent_link: base child_link: link_1

```


```TrashCan semantics
link_0 hinge door
link_1 free trashcan_body

```

Links:
link_0: Link_0, as the trash can lid, needs to be approached and opened by the robot in order to throw away the trash.
Joints:
joint_0: Joint_0, connecting to link_0, needs to be actuated for the robot to open the trash can lid.
\end{lstlisting}

An example response for the above task ``Throw the trash away''. It is decomposed into multiple substeps, and either action primitive or RL is selected as the algorithm for these substeps. For substep using RL, a reward function is further written. 
\begin{lstlisting}
This task can be decomposed into the following substeps:

substep 1: grasp the trash can lid
```primitive
    rgbs, final_state = grasp_object_link(self, "TrashCan", "link_0")  
    grasped_object, grasped_link = get_grasped_object_and_link_name(self)
    success = (grasped_object == "TrashCan".lower() and grasped_link == "link_0".lower())
```

substep 2: open the trash can lid
```reward
def _compute_reward(self):
    # This reward encourages the end-effector to stay near the lid to grasp it.
    eef_pos = get_eef_pos(self)[0]
    lid_pos = get_link_state(self, "TrashCan", "link_0")
    reward_near = -np.linalg.norm(eef_pos - lid_pos)

    # Get the joint state of the lid. The semantics and the articulation tree show that joint_0 connects link_0 and is the joint that controls the rotation of the lid.
    joint_angle = get_joint_state(self, "TrashCan", "joint_0") 
    # The reward is the negative distance between the current joint angle and the joint angle when the lid is fully open (upper limit).
    joint_limit_low, joint_limit_high = get_joint_limit(self, "TrashCan", "joint_0")
    target_joint_angle = joint_limit_high

    diff = np.abs(joint_angle - target_joint_angle)
    reward_joint =  -diff

    reward = reward_near + 5 * reward_joint

    success = diff < 0.1 * (joint_limit_high - joint_limit_low)

    return reward, success
```

```action space
delta-translation
```

substep 3: grasp the trash
```primitive
    rgbs, final_state = grasp_object(self, "Trash")
    success = get_grasped_object_name(self) == "Trash".lower()
```

substep 4: put the trash into the trash can
```reward
def _compute_reward(self):
    # Get the current trash position
    trash_position = get_position(self, "Trash")

    # This reward encourages the end-effector to stay near the trash to grasp it.
    eef_pos = get_eef_pos(self)[0]
    reward_near = -np.linalg.norm(eef_pos - trash_position)

    # Get the trash can body bounding box
    min_aabb, max_aabb = get_bounding_box_link(self, "TrashCan", "link_1") # from the semantics, link_1 is the body of the trash can.
    diff = np.array(max_aabb) - np.array(min_aabb)
    min_aabb = np.array(min_aabb) + 0.05 * diff  # shrink the bounding box a bit
    max_aabb = np.array(max_aabb) - 0.05 * diff
    center = (np.array(max_aabb) + np.array(min_aabb)) / 2

    # another reward is one if the trash is inside the trash can bounding box
    reward_in = 0
    if in_bbox(self, trash_position, min_aabb, max_aabb): reward_in += 1

    # another reward is to encourage the robot to move the trash to be near the trash can
    # we need this to give a dense reward signal for the robot to learn to perform this task. 
    reward_reaching = -np.linalg.norm(center - trash_position)

    # The task is considered to be successful if the trash is inside the trash can bounding box
    success = in_bbox(self, trash_position, min_aabb, max_aabb)

    # We give more weight to reward_in, which is the major goal of the task.
    reward = 5 * reward_in + reward_reaching + reward_near
    return reward, success
```

```action space
normalized-direct-translation
```
Since this substep requires moving the item to a target location, we use the normalized-direct-translation.

substep 5: release the trash
```primitive
    rgbs, final_state = release_grasp(self)
    success = get_grasped_object_name(self) == None
```

substep 6: grasp the trash can lid again
```primitive
    rgbs, final_state = grasp_object_link(self, "TrashCan", "link_0")
    grasped_object, grasped_link = get_grasped_object_and_link_name(self)
    success = (grasped_object == "TrashCan".lower() and grasped_link == "link_0".lower())  
```

substep 7: close the trash can lid
```reward
def _compute_reward(self):
    # This reward encourages the end-effector to stay near the lid
    eef_pos = get_eef_pos(self)[0]
    lid_pos = get_link_state(self, "TrashCan", "link_0")
    reward_near = -np.linalg.norm(eef_pos - lid_pos)

    # Get the joint state of the lid. The semantics and the articulation tree show that joint_0 connects link_0 and is the joint that controls the rotation of the lid.
    joint_angle = get_joint_state(self, "TrashCan", "joint_0") 
    # The reward encourages the robot to make joint angle of the lid to be the lower limit to close it.
    joint_limit_low, joint_limit_high = get_joint_limit(self, "TrashCan", "joint_0")
    target_joint_angle = joint_limit_low
    
    diff = np.abs(target_joint_angle - joint_angle)
    reward_joint =  -diff

    reward = reward_near + 5 * reward_joint

    success = diff < 0.1 * (joint_limit_high - joint_limit_low)       

    return reward, success
```

```action space
delta-translation
```
\end{lstlisting}

\end{document}